\documentclass[preprint,11.5pt,authoryear]{elsarticle}

\usepackage{amssymb}
\usepackage{amsthm, amssymb, amsmath}
\newtheorem{Proposition}{Proposition}

\usepackage{hyperref}

\journal{Omega}

\begin{document}

\begin{frontmatter}



\title{A hybrid machine learning framework for analyzing human decision making through learning preferences}


\author[label1,label2]{Mengzhuo Guo}
\ead{mengzhguo2-c@my.cityu.edu.hk}
\author[label2]{Qingpeng Zhang\corref{cor1}}
\ead{qingpeng.zhang@cityu.edu.hk}
\author[label1]{Xiuwu Liao}
\ead{liaoxiuwu@mail.xjtu.edu.cn}
\author[label3]{Frank Youhua Chen}
\ead{youhchen@cityu.edu.hk}
\author[label4,label5]{Daniel Dajun Zeng}
\ead{dajun.zeng@ia.ac.cn}

\address[label1]{School of Management, Xi'an Jiaotong University}
\address[label2]{School of Data Science, City University of Hong Kong}
\address[label3]{Department of Management Science, City University of Hong Kong}
\address[label4]{The State Key Laboratory of Management and Control for Complex Systems, Institute of Automation, Chinese Academy of Sciences}
\address[label5]{Department of Management Information Systems, University of Arizona}
\cortext[cor1]{Corresponding author}
\begin{abstract}
Machine learning has recently been widely adopted to address the managerial decision making problems, in which the decision maker needs to be able to interpret the contributions of individual attributes in an explicit form. However, there is a trade-off between performance and interpretability. Full complexity models (such as neural network-based models) are non-traceable black-box, whereas classic interpretable models (such as logistic regression) are usually simplified with lower accuracy. This trade-off limits the application of state-of-the-art machine learning models in management problems, which requires high prediction performance, as well as the understanding of individual attributes' contributions to the model outcome. Multiple criteria decision aiding (MCDA) is a family of analytic approaches to depicting the rationale of human decision. It is also limited by strong assumptions (e.g. preference independence). To meet the decision maker's demand for more interpretable machine learning models, we propose a novel hybrid method, namely \textbf{N}eural \textbf{N}etwork-based \textbf{M}ultiple \textbf{C}riteria \textbf{D}ecision \textbf{A}iding (NN-MCDA), which combines an additive value model and a fully-connected multilayer perceptron (MLP) to achieve good performance while capturing the explicit relationships between individual attributes and the prediction. NN-MCDA has a linear component (in an additive form of a set of polynomial functions) to characterize such relationships through providing explicit marginal value functions, and a nonlinear component (in a standard MLP form) to capture the implicit high-order interactions between attributes and their complex nonlinear transformations. We demonstrate the effectiveness of NN-MCDA with extensive simulation studies and three real-world datasets. To the best of our knowledge, this research is the first to enhance the interpretability of machine learning models with MCDA techniques. The proposed framework also sheds light on how to use machine learning techniques to free MCDA from strong assumptions.
\end{abstract}



\end{frontmatter}



\section{Introduction}
\label{Sec-intro}

Machine learning has recently been widely adopted to address challenging decision making problems in a variety of managerial contexts such as marketing \citep{cui2005prediction}, credit-risk evaluation \citep{baesens2003using} and healthcare management \citep{gartner2015machine}. Many machine learning models, such as support vector machines (SVMs) \citep{cortes1995support}, boosted trees \citep{friedman2001greedy} and neural network-based deep learning methods \citep{lecun2015deep}, are capable of handling high-dimensional data because of the high complexity of the model. However, it comes at the expense of interpretability \citep{lou2012intelligible}. 

The application of machine learning in management research could benefit from the interpretability of models. In practice, the decision maker (DM) needs to be able to characterize the contributions of attributes in an explicit form. Such interpretability can make the model trustworthy to the DM \citep{ribeiro2016should}, help DM understand the causality \citep{miller2018explanation}, and improve the model through incorporating DM's domain knowledge \citep{aggarwal2019modelling}. Therefore, technology giants like Google, IBM, and Microsoft, have been investigating on the techniques in enhancing the model interpretability recently. As stated in a comprehensive overview conducted by David Gunning, the program manager in the Information Innovation Office (I2O) of the Defense Advanced Research Projects Agency (DARPA), ``\textit{machine learning models are opaque, non-intuitive and difficult for people to understand}'' \citep{gunning2017explainable}. DARPA has since funded for developing interpretable machine learning techniques among academics. In the latest budget plan of DARPA, explainable artificial intelligence (XAI) has been listed as the key funding area in the fiscal year 2019-2020, with the total amount of 26.05 million US dollars \footnote{\url{https://www.darpa.mil/about-us/budget}}. 

\subsection{The need for interpretable models in management problems.}
\label{subsec-whobene}

In many management problems, the ability in understanding the contributions of individual attributes to the prediction outcome is heavily needed \citep{guo2019consumer, barbati2018optimization}. An explicit form of such contributions, for instance giving a value function that describes the detailed relationship between each attribute and outcome, can help the DM exploit and correct the obtained patterns and rules with prior knowledge, facilitate the downstream managerial decision making, and incorporate critical domain knowledge \citep{miller2018explanation, ribeiro2016should, doumpos2011preference, lou2012intelligible}. The resulted interpretable model can provide in-depth understanding of the data and patterns \citep{aggarwal2019modelling}. In practice, model interpretability is as important as (if not more important than) accuracy in many managerial decision making tasks such as clinical diagnoses, in which the understanding of how the model makes the prediction is the key to facilitate physicians to trust the model and utilize the prediction results \citep{ caruana2015intelligible}. For example, the risk of depression is usually assumed to be positively correlated with the age \citep{blazer1991association}. However, the strength of the correlation could vary over the course of aging \citep{li2014meta}. A simple regression analysis assumes that such correlation is consistent, which sometimes simplify the real scenarios. If we can provide the physicians with a value function that characterizes the age's contribution on the risk with a more explicit and concrete form, the predictions seem to be more convincing. In this study, by interpretability we mean that the model capacity to provide such explicit value functions characterizing the contributions of each individual attributes \citep{lou2012intelligible}.

\subsection{Multiple criteria decision aiding}
\label{subsec-mcda-intro}

Multiple criteria decision aiding (MCDA) has been a fast growing area of operational research during the last several decades \citep{dyer1992multiple, stewart1992critical, wallenius2008multiple, ciomek2018predictive}. It involves a finite set of alternatives (e.g. actions, items, policies) that are evaluated from a set of conflicting multiple criteria or attributes\footnote{In machine learning, criteria refer to attributes or features with preference order scales \citep{corrente2013robust}. For consistency, we use ``attribute'' in this paper.}. The DM's decision is driven by his/her underlying \textit{global value (utility) function} \citep{keeney1976group}. This global value measures the DM's desirability for an alternative and can be disaggregated into a set of per-attribute \textit{marginal value functions} that represent the DM's evaluation of the corresponding attribute. These marginal value functions can be learned by the DM's judgments on learning examples (e.g. pairwise comparison between two alternatives). Once the marginal value functions are deciphered, we can understand the decision making rationale, based on which we can predict the judgment of the DM. This process is referred as the preference disaggregation approaches of MCDA.

Many machine learning framework can help MCDA accomplish the learning objectives because both of them aim to learn a decision model from data. Thus, MCDA and machine learning naturally have reciprocal interactions \citep{doumpos2011preference}. MCDA and machine learning are integrated in two directions. First, we can apply machine learning techniques to various tasks in a decision aiding context, such as learning to rank, multi-label classification, etc. The opposite direction is to implement MCDA concepts in a machine learning framework. Utilizing MCDA approaches to adapt the machine learning models to various topics (such as feature selection and extraction, pruning decision rules and multiple objective optimization) has become a trend recently. Our work belongs to the second stream. We aim to construct a hybrid model, which utilizes value function-based preference disaggregation approaches of MCDA to enhance the interpretability of ``black-box'' machine learning models. 

The motivation of introducing the value function-based preference disaggregation approaches of MCDA to machine learning stems from its powerful capacity in depicting the human decision-making process. The deciphered marginal value functions reveal the rationale of DM's judgment, and thus provide convincing evidence to assist comprehending the decision making \citep{aggarwal2019modelling, lou2012intelligible}. Our task of learning an interpretable model is essentially to capture the characteristics of the marginal value functions, based on which we obtain a certain degree of interpretability.

\subsection{An overview of this paper.}
\label{subsec-overpaper}

This paper proposes a framework for a \textbf{N}eural \textbf{N}etwork-based \textbf{M}ultiple \textbf{C}riteria \textbf{D}ecision \textbf{A}iding (NN-MCDA) approach. NN-MCDA combines an additive model and a fully-connected multilayer perceptron (MLP) to achieve good performance while capturing the explicit
relationships between individual attributes and the prediction. The additive model uses marginal value functions to approximate the explicit relationship between the outcome and individual attributes whereas the MLP is used to capture the implicit high-order correlations between attributes in the model. We estimate the parameters in the model under a neural network framework that automatically balances the trade-off between two components.

We validate the proposed model using a set of synthetic datasets and three real datasets. Specifically, the simulation experiments respectively show the impact of pre-defined parameters on model performance when data is either extremely complex or simple. Three real datasets on ranking universities regarding employment reputation, predicting the risk for geriatric depression, and predicting the success of bank telemarketing are used to illustrate the proposed model in real scenarios. We explain the obtained model and compare its performance with that of baseline interpretable models, i.e., GAM and logistic regression models, for the first two datasets. We further verify the efficacy of the proposed model with the third dataset through comparing its performance with the results reported in the literature.

The contributions of this paper are fourfold. First, we advocate a new perspective of a hybrid machine learning model that both quantifies the explicit impact of individual attributes on the outcome and captures the implicit high-order correlations among attributes. It helps the DM understand the main effect of single attribute, and at the same time, make better decisions. Second, to the best of our knowledge, this paper is the first pilot work that introduces the value function-based preference disaggregation approaches of MCDA to the machine learning models to enhance the model interpretability. The trained parameters in the proposed framework determine the shape for marginal value functions in the additive models. The proposed model is free from preference independence, preference monotonicity, and small learning set assumptions in MCDA approaches, and thus makes MCDA approaches more general and practical for real-world management problems. Third, we examine the model effectiveness given different model parameters and datasets. The empirical conclusions about the relationships between model interpretability and data complexity are managerially intuitive for DM and insightful for future research. Fourth, the proposed framework is flexible and extendible, especially the nonlinear part, which can be modified or replaced by other network structures or modeling schemes according to different tasks.

The rest of the paper is organized as follows. We discuss the related work in Section \ref{sec-related}. In Section \ref{sec-framework}, we introduce the framework for the proposed model. The simulation and real case experiments are presented in Section \ref{sec-experiment} and some discussions about the proposed framework is provided in Section \ref{sec-dis}. We conclude the paper in Section \ref{sec-conc}.

\section{Related work.}
\label{sec-related}

\subsection{Value function-based preference disaggregation approach of MCDA.}
\label{subsec-MCDA}

The value function-based preference disaggregation approaches of MCDA provide explicit marginal value functions and numerical scores. A DM can understand the importance of a particular attribute and how the individual attributes contribute to the final decision. This procedure encourages the DM to participate in the decision making process and it provides a comprehensive preference model. These approaches have been successfully applied to many scenarios, such as consumer preference analysis \citep{guo2019consumer}, financial decisions \citep{barbati2018optimization}, nano-particles synthesis assessment \citep{kadzinski2018co}, territorial transformation management \citep{ciomek2018predictive}, and medical therapy \citep{hasan2019multi}. However, the applications of value function-based preference disaggregation approaches are limited due to some strong assumptions, such as (1) preference independence, (2) monotonic preference, and (3) small set of alternatives.

Recently, many novel models haven been proposed to generalize the value function-based preference disaggregation approaches of MCDA. Preference independence allows the model to be additive. Considering interacted attributes, \cite{angilella2010non} utilize a fuzzy measure to model the preference system where the alternatives are now evaluated in terms of the Choquet integral. However, it is difficult for the DM to understand the impact of individual attribute evaluated from the Choquet integral. \cite{angilella2014musa} account for positive and negative interactions among attributes, and add an interaction term to the additive global value function for each alternative. They require the DM to provide some knowledge about the interacted pairs that are mined by the models. These studies only consider the interaction between pairs of attributes because higher-order interactions require more cognitive efforts and more computational cost.

The majority of existing researches assume the marginal value functions are monotonic piece-wise linear. This assumption reduces the model complexity, but it fails to describe preference inflexions. Addressing this problem, \cite{ghaderi2017linear} and \cite{liu2019preference} relax this assumption and constrain on variations of the slope to obtain non-monotonic marginal value functions without serious over-fitting problem. Both of their approaches obtain non-smooth value functions which are difficult to interpret attitudes towards risks due to the use of non-derivative functions. Since a differentiable marginal value function is essential to analyze consumer behavior, \cite{sobrie2018uta} utilize semidefinite programming to infer the key parameters for polynomial marginal value functions. It gives a more flexible and interpretable preference model. However, it still assumes that the DM preference is monotonic.

The monotonic piece-wise form of the marginal value functions has a low expressibility for large learning sets \citep{sobrie2018uta}. Nowadays, MCDA approaches are expected to deal with large amount of data in many disciplines \citep{pelissari2019smaa}. \cite{liu2019preference} embed the MCDA approach into a regularization framework to approximate marginal value functions in any piece-wise linear shapes, and provide efficient algorithms to handle larger learning sets.

Most existing researches focus on expanding the MCDA approaches from only one perspective. Comparing with these recent advances, the proposed framework tries to solve all aforementioned limitations of MCDA by providing a non-monotonic, smoother, and more powerful MCDA approach for real-world applications considering more complex decision making scenarios.

\subsection{Interpretable models in management.}
\label{subsec-interModel}

Generalized additive model (GAM) uses a link function to build a connection between the mean of the prediction and a smooth function of the variables \citep{hastie1986generalized}. It is good at both dealing with and presenting the nonlinear and non-monotonic relationship between the variables and the prediction \citep{lou2012intelligible}. Therefore, GAM is usually more accurate than linear additive models. Although GAM does not outperform full complexity models, it possesses more interpretability than these ``black-box'' models. \cite{lou2013accurate} explore the co-effect of pairwise interactions and apply the improved GAM to predicting pneumonia risk and 30-day readmission. This model helps the DM (physician) to find useful patterns in the data and quantifies the contributions of individual attributes. Based on these promising results, they argue that it is necessary to develop more interpretable models in mission-critical applications such as management problems \citep{caruana2015intelligible}.

A recent work proposes a structured-effect model for forecasting the remaining useful life of machines \citep{kraus2019forecasting}. The model combines non-parametric approaches and a deep recurrent neural network. The benefit of the former part is to provide a linear structure where one can understand the importance of the predictors by comparing the obtained coefficients, while the latter helps to improve the flexibility of the model, thereby increasing the model performance. The proposed model can achieve state-of-the-art performance compared to some traditional approaches but remains interpretable. 

Another solution for explaining the predictions is to infer a new model to approximate the true black-box model. The new model may not be as accurate as the original black-box model, but can identify patterns and rules to explain how the predictions are made. In \cite{baesens2003using}, explanatory rules are extracted to help the credit-risk managers in explaining their decisions. Similarly, \cite{ letham2015interpretable} discretize a high-dimensional attribute space into a series of simpler interpretable \textit{if-then} statements. They firstly make predictions using complex machine learning techniques and then use Bayesian rule lists to reconstruct the stroke predictions. Given approximately accurate predictions, the obtained model is more interpretable.

\section{Framework for the intelligible model.}
\label{sec-framework}

Let $\mathcal{D} = \{ (\mathbf{x}_i,y_i) \}^N_1$ be the dataset of size $N$, $\mathbf{x}_i = (x_{i,1},\cdots,x_{i,n})^T$ be the $i$-th attribute vector with $n$ attributes\footnote{In MCDA, $\mathbf{x}_i$ is called an alternative with $n$ criteria/attributes.}, and $y_i$ be the target/response value. In this study, we consider a binary classification problem where $y_i \in \{0, 1\}$. The proposed framework can be easily extended to multi-classification and regression problems.

\subsection{The additive model.}
\label{subsec-addiModel}

The value function-based preference disaggregation approaches of MCDA assume that for each attribute vector $\mathbf{x}_i \subseteq \mathbb{R}^n$, there is a global value function in the following form:
\begin{equation}
F(\mathbf{x}_i) = w_1\cdot v_1(x_{i,1})+w_2\cdot v_2(x_{i,2})+\cdots + w_n\cdot v_n(x_{i,n}) = \sum\nolimits_{j=1}^{n} {w_j v_j(x_{i,j})} 
\label{eq-global-val}
\end{equation}
where $0 \leq w_j \leq 1, j= \{ 1,2,\cdots,n\}$ represents the importance of the $j$-th attribute and $v_j(\cdot)$ is a marginal value function. Note that we reply the shape and positive/negative effect of the marginal value function to capture the contribution of individual attributes. Thus we set the weight $w_j$ to be positive to represent the relative importance of the $j$-th attribute, which can positively or negatively affect the global value. The \textit{global value function} $F(\cdot)$ linearly sums contributions of individual attributes.

Although the global value function is in an additive linear form, the marginal value functions themselves can be in any forms, often nonlinear. It has been recognized that the preference in human decision behaviors is \textbf{rational}, and thus the marginal value functions should be stable and smooth. In the literature, the marginal value function can be in a simple linear (weighted sum) form \citep{rezaei2016best, korhonen2012can}, monotonic and non-monotonic piecewise linear forms \citep{stewart1993use, jacquet2001preference, greco2008ordinal, ghaderi2017linear}, and monotonic polynomial form \citep{sobrie2018uta}. To capture the first-order (e.g. monotonicity) and second-order (e.g. marginal rate in substitution) derivative patterns of the attributes' contributions to the prediction, we extend and generalize state-of-the-art MCDA models \citep{liu2019preference, sobrie2018uta} to allow the marginal value function in any polynomial forms. In this paper, we allow the $j$-th marginal value function to be in a smooth and non-monotonic form of $D_j$ degrees:
\begin{equation}
v_j(x_{i,j}) = p_{j,1}x_{i,j}^1 + p_{j,2}x_{i,j}^2 + \cdots + p_{j,D_j}x_{i,j}^{D_j}
\label{eq-marginal-val}
\end{equation}
where $p_{j,d} \in \mathbb{R}, d =\{ 1,2,\cdots,D_j\}$ is the coefficient of the $d$-th degree and $D_j$ is the highest order of degree on the $j$-th attribute.

The motivations using Eq.(\ref{eq-marginal-val}) as a marginal value function are derived from two facets. First, we enhance the expressiveness of the preference model to capture non-monotonic preferences. For example, piecewise linear or monotonic polynomial functions fail to restore all information in a larger learning set \citep{sobrie2018uta}. The nonlinearity and non-monotonicity of Eq.(\ref{eq-marginal-val}) can better fit complex relationships between attributes and the outcome, leading to a better model performance. Second, while analyzing human behavior, it is critical to examine the trade-offs or marginal rates of substitution in economics and management studies. A non-derivative value function, for instance the boosted bagged trees model in \cite{lou2012intelligible}, cannot capture the inflexion point where the marginal rate of substitution grows or diminishes more quickly. A model exploiting human behaviors seems convincing and has more managerial meaning for the DM in management scenarios. 

\subsection{Neural network-based MCDA.}
\label{subsec-nnmcda}

Full complexity models perform well on many machine learning tasks because they can model both the nonlinearity and the interactions between attributes. An additive model like Eq.(\ref{eq-global-val}) does not model any interactions between attributes. Therefore, we propose a neural network-based multiple criteria decision aiding (NN-MCDA) model in the following form
\begin{eqnarray}
U(\mathbf{x}_i) &&=\alpha\sum\nolimits_{j=1}^{n} {w_j v_j(x_{i,j})} + (1-\alpha) f(x_{i,1},x_{i,2},\cdots,x_{i,n}) \label{eq-global-inter-1}
\\&&=  \alpha F(\mathbf{x}_i) + (1-\alpha) f(x_{i,1},x_{i,2},\cdots,x_{i,n})
\label{eq-global-inter}
\end{eqnarray}
where $U(\mathbf{x}_i)$ is the \textit{global score} of $\mathbf{x}_i$, $f(\cdot)$ is a latent function of all attributes, and $\alpha \in [0,1]$ is a trade-off coefficient. Eq.(\ref{eq-global-inter}) describes (a) a regression model if $U(\cdot)$ is the identity, and (b) a classification model if $U(\cdot)$ is the logistic function of the identity. $f(\cdot)$ is used to capture the high-order interrelations between attributes in the model. We can use any complexity models to fit $f(\cdot)$ for better performance, for instance we use a MLP in this paper \citep{rosenblatt1958perceptron}. While using an MLP form of $f(\cdot)$, it is not transparent, meaning that we do not know the exact structure of $f(\cdot)$. Since we have the $F(\cdot)$ to capture the explainable form of the marginal value functions, the non-transparent $f(\cdot)$ describes the complex patterns that are not readily useful to the DM. Coefficient $\alpha$ balances the trade-off between $F(\cdot)$ and $f(\cdot)$. If $\alpha$ is close to 1, the model tends to be in a simple additive form. If $\alpha$ is close to 0, we obtain a full complexity model.

\begin{figure}[htbp]
\centering
{\includegraphics[scale=0.4]{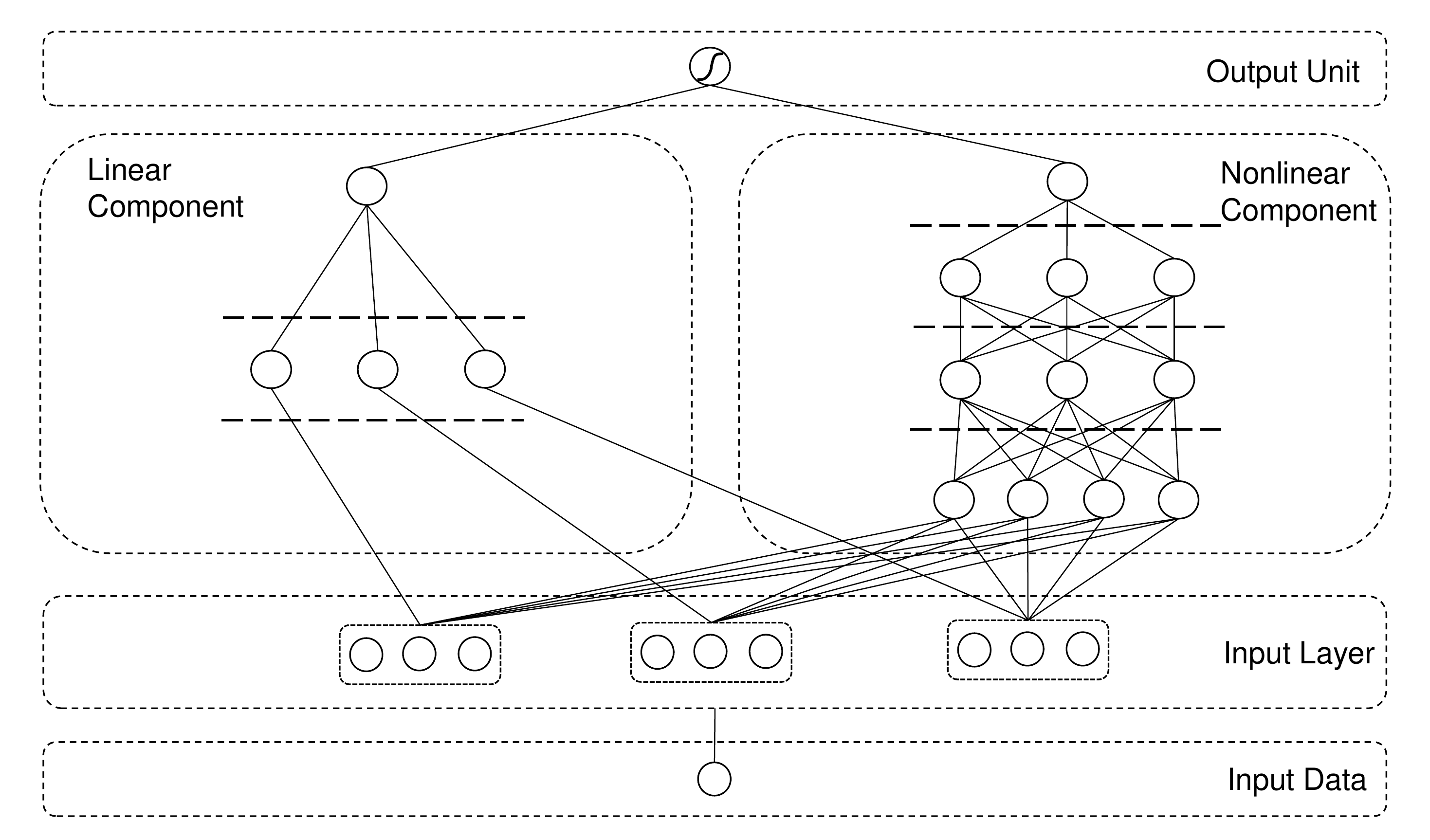}}
\caption{The framework for NN-MCDA.\label{fig-framework}}
\end{figure}

The utilized joint training process is shown in Figure \ref{fig-framework}. The input attribute vectors should be transformed into polynomial forms, i.e., $\Phi(\mathbf{x}_i) =  (x_{i,1}^1,\cdots,x_{i,1}^{D_1},x_{i,2}^1,\cdots,x_{i,2}^{D_2},\cdots,x_{i,n}^{1},\cdots,x_{i,n}^{D_n})^T$. In the input layer, a single-layer network without any activation functions is provided to reconstruct Eq.(\ref{eq-global-val}). It has $\sum\nolimits_{j = 1}^m { D_j }$ units and the weight for each unit corresponds to a particular $p_{j,d}$. We denote the output of the linear component with $z_i^{linear}$, and 
\begin{eqnarray}
\label{eq-z}
z_i^{linear} &&= (w_1, \cdots, w_j, \cdots, w_n) \left( \begin{array}{l}
{\mathbf{p}}_1 ^T \phi ({x_{i,1}})\\
\quad \quad \vdots \\
{\mathbf{p}}_j ^T \phi ({x_{i,j}})\\
\quad \quad \vdots \\
{\mathbf{p}}_m ^T \phi ({x_{i,n}})
\end{array} \right) = \mathbf{w}^T \mathbf{P}_i
\end{eqnarray}
where $\mathbf{p}_j = (p_{j,1},p_{j,2},\cdots,p_{j,D_j})^T$ is the vector of coefficients in the $j$-th polynomial marginal value function, $\phi(x_{i,j})=(x_{i,j}^1,\cdots,x_{i,j}^{D_j})^T$ is the transformed input vector, $\mathbf{w}$ is the vector of weights of attributes, and $\mathbf{P}_i$ contains marginal values of $i$-th attribute vector. Note that, Eq.(\ref{eq-z}) is actually a specific case of Eq.(\ref{eq-global-val}). In Eq.(\ref{eq-global-val}), the marginal value functions $v(\cdot)$ can be in \textit{any} shapes (e.g. piece-wise linear). However, in this study, we allow them to be in a polynomial form in Eq.(\ref{eq-marginal-val}). Thus, $F(\cdot)$ is a generalization of $z_i^{linear}$.

The nonlinear component is a standard MLP. It is used to learn high-order correlations between attributes. Similarly, by summing every $D_j$ units we can obtain a marginal value on the $j$-th attribute. For activation functions, we opt for Rectifier (ReLU), which is is the most commonly used activation function in neural networks \citep{glorot2011deep}. We can also use other activation functions such as Sigmoid and TanH functions. An $L$-layer MLP is defined as:
\begin{eqnarray}
\nonumber
\mathbf{z}_1(\mathbf{x}_i) &&= \Phi(\mathbf{x}_i),\\
\nonumber
\mathbf{z}_2(\mathbf{x}_i) &&= a_1(\mathbf{W}_1^T \mathbf{z}_1 + \mathbf{b}_1),\\
\cdots \\
\nonumber
\mathbf{z}_L(\mathbf{x}_i) &&= a_{L-1}(\mathbf{W}_{L-1}^T \mathbf{z}_{L-1} + \mathbf{b}_{L-1}), \\
\nonumber
{z}_{i}^{nonlinear} &&= \mathbf{h}^T \mathbf{z}_L(\mathbf{x}_i),
\end{eqnarray}
where $\mathbf{W}_l$, $\mathbf{b}_l $ and $a_l $ denote the weight matrix, bias vector and activation function for the $l$-th layer, respectively. The input of the MLP model is the same as the input for the linear part, i.e., $\Phi (\mathbf{x}_i)$.

The output is the probability of $y_i = 1$, we have
\begin{equation}
P(\hat{y}_i=1 | \mathbf{x}_i)=\sigma(\alpha {z}_{i}^{linear} + (1-\alpha) {z}_{i}^{nonlinear})
\end{equation}
where $\sigma(\cdot)$ is a sigmoid function. To estimate the parameters, we minimize the mean square error (MSE):
\begin{equation}
MSE = \frac{1}{N} \sum\nolimits_{i = 1}^N { (P(\hat{y}_{i} = 1|\mathbf{x}_i) - y_i)^2} 
\label{eq-mse}
\end{equation}

We can adopt a variety of optimization methods to minimize Eq.(\ref{eq-mse}), such as Stochastic Gradient Descent (SGD), Adaptive Gradient Algorithm (Adagrad) and Adaptive Moment Estimation (Adam). Please refer to \cite{le2011optimization} for details of the optimization procedure. The interpretability of the model refers to the capacity in developing marginal value functions, which capture the relationship between individual attributes and prediction. With the proposed model, the DM can know what attributes are more important for the prediction, what values of an attribute are positively or negatively associated to the prediction, and where the convexity and concavity of the function are changed.

\begin{Proposition}
When $a_l(\cdot), l=1,\dots,L$ are linear activation functions, the proposed NN-MCDA model degenerates to an additive model.\label{prop-1}
\end{Proposition}
\proof{See \ref{app-1}}

\begin{Proposition}
When $a_1(\cdot)$ is a nonlinear activation function, if $L=1$, the lower-order interacting attributes can be explicitly expressed. If $L\ge2$, the higher-order interacting attributes cannot be explicitly expressed. \label{prop-2}
\end{Proposition}
\proof{See \ref{app-2}}

Given Propositions \ref{prop-1} and \ref{prop-2}, a neural network with nonlinear activation functions and more than one layers will bring extremely complex transformations of attribute interactions that are hard to be explicitly presented, thereby detecting the main effects of the individual attributes in the presence of these interactions is an alternative way to interpret the predictions.

\subsection{Application to multiple criteria ranking problems.}
\label{subsec-apptoMCDA}

In this subsection, we will show how to apply NN-MCDA to traditional multiple criteria ranking problems where alternatives are ranked based on the DM's judgment. In this paper, alternatives are represented as attribute vectors.

Let $\mathbf{x}_i \succsim \mathbf{x}_k$ denote that an attribute vector $\mathbf{x}_i$ is at least as good as $\mathbf{x}_k$, and $\mathbf{x}_i \succ \mathbf{x}_k$ denote that $\mathbf{x}_i$ is better than $\mathbf{x}_k$. Note that the symbol `$\succsim$'(or `$\succ$') does not necessarily require that each element in $\mathbf{x}_i$ is at least as good as (or better than) that in $\mathbf{x}_k$. It actually indicates that one alternative is at least as good as (or better than) another one based on the DM's judgment. For each pair $(\mathbf{x}_i, \mathbf{x}_k)$, we define $y_{i,k}$ as follows:
\begin{equation}
{{y_{i,k} = }}\left\{ \begin{array}{l}
1,\quad \textit{if} \quad \mathbf{x}_i \succsim \mathbf{x}_k,\\
0,\quad \textit{if} \quad \mathbf{x}_k \succ \mathbf{x}_i,
\end{array} \right.
\end{equation}
and the difference between global scores of $\mathbf{x}_i$ and $\mathbf{x}_k$ is:
\begin{eqnarray*}
U(\mathbf{x}_i) - U(\mathbf{x}_k) &&= \alpha \sum\nolimits_{j = 1}^n {{w_j}{v_j}(x_{i,j})} +  (1-\alpha) f(\mathbf{x}_i) \\ && \quad - [ \alpha \sum\nolimits_{j = 1}^n {{w_j}{v_j}(x_{k,j})} + (1-\alpha) f(\mathbf{x}_k)] \nonumber \\
&&= \alpha \sum\nolimits_{j = 1}^n {{w_j} ( {v_j}(x_{i,j}) - {v_j}(x_{k,j}) )} + (1-\alpha) [f( \mathbf{x}_i ) - f(\mathbf{x}_k)] \nonumber \\
&&= \alpha \sum\nolimits_{j = 1}^n {{w_j}\sum\nolimits_{d = 1}^{{D_j}} {p_{j,d}(x_{i,j}^d - x_{k,j}^d)} } + (1-\alpha) [f( \mathbf{x}_i ) - f(\mathbf{x}_k)] \nonumber \\
&&= \alpha \sum\nolimits_{j = 1}^n {{w_j}(\mathbf{p}_j^T \times \phi {{({x}_{i,j},{x}_{k,j})}})}  + (1-\alpha) \Theta(\Phi(\mathbf{x}_i, \mathbf{x}_k))  \nonumber \\
&&= \alpha \mathbf{w}^T ( \mathbf{P}_i - \mathbf{P}_k) + (1-\alpha) \Theta(\Phi(\mathbf{x}_i, \mathbf{x}_k))\nonumber  \label{eq-global-value}
\end{eqnarray*}
where $\phi(x_{i,j},x_{k,j}) = (x_{i,j}^1 - x_{k,j}^1,x_{i,j}^2 - x_{k,j}^2,\cdots, x_{i,j}^{D_j} - x_{k,j}^{D_j})^T$. Let $\Phi(\mathbf{x}_i, \mathbf{x}_k)$ be the aggregated vector for $\phi(x_{i,j},x_{k,j})$:
\begin{equation}
\footnotesize
\Phi(\mathbf{x}_i, \mathbf{x}_k) = \left( \underbrace {x_{i,1}^1 - x_{k,1}^1, \ldots x_{i,1}^{{D_1}} - x_{k,1}^{{D_1}}}_{{D_1}},\underbrace {x_{i,2}^1 - x_{k,2}^1, \ldots ,x_{i,2}^{{D_2}} - x_{k,2}^{{D_2}}}_{{D_2}},\underbrace {x_{i,3}^1 - x_{k,3}^1, \ldots }_{{D_3}}, \ldots ,\underbrace { \ldots ,x_{i,n}^{{D_n}} - x_{k,n}^{{D_n}}}_{{D_n}} \right) ^T
\nonumber
\end{equation}
and $\Theta(\Phi(\mathbf{x}_i, \mathbf{x}_k))$ be a function of $\Phi(\mathbf{x}_i, \mathbf{x}_k)$. We fit $\Theta(\cdot)$ function to approximate the value of $f( \mathbf{x}_i ) - f(\mathbf{x}_k)$. Note that in some decision problems, the attribute weights $w_j,j=1,\cdots,n$ in Eq.(\ref{eq-global-inter-1}) are normalized to $[0,1]$ and $\sum\nolimits_{j=1}^{n} {w_j}=1$, which are useful for interpreting the trade-offs between attributes\footnote{Note that the trade-off between attributes is similar to attribute importance, but the trade-off emphasizes that assigning more weight to an attribute would decrease other attributes. That usually leads to a situation where some attributes have almost no effects on the predictions, which is unexpected because the selected attributes are often summarized based on DM's prior knowledge and their requirements. In this regard, we tend to train our model without normalization but provide normalized weights to evaluate the trade-offs between attributes \citep{liu2019preference}. Moreover, there are few minor differences on performances when using normalized weights or not.}. To address this issue, we apply the following transformation:
\begin{itemize}
\item For each attribute $g_j,j=1,\cdots,n$, the normalized weight is $w_j'=\frac{w_j}{ \sum\nolimits_{j=1}^{n} {w_j}}$.
\item The new global score is $U'(\mathbf{x}_i)=\alpha \sum\nolimits_{j=1}^{n} {w_j' v_j(x_{i,j})} + \frac{(1-\alpha)}{ \sum\nolimits_{j=1}^{n} {w_j}} f(x_{i,1},x_{i,2},\cdots,x_{i,n})$. Moreover, the ordinal relations among all attribute vectors are preserved since $ U'(\mathbf{x}_i) - U'(\mathbf{x}_k) =\frac{ U(\mathbf{x}_i) - U(\mathbf{x}_k)}{\sum\nolimits_{j=1}^{n} {w_j}}$ and $U(\mathbf{x}_i)\geq U(\mathbf{x}_k) \Leftrightarrow U'(\mathbf{x}_i)\geq U'(\mathbf{x}_k)$.
\end{itemize}

Given the input data $\mathcal{D} = \{ ( \Phi(\mathbf{x}_i, \mathbf{x}_k),y_{i,k} ) \}$, instead of mathematical programming, we can now use the machine learning scheme in section \ref{subsec-nnmcda} to infer the preference model and rank other attribute vectors. The output $\hat{y}_{i,k} = \sigma(U(\mathbf{x}_i) - U(\mathbf{x}_k))$ is the probability that $\mathbf{x}_i$ is at least as good as $\mathbf{x}_k$. We can pre-define two thresholds $\eta^1$ and $\eta^2$, where $0\leq \eta^1 \leq \eta^2 \leq 1$. If $\eta^2 \leq \hat{y}_{i,k}$, then $\mathbf{x}_i \succ \mathbf{x}_k$, and if $\eta^1 \leq \hat{y}_{i,k} \leq \eta^2$, then $\mathbf{x}_i \sim \mathbf{x}_k$ and otherwise, $\mathbf{x}_k \succ \mathbf{x}_i$. If we use the normalized weights, since the probability $\hat{y}_{i,k}' = \sigma (U'(\mathbf{x}_i) - U'(\mathbf{x}_k))$ is transformed nonlinearly, the pre-defined thresholds should also be transformed as follows, $\eta^1_{i,k}=\frac{\hat{y}_{i,k}'}{y_{i,k}} \eta^1, \eta^2_{i,k}=\frac{\hat{y}_{i,k}'}{y_{i,k}} \eta^2 $, to preserve the ordinal relations. In this way, the traditional multiple criteria ranking approaches can handle larger datasets and obtain smoother and more flexible marginal value functions to assist the DM. We present the simulation results in Section \ref{subsec-sim} and the results using real datasets in Section \ref{subsec-real}.

\subsection{Usefulness of the proposed framework in decision making.}
\label{subsec-usefulness}

As we introduce MCDA into machine learning, the main objective is shifted from achieving the best predictive performance to facilitating the DM in gaining insights into the characteristics of the decision making process and the interpretations of the results \citep{doumpos2011preference}. Once the marginal value functions are obtained by the proposed NN-MCDA framework, we can further analyze the DM's preference from the following perspectives.

\begin{figure}[htbp]
\centering
\includegraphics[scale=0.3]{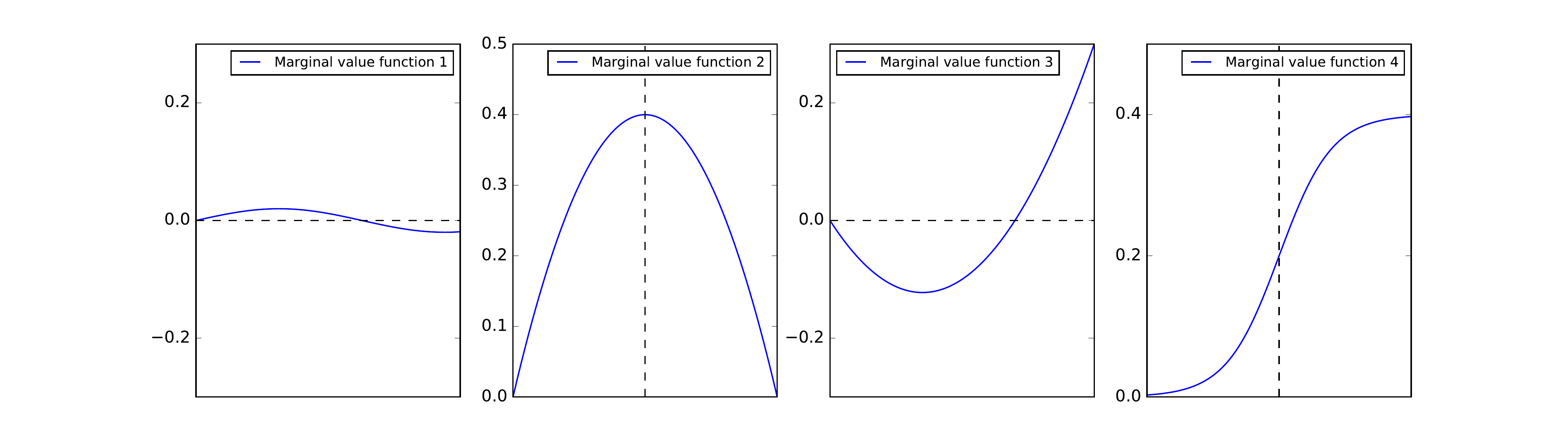}
\caption{Exemplary marginal value functions.\label{fig-exammvf}}
\end{figure}

First, the attribute importance usually has a long-tail distribution, with a few of them being very important and the majority of them being less important \citep{caruana2015intelligible}. The characteristics of the marginal value functions can reveal the importance of the corresponding attribute. If a marginal value function is close to 0 for the whole scale of the attribute values, it indicates that the attribute is either not important to the DM or the characteristic of the marginal value function is wrongly captured, because the change of this attribute has little influence on the predictions. When this is the case, we need to interact with the DM to determine whether we preserve this attribute or calibrate the model. In this regard, the proposed framework can perform model selection and modification (similar to statistical approaches like LASSO). For example, while predicting if a patient has the flu, the marginal value function of ``room humidity'' is in a shape like the marginal value function 1 in Figure \ref{fig-exammvf}, it is possible that ``room humidity'' has little contribution to the flu. However, whether we abandon it should be determined by a physician.

Second, the increasing and decreasing tendencies of the marginal value function curves reveal the change of the DM's non-monotonic preference. We focus on the monotonicity inflexion points because they can determine that to what attribute values, the DM is more sensitive. Moreover, if we partition the marginal value function curve by these points, we can discretize the continuous attribute into smaller ranges in which the DM's preference is monotonic. Such smaller intervals are useful for personalization (e.g. customer segmentation) and strategy-making tasks in management. For example, while evaluating the company's performance, if the marginal value function of ``cash to total assets ratio'' is like the second function in Figure \ref{fig-exammvf}, we can learn that a company with a very small or large ``cash to total assets ratio'' is in a bad condition. Companies with a large ratio are suggested to use the cash to do more investigations, whereas companies with a small ratio are suggested to save general expenses so that more cash can be used in new investigations.

Third, since the marginal value function returns a ``score'' that is added to the global value, it is crucial to determine whether the attribute positively or negatively contributes to the outcome. If a marginal value function is above/below \textit{zero}, the corresponding attribute is positively/negatively associated with the prediction. The marginal value function can capture the sign change (if any) of an attribute's contribution and provide the DM an exact attribute value where the sign changes. This is more informative than the statistical models that only provide a fixed coefficient representing either positive or negative effect of the attribute. For example, when predicting the risk of depression among adults, the marginal value function of ``age'' may has a shape similar to the third function in Figure \ref{fig-exammvf} (please also refer to Figure \ref{fig-depre-nnmcda}, which is drawn from the real-data). The shape of this curve indicates that the risk of depression does not increase while aging if the adult is younger than a threshold. The risk will increase once the adult is older than that threshold (the threshold is 71.58 in the real data introduced in section \ref{subsec-preddeprisk}). Statistical models, on the other hand, can only conclude that age has either a negative or positive effect on the depression risk. We need to segment the adults to pre-defined age groups to capture such sign change effect.

Fourth, the concavity (and convexity) of the marginal value function can directly reflect the changing rate of the DM's preference. Such information is important to both economics and marketing problems. For example, if the consumer's preference to ``discount rate'' is in a same shape of marginal value function 4 in Figure \ref{fig-exammvf}, it implies that at the beginning, along with the increase of the discount rate, the consumer's utility (propensity to consume the product) grows more quickly. However, when the discount rate is over a specific value, it gives a signal that the product is possibly of bad quality. Although the consumer's utility still grows, its rate of increase starts to slow down. This provides the DM with a conclusion that keeping the discount rate at a medium level could maximize the profit.
\section{Experiments.}
\label{sec-experiment}

To validate the proposed NN-MCDA model, we perform experiments with both synthetic and real datasets. We use \textit{area under the curve} (AUC) of \textit{receiver operating characteristic} (ROC) curve to measure the model performance. In subsection \ref{subsec-sim}, three simulation experiments examine (a) the influence of the degree of polynomial on the prediction performance, (b) the influence of the value of $\alpha$, the trade-off coefficient, on prediction performance, and (c) the goodness of the proposed NN-MCDA approach in fitting the given marginal value functions. In Section \ref{subsec-real}, we first apply the NN-MCDA model to a multiple criteria decision problem where we rank universities based on the employer reputation. Then we predict the risk for geriatric depression with useful interpretations of the risk factors with a higher resolution. At last, we perform the NN-MCDA on an open access data to compare the results with a published paper.

\subsection{Simulations.}
\label{subsec-sim}

For brevity, we set equal pre-defined degrees of polynomial for all marginal value functions in the subsequent experiments. We generate three typical synthetic datasets (from the simplest to very complex) as follows:
\begin{enumerate}
\item Uniformly draw $N$ attribute vectors with $n$ attributes whose values are within [0,1].
\item We generate three datasets. (a) For the first dataset $\mathcal{D}^n_l$, all $n$ attributes have equal importance and the actual marginal value functions are identity functions. The global score for each attribute vector is a linear summation of $n$ attribute values without any attribute interactions and an additional noise term that is in a standard normal distribution; (b) The second dataset $\mathcal{D}^n_{polynomial-3}$ randomly generates 3-degree polynomials marginal value functions for $n$ attributes, and the global score is the summation of marginal values, all $\binom{n}{2}$ attribute interactions and a standard normal noise term. (c) The third dataset $\mathcal{D}^n_{polynomial-15}$ is extremely complex. The global score is the summation of $n$ 15-degree polynomial marginal values, all possible attribute interactions (pairwise, triple-wise and higher interactions) and a standard normal noise term.
\item We compare global scores between each pair of attribute vectors. If $U(\mathbf{x}_i) - U(\mathbf{x}_k) \geq 0$, then $y_{i,k} = 1$, otherwise, $y_{i,k}=0$. Note that the actual input is the transformed attribute vector.
\end{enumerate}

\subsubsection{Experiment I: Relationship between degree of polynomial and model performance}
\label{subsubsec-exp1}

The first simulated experiment aims at exploring the relationship between the pre-defined degree of polynomial and AUC. The parameters used in the experiment is shown in Table \ref{tab-setExpI}. For each setting, we iteratively repeat the experiment for 10 times and record the averaged AUC. In this experiment, the numbers of iterations are determined using fivefold cross-validation: We partition the training set into five sets and set aside one of them as a validation set. We then train the model using the other four partitions and use the validation set to check the convergence. This procedure is repeated five times and the averaged number of iterations is used to train the final model with the whole dataset \citep{lou2012intelligible}.

\begin{table}[htbp]
\centering
\caption{Parameter settings for simulation experiment I. \label{tab-setExpI}}
\resizebox{0.95\textwidth}{!}{
\begin{tabular}{lllllll}
    \hline
          & N     & \# Comparisons & Training size & Pre-defined $D_j$ & \# Attributes \\
    \hline
    Setting & 250   & $\frac{250*249}{2}$ & 0.5, 0.6, 0.7, 0.8, 0.9 & 1, 2, 3, 5, 10 & 3, 5 \\
    \hline
    \end{tabular}%
    }
\end{table}

\begin{figure}[htbp]
\centering
{\includegraphics[scale=0.25]{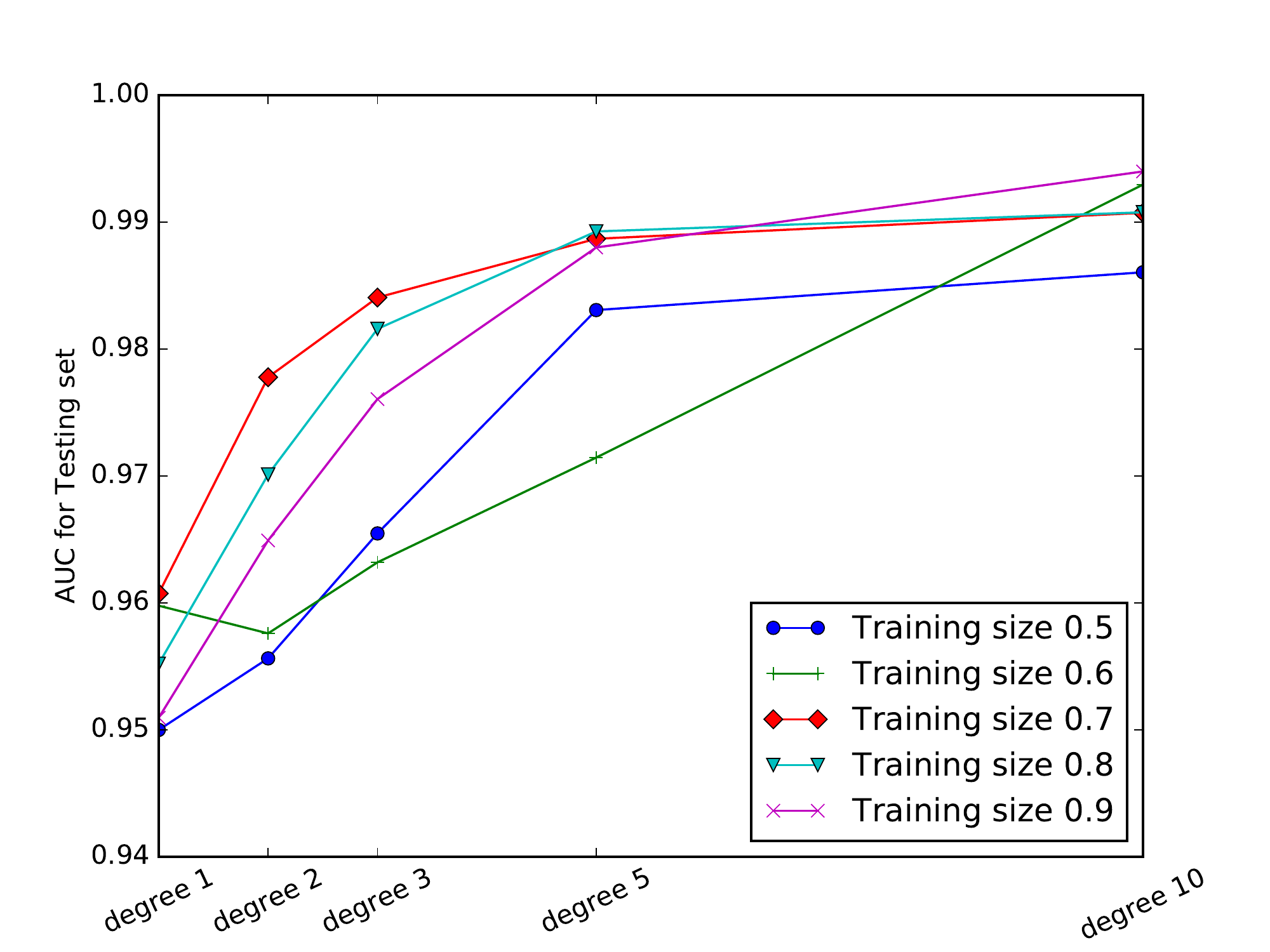}
\includegraphics[scale=0.25]{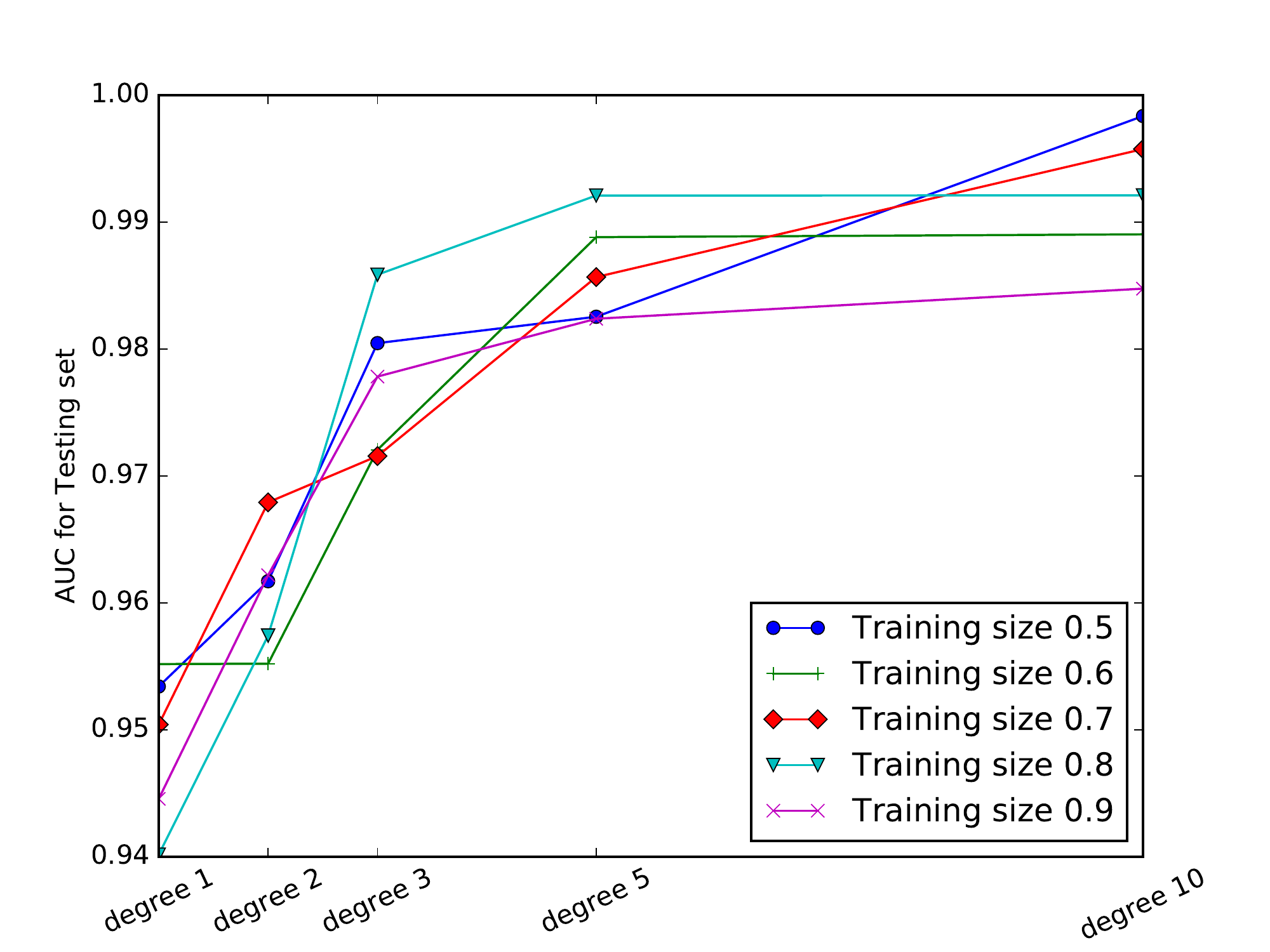}
}
\caption{The average AUC for the testing set using training set with different sizes from datasets $\mathcal{D}^3_l$ and $\mathcal{D}^5_l$.\label{fig-exp1testlinear}}
\end{figure}
\begin{figure}[htbp]
\centering
{\includegraphics[scale=0.25]{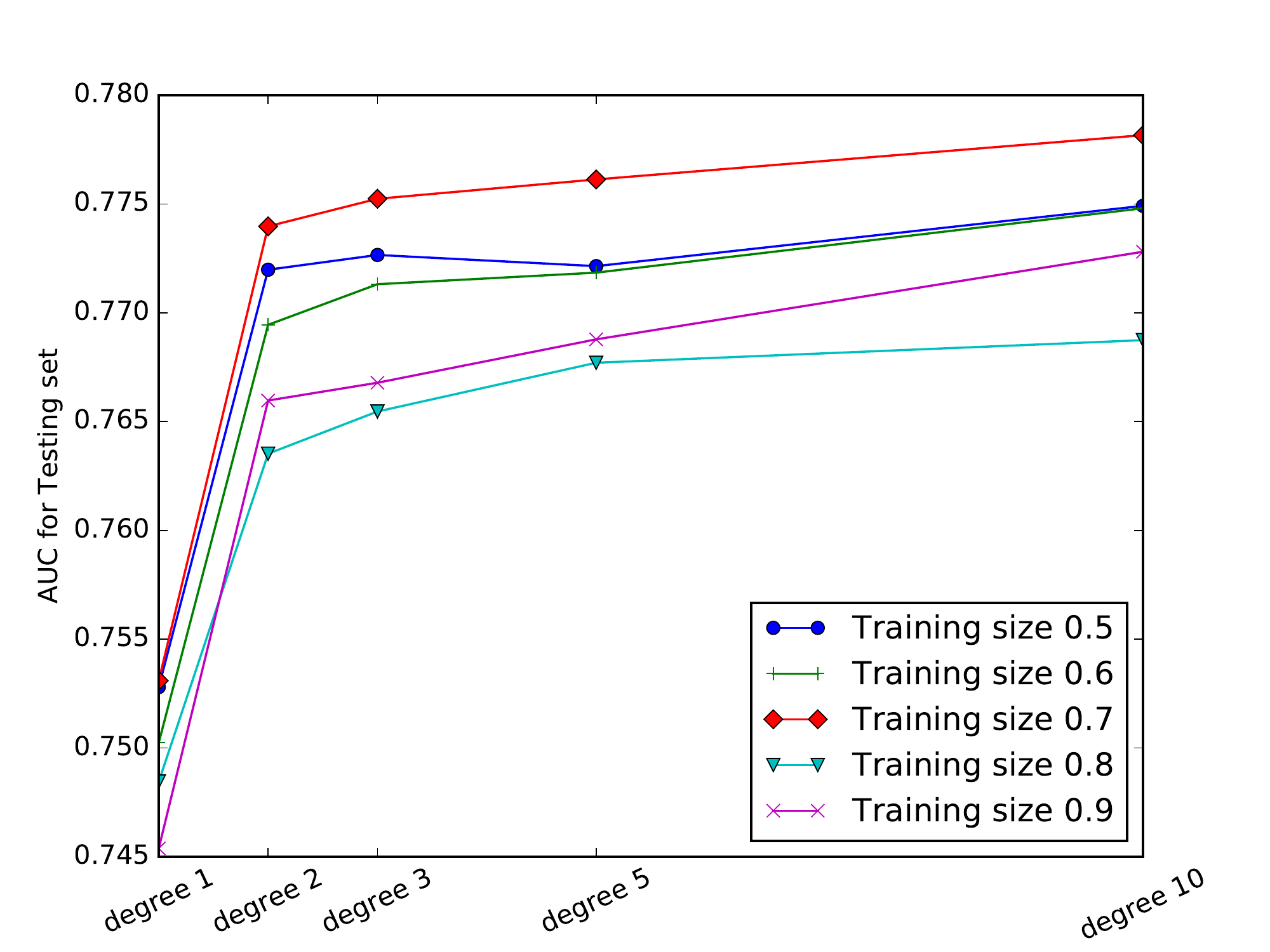}
\includegraphics[scale=0.25]{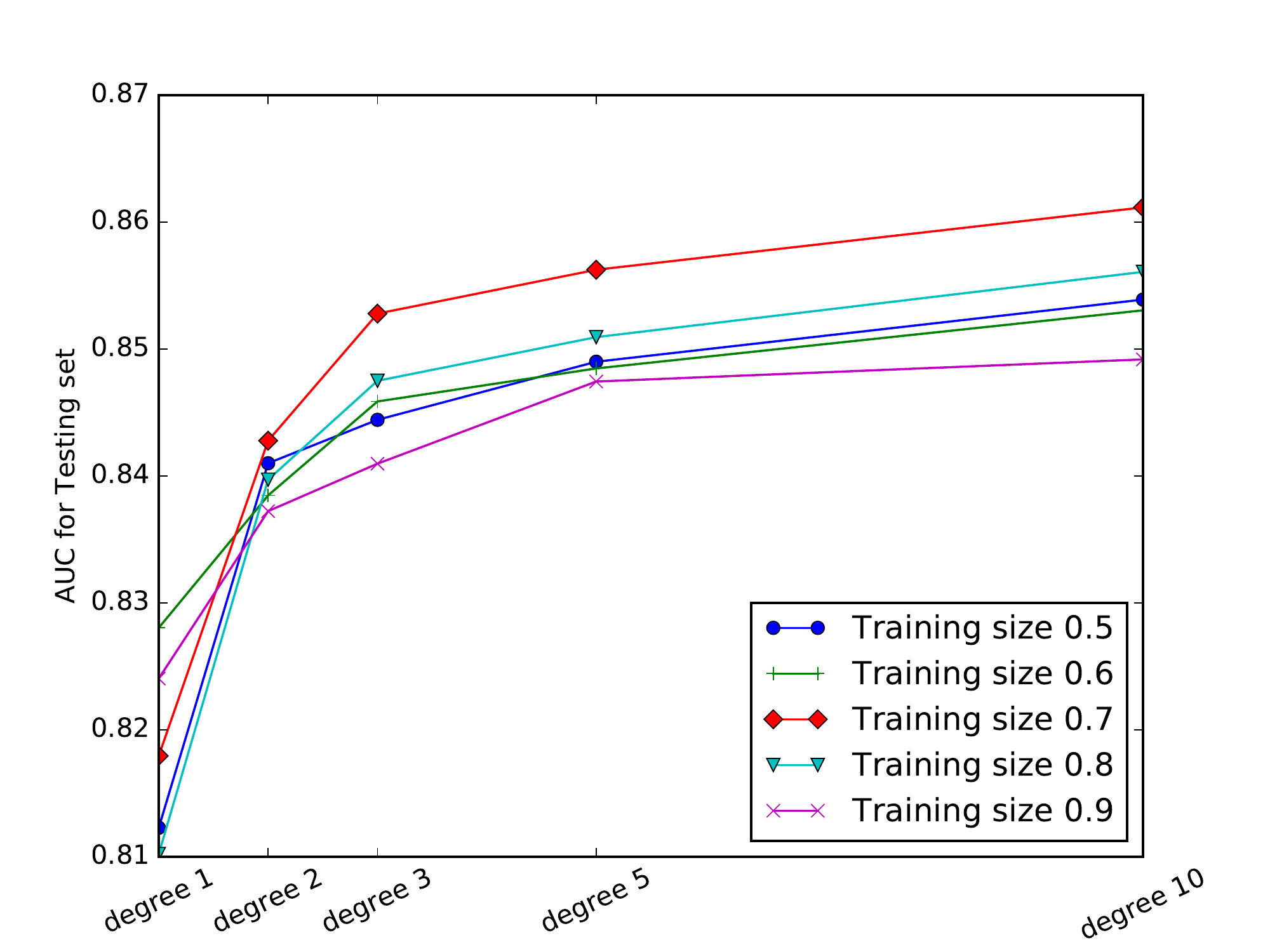}
}
\caption{The average AUC for the testing set using training set with different sizes from datasets $\mathcal{D}^3_{polynomial-3}$ and $\mathcal{D}^5_{polynomial-3}$.\label{fig-exp1testrandom}}
\end{figure}
\begin{figure}[htbp]
\centering
{\includegraphics[scale=0.25]{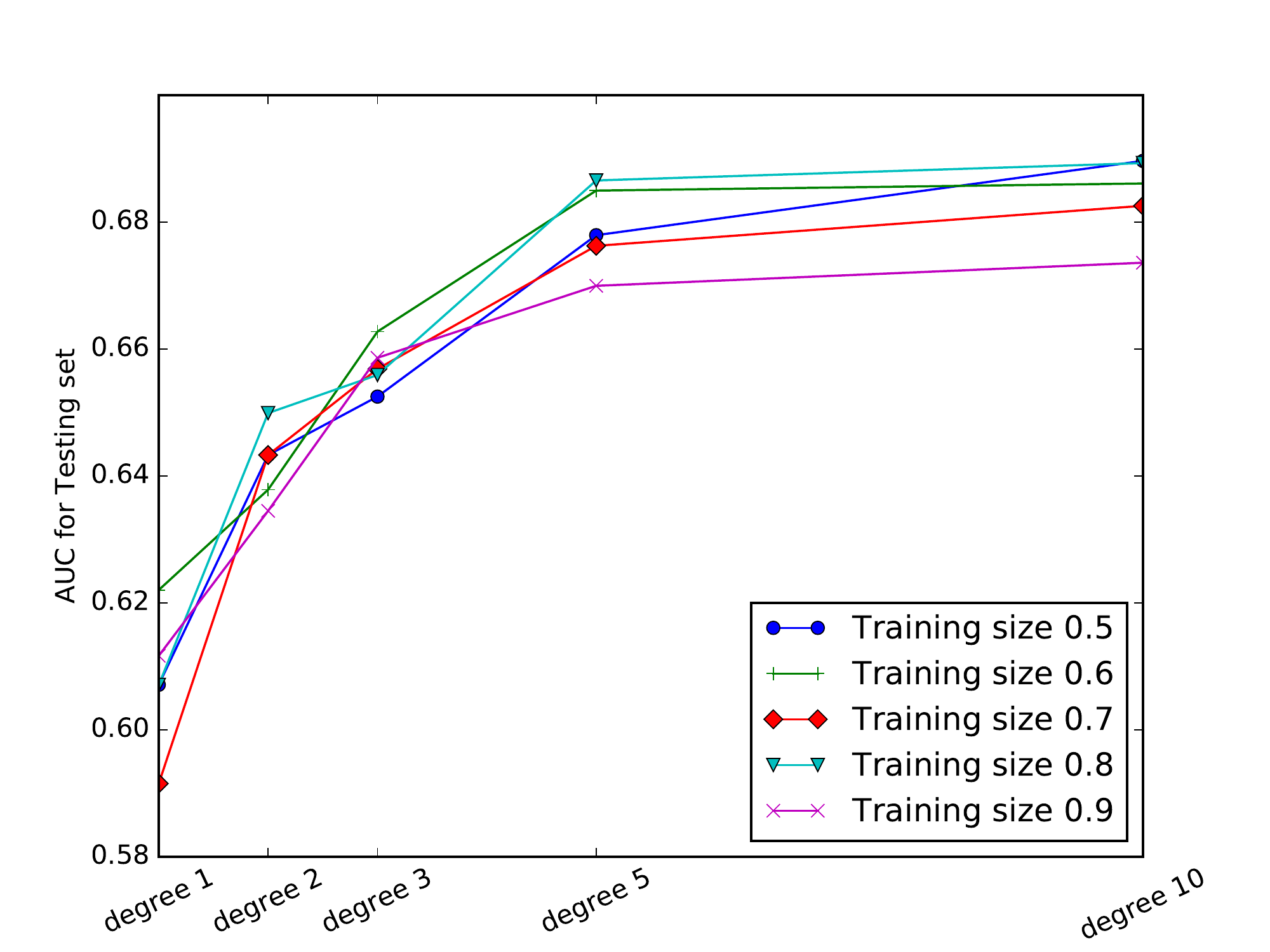}
\includegraphics[scale=0.25]{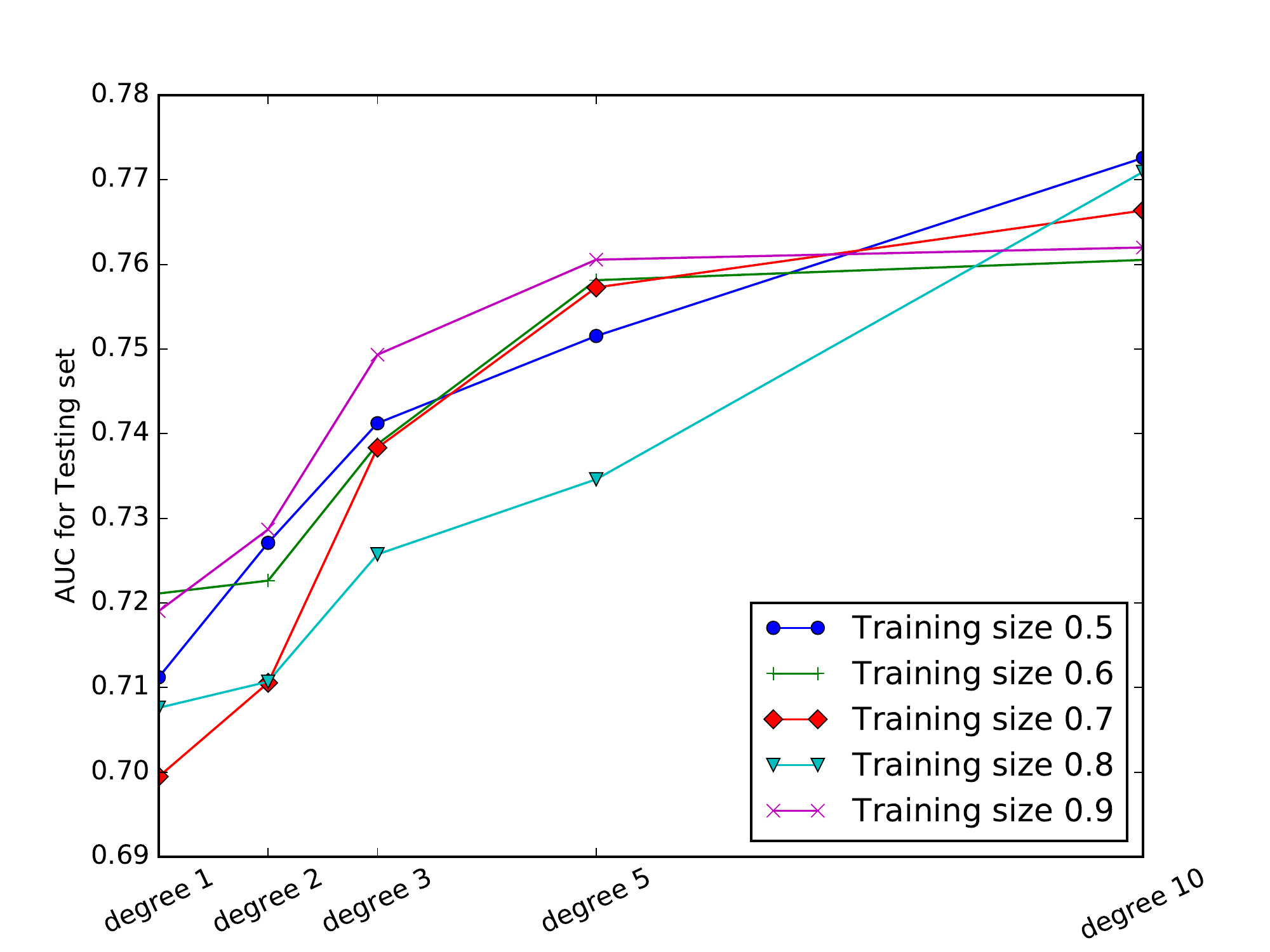}
}
\caption{average AUC for the testing set using training set with different sizes from datasets $\mathcal{D}^3_{polynomial-15}$ and $\mathcal{D}^5_{polynomial-15}$.\label{fig-exp1testrandom2}}
\end{figure}

\begin{figure}[h]
\centering
{\includegraphics[scale=0.3]{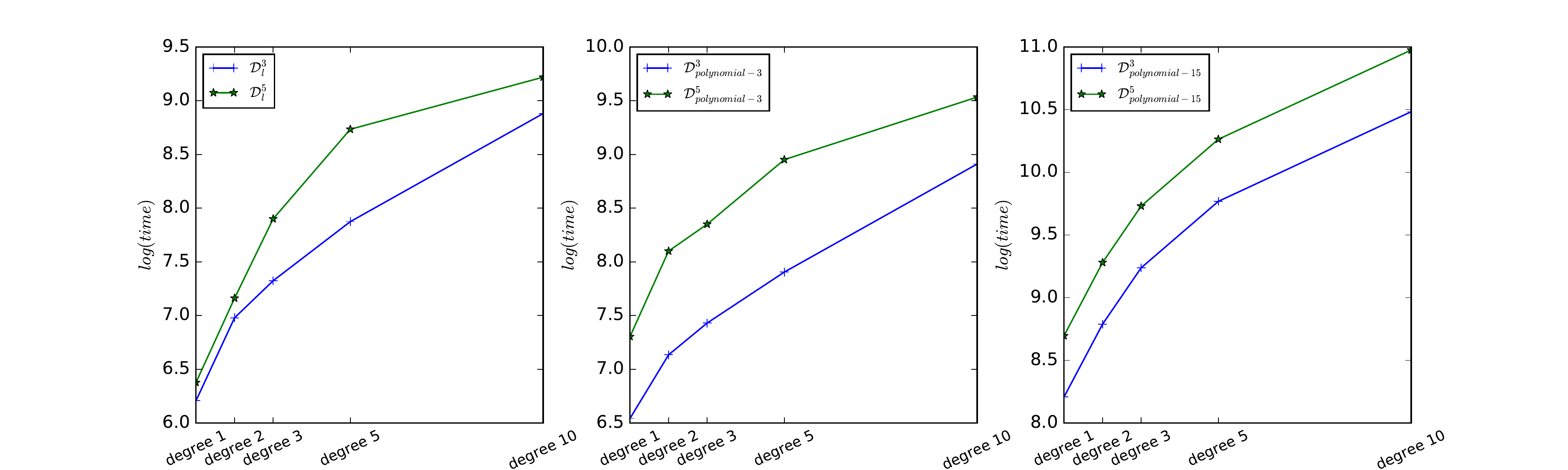}}
\caption{The average computational time for different datasets. The recoded computational times in the figure are the averaged time for training all five training sizes given a pre-defined degree of polynomials.\label{fig-exp1time}}
\end{figure}

Figures \ref{fig-exp1testlinear}, \ref{fig-exp1testrandom} and \ref{fig-exp1testrandom2} report the averaged AUC for the testing set with different training sizes using the three synthetic datasets. Though there are no obvious relationships between the training sizes and the model performance, we find two interesting patterns. First, higher pre-defined degrees of polynomials can lead to higher accuracy when convergence. That results from the ability of the underlying model to capture more complicated nonlinearity. However, higher degrees of polynomials usually require more iterations to converge. More specifically, we depict the averaged computational time for each training process in Figure \ref{fig-exp1time}. Apparently, while increasing the model complexity, for example, using higher degrees of polynomial marginal value function and considering more attributes, the average computational time to converge also increases almost linearly. Another interesting pattern is that the shape of the AUC curves (Figures \ref{fig-exp1testlinear}, \ref{fig-exp1testrandom} and \ref{fig-exp1testrandom2}) can fit a concave function in general. When the degree increases, the AUC improvement (over the model with the immediate smaller degree) is becoming smaller. For example, the improvement is more obvious if we change the pre-defined degree from 1 to 3 than that if we change from 3 to 5 and from 5 to 10. The improvement diminishes quickly along with the increase of pre-defined degree of polynomials. The greatest AUC improvement happens if we increase the degree to 3, while the improvement resulted from further increasing the degree to 5 and 10 is slim. The results suggest that it is not necessary to set a very large $D_j$ for the seek of minor improvement because the computational cost increases much faster when $D_j$ increases. Generally, we believe that a polynomial of 3 degree is sufficient to capture the characteristics for all the three datasets. Higher degrees of polynomials have a risk for over-fitting and obviously cost more computational time, but contribute little to accuracy.

\begin{table}[htbp]
  \centering
  \caption{The average AUCs $\pm$ one deviation on dataset $\mathcal{D}^3_{l}$ and $\mathcal{D}^5_{l}$ using different machine learning algorithms.}
\resizebox{1.0\textwidth}{!}{    
    \begin{tabular}{lllllllllll}
    \hline
          & \multicolumn{5}{c}{n=3}               & \multicolumn{5}{c}{n=5} \\
    \hline
    Training size & \multicolumn{1}{c}{0.5} & \multicolumn{1}{c}{0.6} & \multicolumn{1}{c}{0.7} & \multicolumn{1}{c}{0.8} & \multicolumn{1}{c}{0.9} & \multicolumn{1}{c}{0.5} & \multicolumn{1}{c}{0.6} & \multicolumn{1}{c}{0.7} & \multicolumn{1}{c}{0.8} & \multicolumn{1}{c}{0.9} \\
    \hline
    NN-MCDA-D1 & 0.949$\pm$0.023 & 0.959$\pm$0.021 & 0.960$\pm$0.011 & 0.955$\pm$0.009 & 0.951$\pm$0.005 & 0.953$\pm$0.021 & 0.955$\pm$0.014 & 0.951$\pm$0.019 & 0.941$\pm$0.013 & 0.945$\pm$0.012 \\
    \hline
    NN-MCDA-D2 & 0.956$\pm$0.022 & 0.958$\pm$0.027 & 0.978$\pm$0.019 & 0.971$\pm$0.007 & 0.965$\pm$0.003 & 0.962$\pm$0.018 & 0.955$\pm$0.021 & 0.968$\pm$0.009 & 0.957$\pm$0.022 & 0.962$\pm$0.018 \\
    \hline
    NN-MCDA-D3 & 0.965$\pm$0.031 & 0.963$\pm$0.024 & 0.984$\pm$0.019 & 0.981$\pm$0.004 & 0.976$\pm$0.003 & 0.980$\pm$0.019 & 0.972$\pm$0.019 & 0.972$\pm$0.021 & 0.986$\pm$0.013 & 0.978$\pm$0.010 \\
    \hline
    NN-MCDA-D5 & 0.983$\pm$0.026 & 0.971$\pm$0.020 & 0.988$\pm$0.019 & 0.989$\pm$0.005 & 0.988$\pm$0.001 & 0.983$\pm$0.028 & 0.989$\pm$0.018 & 0.876$\pm$0.037 & 0.992$\pm$0.009 & 0.982$\pm$0.010 \\
    \hline
    NN-MCDA-D10 & 0.986$\pm$0.011 & 0.993$\pm$0.010 & 0.991$\pm$0.009 & 0.991$\pm$0.002 & 0.994$\pm$0.001 & 0.998$\pm$0.013 & 0.989$\pm$0.029 & 0.996$\pm$0.009 & 0.992$\pm$0.007 & 0.985$\pm$0.008 \\
    \hline
    MLP-D1 & 0.969$\pm$0.001 & 0.951$\pm$0.005 & 0.960$\pm$0.000 & 0.957$\pm$0.000 & 0.998$\pm$0.001 & 0.970$\pm$0.001 & 0.969$\pm$0.002 & 0.977$\pm$0.001 & 0.981$\pm$0.001 & 0.966$\pm$0.000 \\
    \hline
    MLP-D2 & 0.970$\pm$0.002 & 0.969$\pm$0.003 & 0.973$\pm$0.003 & 0.970$\pm$0.001 & 0.999$\pm$0.002 & 0.979$\pm$0.001 & 0.980$\pm$0.000 & 0.989$\pm$0.001 & 0.988$\pm$0.001 & 0.988$\pm$0.001 \\
    \hline
    MLP-D3 & 0.992$\pm$0.001 & 0.990$\pm$0.001 & 0.987$\pm$0.002 & 0.989$\pm$0.001 & 0.973$\pm$0.001 & 0.987$\pm$0.002 & 0.991$\pm$0.001 & 0.991$\pm$0.001 & 0.995$\pm$0.000 & 0.996$\pm$0.000 \\
    \hline
    MLP-D5 & 0.999$\pm$0.001 & 0.993$\pm$0.000 & 0.990$\pm$0.001 & 0.998$\pm$0.000 & 0.999$\pm$0.001 & 0.995$\pm$0.001 & 0.996$\pm$0.001 & 0.995$\pm$0.002 & 0.996$\pm$0.000 & 0.996$\pm$0.000 \\
    \hline
    MLP-D10 & 0.999$\pm$0.002 & 0.997$\pm$0.001 & 0.995$\pm$0.000 & 0.999$\pm$0.001 & 0.999$\pm$0.001 & 0.999$\pm$0.001 & 0.999$\pm$0.001 & 0.998$\pm$0.001 & 0.998$\pm$0.001 & 0.999$\pm$0.000 \\
    \hline
    SVM-Linear & 0.998$\pm$0.001 & 0.998$\pm$0.001 & 0.998$\pm$0.001 & 0.998$\pm$0.000 & 0.998$\pm$0.000 & 0.997$\pm$0.002 & 0.998$\pm$0.003 & 0.997$\pm$0.001 & 0.997$\pm$0.000 & 0.997$\pm$0.000 \\
    \hline
    SVM-RBF & 0.998$\pm$0.001 & 0.997$\pm$0.000 & 0.998$\pm$0.001 & 0.998$\pm$0.000 & 0.999$\pm$0.001 & 0.995$\pm$0.001 & 0.996$\pm$0.000 & 0.996$\pm$0.003 & 0.996$\pm$0.001 & 0.996$\pm$0.001 \\
    \hline
    SVM-D3poly & 0.967$\pm$0.002 & 0.969$\pm$0.001 & 0.962$\pm$0.000 & 0.971$\pm$0.001 & 0.973$\pm$0.000 & 0.987$\pm$0.003 & 0.991$\pm$0.000 & 0.986$\pm$0.001 & 0.988$\pm$0.002 & 0.989$\pm$0.001 \\
    \hline
    GAM-D3 & 0.983$\pm$0.011 & 0.983$\pm$0.017 & 0.983$\pm$0.009 & 0.982$\pm$0.003 & 0.985$\pm$0.002 & 0.986$\pm$0.010 & 0.985$\pm$0.011 & 0.984$\pm$0.010 & 0.986$\pm$0.009 & 0.984$\pm$0.010 \\
    \hline
    GAM-D10 & 0.985$\pm$0.010 & 0.984$\pm$0.018 & 0.984$\pm$0.010 & 0.983$\pm$0.002 & 0.985$\pm$0.001 & 0.987$\pm$0.008 & 0.985$\pm$0.009 & 0.985$\pm$0.011 & 0.986$\pm$0.010 & 0.986$\pm$0.011 \\
    \hline
    DeciTr-MaxDep6 & 0.842$\pm$0.001 & 0.842$\pm$0.003 & 0.843$\pm$0.001 & 0.835$\pm$0.001 & 0.831$\pm$0.000 & 0.931$\pm$0.001 & 0.927$\pm$0.000 & 0.937$\pm$0.000 & 0.933$\pm$0.000 & 0.938$\pm$0.000 \\
    \hline
    DeciTr-MaxDep10 & 0.896$\pm$0.002 & 0.893$\pm$0.000 & 0.890$\pm$0.002 & 0.897$\pm$0.001 & 0.898$\pm$0.000 & 0.965$\pm$0.001 & 0.965$\pm$0.000 & 0.971$\pm$0.001 & 0.971$\pm$0.000 & 0.967$\pm$0.000 \\
    \hline
    DeciTr-MaxDep20 & 0.900$\pm$0.003 & 0.898$\pm$0.001 & 0.895$\pm$0.000 & 0.908$\pm$0.000 & 0.903$\pm$0.000 & 0.966$\pm$0.001 & 0.966$\pm$0.001 & 0.972$\pm$0.001 & 0.974$\pm$0.001 & 0.970$\pm$0.000 \\
    \hline
    PLR-D1 & 0.899$\pm$0.014 & 0.903$\pm$0.022 & 0.901$\pm$0.021 & 0.900$\pm$0.011 & 0.901$\pm$0.000 & 0.903$\pm$0.011 & 0.902$\pm$0.019 & 0.900$\pm$0.016 & 0.910$\pm$0.018 & 0.911$\pm$0.009 \\
    \hline
    PLR-D2 & 0.910$\pm$0.017 & 0.913$\pm$0.023 & 0.928$\pm$0.012 & 0.923$\pm$0.016 & 0.922$\pm$0.001 & 0.923$\pm$0.017 & 0.931$\pm$0.011 & 0.929$\pm$0.019 & 0.931$\pm$0.011 & 0.936$\pm$0.016 \\
    \hline
    PLR-D3 & 0.945$\pm$0.020 & 0.933$\pm$0.017 & 0.945$\pm$0.014 & 0.941$\pm$0.004 & 0.914$\pm$0.001 & 0.941$\pm$0.010 & 0.952$\pm$0.009 & 0.946$\pm$0.014 & 0.956$\pm$0.012 & 0.944$\pm$0.010 \\
    \hline
    PLR-D5 & 0.950$\pm$0.018 & 0.941$\pm$0.023 & 0.955$\pm$0.020 & 0.949$\pm$0.003 & 0.925$\pm$0.000 & 0.958$\pm$0.011 & 0.959$\pm$0.012 & 0.961$\pm$0.012 & 0.964$\pm$0.010 & 0.961$\pm$0.011 \\
    \hline
    PLR-D10 & 0.959$\pm$0.011 & 0.957$\pm$0.009 & 0.960$\pm$0.008 & 0.955$\pm$0.010 & 0.933$\pm$0.000 & 0.963$\pm$0.013 & 0.970$\pm$0.023 & 0.972$\pm$0.014 & 0.977$\pm$0.009 & 0.973$\pm$0.009 \\
    \hline
    Mean & 0.957$\pm$0.010 & 0.955$\pm$0.011 & 0.955$\pm$0.008& 0.958$\pm$0.004 &0.957$\pm$0.001	& 0.970$\pm$0.011 &0.958$\pm$0.009 & 0.957	$\pm$0.009&	0.974$\pm$0.007	&0.972$\pm$0.006 \\
    \hline
    \end{tabular}}
  \label{tab-simexp1data1}
\end{table}%

\begin{table}[htbp]
  \centering
  \caption{The average AUCs $\pm$ one deviation on dataset $\mathcal{D}^3_{polynomial-3}$ and $\mathcal{D}^5_{polynomial-3}$ using different machine learning algorithms.}
    \resizebox{1.0\textwidth}{!}{
    \begin{tabular}{lllllllllll}
    \hline
          & \multicolumn{5}{c}{n=3}               & \multicolumn{5}{c}{n=5} \\
    \hline
    Training size & \multicolumn{1}{c}{0.5} & \multicolumn{1}{c}{0.6} & \multicolumn{1}{c}{0.7} & \multicolumn{1}{c}{0.8} & \multicolumn{1}{c}{0.9} & \multicolumn{1}{c}{0.5} & \multicolumn{1}{c}{0.6} & \multicolumn{1}{c}{0.7} & \multicolumn{1}{c}{0.8} & \multicolumn{1}{c}{0.9} \\
    \hline
    NN-MCDA-D1 & 0.753$\pm$0.018 & 0.750$\pm$0.023 & 0.753$\pm$0.021 & 0.748$\pm$0.010 & 0.745$\pm$0.008 & 0.812$\pm$0.021 & 0.828$\pm$0.038 & 0.818$\pm$0.029 & 0.810$\pm$0.021 & 0.824$\pm$0.015 \\
    \hline
    NN-MCDA-D2 & 0.772$\pm$0.024 & 0.769$\pm$0.021 & 0.774$\pm$0.012 & 0.764$\pm$0.008 & 0.766$\pm$0.009 & 0.841$\pm$0.023 & 0.838$\pm$0.033 & 0.843$\pm$0.021 & 0.840$\pm$0.019 & 0.837$\pm$0.021 \\
    \hline
    NN-MCDA-D3 & 0.773$\pm$0.026 & 0.771$\pm$0.020 & 0.775$\pm$0.021 & 0.765$\pm$0.011 & 0.767$\pm$0.011 & 0.844$\pm$0.031 & 0.846$\pm$0.052 & 0.853$\pm$0.041 & 0.847$\pm$0.037 & 0.841$\pm$0.031 \\
    \hline
    NN-MCDA-D5 & 0.772$\pm$0.018 & 0.772$\pm$0.019 & 0.776$\pm$0.015 & 0.768$\pm$0.011 & 0.769$\pm$0.008 & 0.849$\pm$0.029 & 0.853$\pm$0.027 & 0.856$\pm$0.039 & 0.851$\pm$0.015 & 0.847$\pm$0.011 \\
    \hline
    NN-MCDA-D10 & 0.775$\pm$0.020 & 0.775$\pm$0.021 & 0.778$\pm$0.012 & 0.769$\pm$0.010 & 0.773$\pm$0.006 & 0.854$\pm$0.032 & 0.853$\pm$0.052 & 0.861$\pm$0.025 & 0.856$\pm$0.030 & 0.849$\pm$0.028 \\
    \hline
    MLP-D1 & 0.787$\pm$0.010 & 0.793$\pm$0.002 & 0.790$\pm$0.001 & 0.791$\pm$0.003 & 0.800$\pm$0.004 & 0.837$\pm$0.002 & 0.835$\pm$0.001 & 0.831$\pm$0.002 & 0.836$\pm$0.001 & 0.833$\pm$0.011 \\
    \hline
    MLP-D2 & 0.790$\pm$0.005 & 0.793$\pm$0.001 & 0.792$\pm$0.003 & 0.797$\pm$0.002 & 0.802$\pm$0.008 & 0.848$\pm$0.001 & 0.851$\pm$0.003 & 0.845$\pm$0.004 & 0.849$\pm$0.010 & 0.847$\pm$0.008 \\
    \hline
    MLP-D3 & 0.791$\pm$0.002 & 0.799$\pm$0.002 & 0.799$\pm$0.001 & 0.801$\pm$0.002 & 0.808$\pm$0.002 & 0.855$\pm$0.004 & 0.855$\pm$0.001 & 0.851$\pm$0.000 & 0.855$\pm$0.000 & 0.857$\pm$0.000 \\
    \hline
    MLP-D5 & 0.793$\pm$0.001 & 0.806$\pm$0.003 & 0.801$\pm$0.002 & 0.805$\pm$0.001 & 0.810$\pm$0.006 & 0.861$\pm$0.002 & 0.856$\pm$0.001 & 0.854$\pm$0.003 & 0.861$\pm$0.001 & 0.860$\pm$0.001 \\
    \hline
    MLP-D10 & 0.795$\pm$0.001 & 0.810$\pm$0.002 & 0.804$\pm$0.004 & 0.810$\pm$0.003 & 0.811$\pm$0.009 & 0.866$\pm$0.004 & 0.864$\pm$0.001 & 0.865$\pm$0.000 & 0.862$\pm$0.001 & 0.862$\pm$0.000 \\
    \hline
    SVM-Linear & 0.687$\pm$0.004 & 0.683$\pm$0.004 & 0.686$\pm$0.004 & 0.681$\pm$0.006 & 0.678$\pm$0.001 & 0.748$\pm$0.003 & 0.748$\pm$0.003 & 0.755$\pm$0.003 & 0.748$\pm$0.007 & 0.743$\pm$0.007 \\
    \hline
    SVM-RBF & 0.688$\pm$0.003 & 0.685$\pm$0.003 & 0.686$\pm$0.007 & 0.682$\pm$0.005 & 0.680$\pm$0.000 & 0.748$\pm$0.003 & 0.748$\pm$0.002 & 0.755$\pm$0.003 & 0.748$\pm$0.004 &  0.743$\pm$0.006 \\
    \hline
    SVM-D3poly & 0.682$\pm$0.004 & 0.680$\pm$0.003 & 0.687$\pm$0.006 & 0.682$\pm$0.005 & 0.676$\pm$0.000 & 0.750$\pm$0.002 & 0.749$\pm$0.003 & 0.756$\pm$0.005 & 0.752$\pm$0.004 & 0.749$\pm$0.008 \\
    \hline
    GAM-D3 & 0.687$\pm$0.002 & 0.684$\pm$0.001 & 0.688$\pm$0.002 & 0.681$\pm$0.001 & 0.681$\pm$0.001 & 0.749$\pm$0.011 & 0.746$\pm$0.006 & 0.752$\pm$0.008 & 0.750$\pm$0.007 & 0.742$\pm$0.003 \\
    \hline
    GAM-D10 & 0.687$\pm$0.004 & 0.684$\pm$0.006 & 0.687$\pm$0.009 & 0.681$\pm$0.001 & 0.680$\pm$0.002 & 0.749$\pm$0.010 & 0.746$\pm$0.010 & 0.752$\pm$0.012 & 0.749$\pm$0.011 & 0.741$\pm$0.009 \\
    \hline
    DeciTr-MaxDep6 & 0.685$\pm$0.003 & 0.678$\pm$0.003 & 0.682$\pm$0.004 & 0.679$\pm$0.007 & 0.676$\pm$0.006 & 0.724$\pm$0.003 & 0.725$\pm$0.002 & 0.734$\pm$0.005 & 0.736$\pm$0.006 & 0.725$\pm$0.009 \\
    \hline
    DeciTr-MaxDep10 & 0.672$\pm$0.002 & 0.669$\pm$0.004 & 0.672$\pm$0.003 & 0.663$\pm$0.007 & 0.665$\pm$0.006 & 0.717$\pm$0.002 & 0.719$\pm$0.002 & 0.721$\pm$0.004 & 0.722$\pm$0.002 & 0.722$\pm$0.008 \\
    \hline
    DeciTr-MaxDep20 & 0.617$\pm$0.003 & 0.623$\pm$0.002 & 0.620$\pm$0.005 & 0.621$\pm$0.009 & 0.630$\pm$0.008 & 0.666$\pm$0.003 & 0.663$\pm$0.004 & 0.672$\pm$0.003 & 0.671$\pm$0.005 & 0.665$\pm$0.009 \\
    \hline
    PLR-D1 & 0.604$\pm$0.011 & 0.609$\pm$0.021 & 0.611$\pm$0.014 & 0.603$\pm$0.011 & 0.610$\pm$0.003 & 0.703$\pm$0.039 & 0.698$\pm$0.010 & 0.700$\pm$0.019 & 0.702$\pm$0.010 & 0.708$\pm$0.017 \\
    \hline
    PLR-D2 & 0.621$\pm$0.013 & 0.632$\pm$0.018 & 0.639$\pm$0.013 & 0.625$\pm$0.010 & 0.621$\pm$0.010 & 0.712$\pm$0.010 & 0.702$\pm$0.018 & 0.710$\pm$0.010 & 0.711$\pm$0.015 & 0.713$\pm$0.029 \\
    \hline
    PLR-D3 & 0.638$\pm$0.021 & 0.644$\pm$0.020 & 0.640$\pm$0.009 & 0.637$\pm$0.009 & 0.635$\pm$0.010 & 0.730$\pm$0.011 & 0.729$\pm$0.021 & 0.721$\pm$0.014 & 0.729$\pm$0.029 & 0.720$\pm$0.018 \\
    \hline
    PLR-D5 & 0.651$\pm$0.017 & 0.657$\pm$0.017 & 0.649$\pm$0.019 & 0.650$\pm$0.011 & 0.650$\pm$0.011 & 0.733$\pm$0.011 & 0.735$\pm$0.013 & 0.729$\pm$0.020 & 0.734$\pm$0.016 & 0.736$\pm$0.019 \\
    \hline
    PLR-D10 & 0.669$\pm$0.009 & 0.667$\pm$0.007 & 0.670$\pm$0.011 & 0.669$\pm$0.011 & 0.668$\pm$0.008 & 0.739$\pm$0.021 & 0.739$\pm$0.023 & 0.732$\pm$0.009 & 0.741$\pm$0.009 & 0.739$\pm$0.014 \\
    \hline
    Mean & 0.713$\pm$0.010	&0.708$\pm$0.010	&0.713$\pm$0.009&	0.712$\pm$	0.006	&0.713$\pm$0.006	&0.780$\pm$0.009&	0.779$\pm$0.014&	0.781$\pm$0.012	&0.775$\pm$0.009	&0.778$\pm$0.012\\
    \hline
    \end{tabular}}
  \label{tab-simexp1data2}
\end{table}%

\begin{table}[htbp]
  \centering
  \caption{The average AUCs $\pm$ one deviation on dataset $\mathcal{D}^3_{polynomial-15}$ and $\mathcal{D}^5_{polynomial-15}$ using different machine learning algorithms.}
    \resizebox{1.0\textwidth}{!}{
    \begin{tabular}{lllllllllll}
    \hline
          & \multicolumn{5}{c}{n=3}               & \multicolumn{5}{c}{n=5} \\
    \hline
    Training size & \multicolumn{1}{c}{0.5} & \multicolumn{1}{c}{0.6} & \multicolumn{1}{c}{0.7} & \multicolumn{1}{c}{0.8} & \multicolumn{1}{c}{0.9} & \multicolumn{1}{c}{0.5} & \multicolumn{1}{c}{0.6} & \multicolumn{1}{c}{0.7} & \multicolumn{1}{c}{0.8} & \multicolumn{1}{c}{0.9} \\
    \hline
    NN-MCDA-D1 & 0.607$\pm$0.010 & 0.622$\pm$0.021 & 0.591$\pm$0.009 & 0.607$\pm$0.010 & 0.612$\pm$0.010 & 0.711$\pm$0.024 & 0.721$\pm$0.032 & 0.699$\pm$0.012 & 0.708$\pm$0.017 & 0.719$\pm$0.022 \\
    \hline
    NN-MCDA-D2 & 0.643$\pm$0.008 & 0.638$\pm$0.016 & 0.643$\pm$0.012 & 0.650$\pm$0.008 & 0.635$\pm$0.010 & 0.727$\pm$0.027 & 0.723$\pm$0.029 & 0.711$\pm$0.026 & 0.711$\pm$0.016 & 0.729$\pm$0.021 \\
    \hline
    NN-MCDA-D3 & 0.653$\pm$0.023 & 0.663$\pm$0.018 & 0.657$\pm$0.019 & 0.656$\pm$0.021 & 0.659$\pm$0.013 & 0.741$\pm$0.028 & 0.739$\pm$0.043 & 0.738$\pm$0.037 & 0.726$\pm$0.029 & 0.749$\pm$0.013 \\
    \hline
    NN-MCDA-D5 & 0.678$\pm$0.019 & 0.685$\pm$0.020 & 0.676$\pm$0.005 & 0.687$\pm$0.023 & 0.670$\pm$0.009 & 0.752$\pm$0.035 & 0.758$\pm$0.039 & 0.757$\pm$0.032 & 0.735$\pm$0.035 & 0.761$\pm$0.010 \\
    \hline
    NN-MCDA-D10 & 0.690$\pm$0.023 & 0.686$\pm$0.021 & 0.683$\pm$0.004 & 0.689$\pm$0.009 & 0.674$\pm$0.006 & 0.773$\pm$0.022 & 0.761$\pm$0.048 & 0.766$\pm$0.019 & 0.771$\pm$0.029 & 0.762$\pm$0.031 \\
    \hline
    MLP-D1 & 0.710$\pm$0.001 & 0.711$\pm$0.005 & 0.710$\pm$0.000 & 0.713$\pm$0.000 & 0.712$\pm$0.001 & 0.790$\pm$0.002 & 0.787$\pm$0.004 & 0.780$\pm$0.001 & 0.779$\pm$0.002 & 0.789$\pm$0.003 \\
    \hline
    MLP-D2 & 0.713$\pm$0.003 & 0.712$\pm$0.002 & 0.714$\pm$0.002 & 0.718$\pm$0.003 & 0.715$\pm$0.002 & 0.794$\pm$0.003 & 0.790$\pm$0.003 & 0.782$\pm$0.001 & 0.783$\pm$0.002 & 0.791$\pm$0.001 \\
    \hline
    MLP-D3 & 0.719$\pm$0.004 & 0.719$\pm$0.003 & 0.720$\pm$0.001 & 0.721$\pm$0.002 & 0.719$\pm$0.003 & 0.799$\pm$0.003 & 0.794$\pm$0.002 & 0.791$\pm$0.002 & 0.788$\pm$0.001 & 0.800$\pm$0.003 \\
    \hline
    MLP-D5 & 0.721$\pm$0.002 & 0.720$\pm$0.000 & 0.724$\pm$0.003 & 0.727$\pm$0.003 & 0.725$\pm$0.004 & 0.803$\pm$0.001 & 0.799$\pm$0.005 & 0.801$\pm$0.002 & 0.794$\pm$0.003 & 0.803$\pm$0.001 \\
    \hline
    MLP-D10 & 0.728$\pm$0.003 & 0.724$\pm$0.001 & 0.729$\pm$0.002 & 0.730$\pm$0.002 & 0.729$\pm$0.003 & 0.809$\pm$0.002 & 0.803$\pm$0.001 & 0.805$\pm$0.002 & 0.806$\pm$0.003 & 0.808$\pm$0.002 \\
    \hline
    SVM-Linear & 0.579$\pm$0.005 & 0.579$\pm$0.002 & 0.584$\pm$0.003 & 0.572$\pm$0.005 & 0.579$\pm$0.006 & 0.658$\pm$0.003 & 0.659$\pm$0.005 & 0.649$\pm$0.004 & 0.645$\pm$0.002 & 0.658$\pm$0.005 \\
    \hline
    SVM-RBF & 0.586$\pm$0.003 & 0.584$\pm$0.004 & 0.586$\pm$0.002 & 0.578$\pm$0.002 & 0.589$\pm$0.005 & 0.660$\pm$0.001 & 0.661$\pm$0.002 & 0.654$\pm$0.003 & 0.648$\pm$0.003 & 0.658$\pm$0.003 \\
    \hline
    SVM-D3poly & 0.574$\pm$0.002 & 0.579$\pm$0.006 & 0.580$\pm$0.006 & 0.572$\pm$0.004 & 0.579$\pm$0.005 & 0.659$\pm$0.004 & 0.662$\pm$0.002 & 0.652$\pm$0.002 & 0.647$\pm$0.002 & 0.655$\pm$0.006 \\
    \hline
    GAM-D3 & 0.579$\pm$0.001 & 0.579$\pm$0.002 & 0.584$\pm$0.003 & 0.572$\pm$0.001 & 0.576$\pm$0.003 & 0.659$\pm$0.002 & 0.659$\pm$0.004 & 0.650$\pm$0.003 & 0.647$\pm$0.005 & 0.656$\pm$0.004 \\
    \hline
    GAM-D10 & 0.583$\pm$0.001 & 0.582$\pm$0.003 & 0.585$\pm$0.003 & 0.579$\pm$0.002 & 0.583$\pm$0.005 & 0.660$\pm$0.004 & 0.659$\pm$0.003 & 0.651$\pm$0.003 & 0.645$\pm$0.002 & 0.657$\pm$0.006 \\
    \hline
    DeciTr-MaxDep6 & 0.583$\pm$0.003 & 0.583$\pm$0.003 & 0.583$\pm$0.003 & 0.581$\pm$0.005 & 0.589$\pm$0.009 & 0.649$\pm$0.002 & 0.650$\pm$0.001 & 0.644$\pm$0.004 & 0.638$\pm$0.005 & 0.651$\pm$0.008 \\
    \hline
    DeciTr-MaxDep10 & 0.565$\pm$0.004 & 0.579$\pm$0.002 & 0.574$\pm$0.004 & 0.570$\pm$0.007 & 0.562$\pm$0.008 & 0.631$\pm$0.004 & 0.635$\pm$0.003 & 0.632$\pm$0.005 & 0.628$\pm$0.006 & 0.638$\pm$0.009 \\
    \hline
    DeciTr-MaxDep20 & 0.536$\pm$0.005 & 0.548$\pm$0.006 & 0.539$\pm$0.004 & 0.541$\pm$0.007 & 0.547$\pm$0.008 & 0.592$\pm$0.003 & 0.583$\pm$0.004 & 0.591$\pm$0.002 & 0.588$\pm$0.008 & 0.582$\pm$0.007 \\
    \hline
    PLR-D1 & 0.560$\pm$0.015 & 0.562$\pm$0.026 & 0.558$\pm$0.021 & 0.559$\pm$0.017 & 0.557$\pm$0.011 & 0.632$\pm$0.024 & 0.630$\pm$0.017 & 0.629$\pm$0.019 & 0.631$\pm$0.022 & 0.633$\pm$0.020 \\
    \hline
    PLR-D2 & 0.562$\pm$0.014 & 0.568$\pm$0.021 & 0.560$\pm$0.020 & 0.560$\pm$0.010 & 0.562$\pm$0.017 & 0.637$\pm$0.019 & 0.631$\pm$0.021 & 0.633$\pm$0.011 & 0.633$\pm$0.017 & 0.635$\pm$0.019 \\
    \hline
    PLR-D3 & 0.569$\pm$0.010 & 0.569$\pm$0.019 & 0.564$\pm$0.019 & 0.565$\pm$0.028 & 0.563$\pm$0.028 & 0.642$\pm$0.023 & 0.637$\pm$0.019 & 0.637$\pm$0.023 & 0.639$\pm$0.028 & 0.639$\pm$0.020 \\
    \hline
    PLR-D5 & 0.571$\pm$0.014 & 0.572$\pm$0.023 & 0.569$\pm$0.020 & 0.569$\pm$0.027 & 0.569$\pm$0.027 & 0.647$\pm$0.022 & 0.642$\pm$0.020 & 0.640$\pm$0.029 & 0.641$\pm$0.018 & 0.641$\pm$0.018 \\
    \hline
    PLR-D10 & 0.578$\pm$0.015 & 0.575$\pm$0.012 & 0.572$\pm$0.023 & 0.575$\pm$0.019 & 0.573$\pm$0.019 & 0.652$\pm$0.021 & 0.649$\pm$0.011 & 0.644$\pm$0.016 & 0.645$\pm$0.019 & 0.647$\pm$0.022 \\
    \hline
    Mean & 0.621$\pm$0.005	&0.623$\pm$0.010	&0.621$\pm$0.008&	0.621$\pm$0.005&	0.621$\pm$0.005&	0.699$\pm$0.012&	0.697$\pm$0.014	& 0.693$\pm$0.011	&0.690$\pm$0.013	&0.698$\pm$0.011 \\
        \hline
    \end{tabular}}
  \label{tab-simexp1data3}
\end{table}%

We also compare the proposed NN-MCDA with baseline machine learning models, including the standard MLP, polynomial linear regression (PLR) with 1, 2, 3, 5 and 10 degrees, GAM with 10 splines that are in 3 and 10 degree of polynomials, SVMs with linear, radial basis function (RBF) and polynomial kernels, and single decision tree (DeciTr) models with 6, 10 and 20 maximum depths. Table \ref{tab-simexp1data1} presents the results for the simplest dataset. All machine learning models, both the interpretable (including NN-MCDA) and full complexity ones, perform well. The performance drops rapidly when we use them to fit the nonlinear and high-order datasets (shown in Tables \ref{tab-simexp1data2} and \ref{tab-simexp1data3}). Since the proposed NN-MCDA model and MLP can model the nonlinearity and attribute interactions, both of them achieve much higher AUCs as compared to the rest. As expected, although the performance of NN-MCDA is lower than MLP, the difference is relatively small. Both NN-MCDA and MLP outperform other baseline machine learning models significantly. More specifically, we can observe that GAM, SVM and DeciTr have similar accuracy, which is higher than that of PLR, but lower than NN-MCDA. NN-MCDA's performance is close to the full complexity model, and at the same time, it preserves a certain degree of interpretability (will be shown in the next experiment).

\subsubsection{Experiment II: Impact of $\alpha$ on AUC}
\label{subsubsec-exp2}

In this section, we focus on assessing how the performance of the NN-MCDA model is affected by the value of $\alpha$, the weight for the linear component. We evenly sample 20 values within [0, 1] as the pre-defined $\alpha$. For each fixed $\alpha$, we train the NN-MCDA model using synthetic datasets introduced in the previous subsection. In this experiment, we use the SGD algorithm to optimize the parameters and the number of iterations are set as 250.

\begin{figure}[h]
\centering
{\includegraphics[scale=0.18]{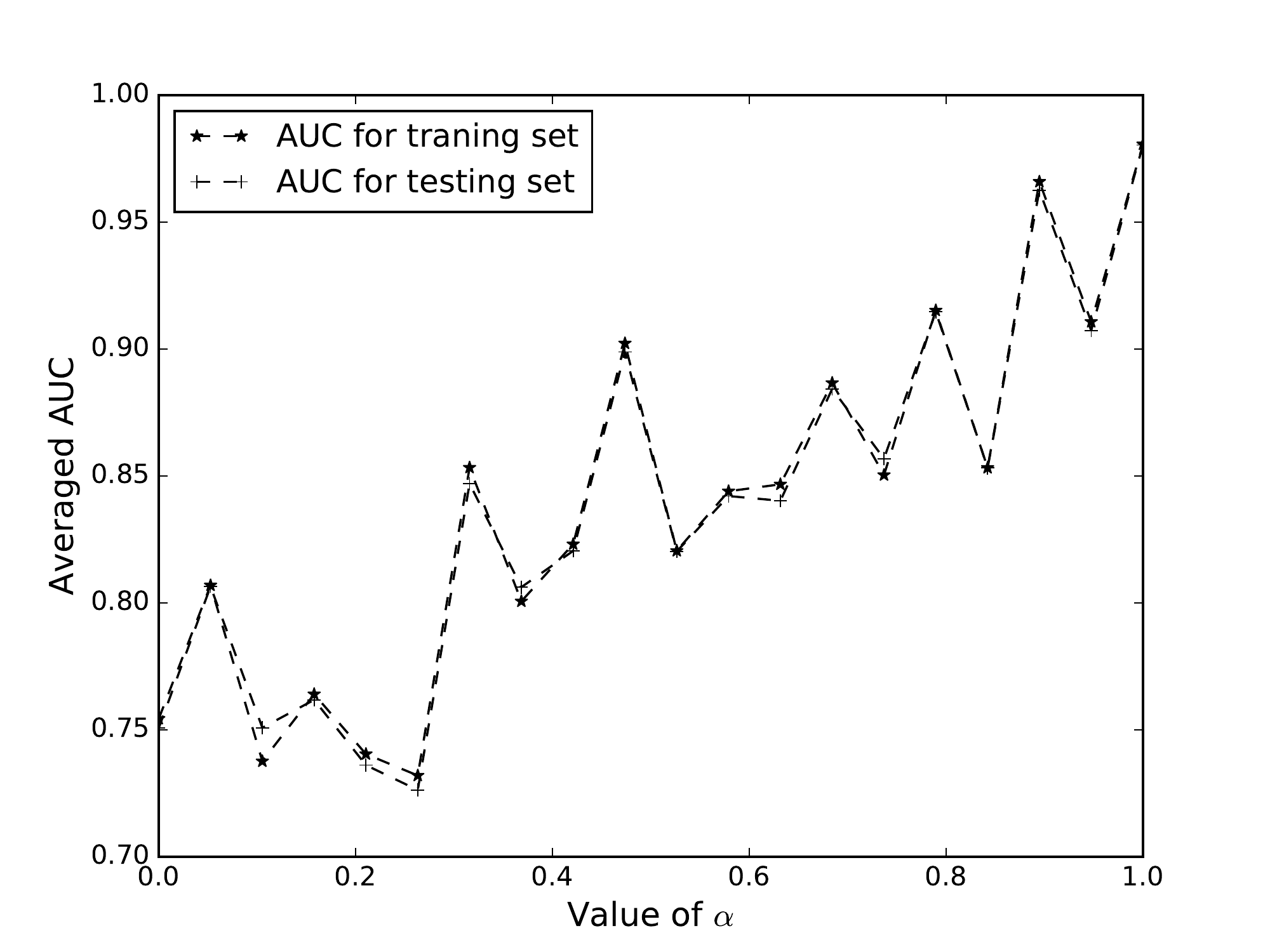}
\includegraphics[scale=0.18]{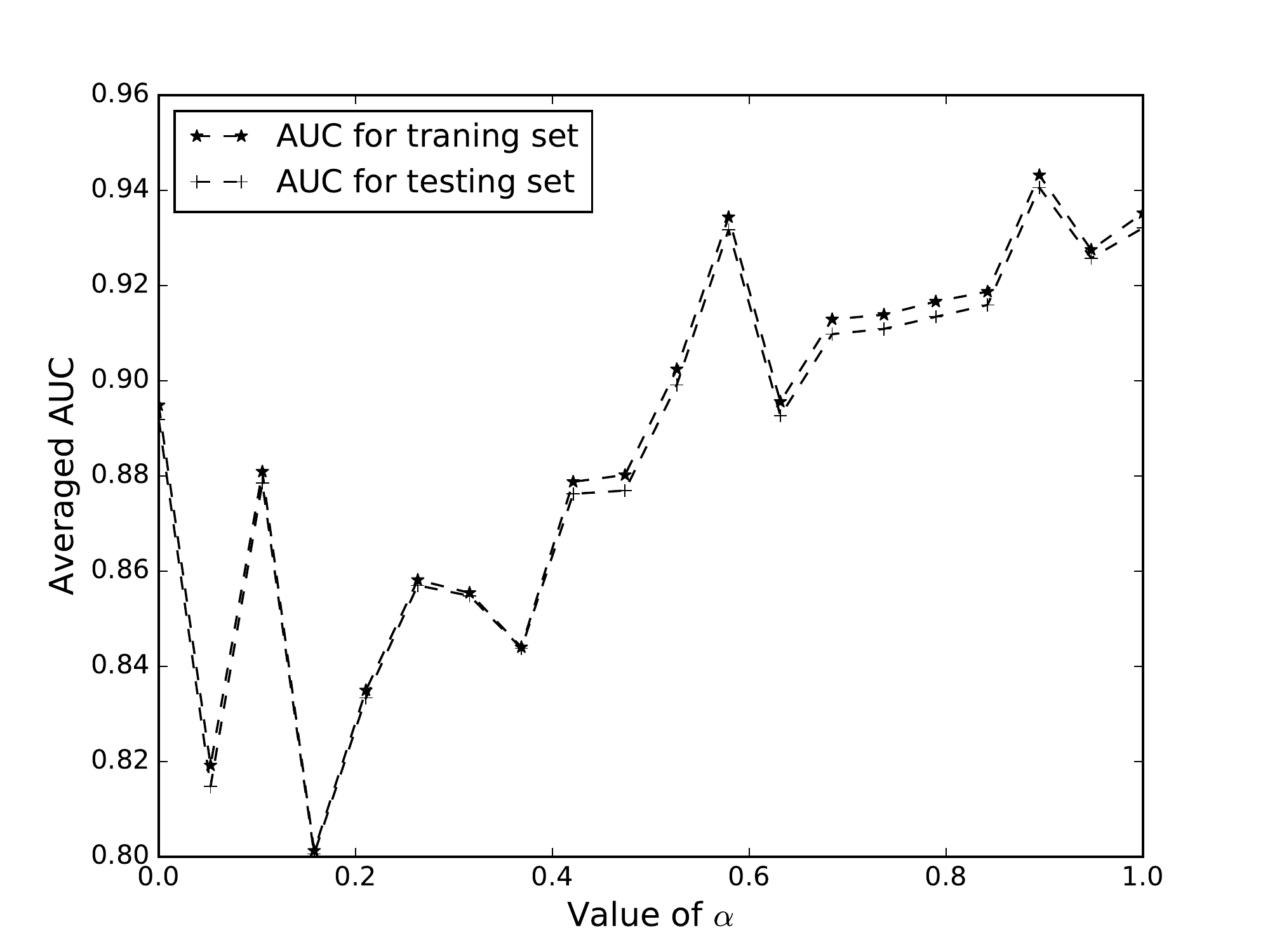}
\includegraphics[scale=0.18]{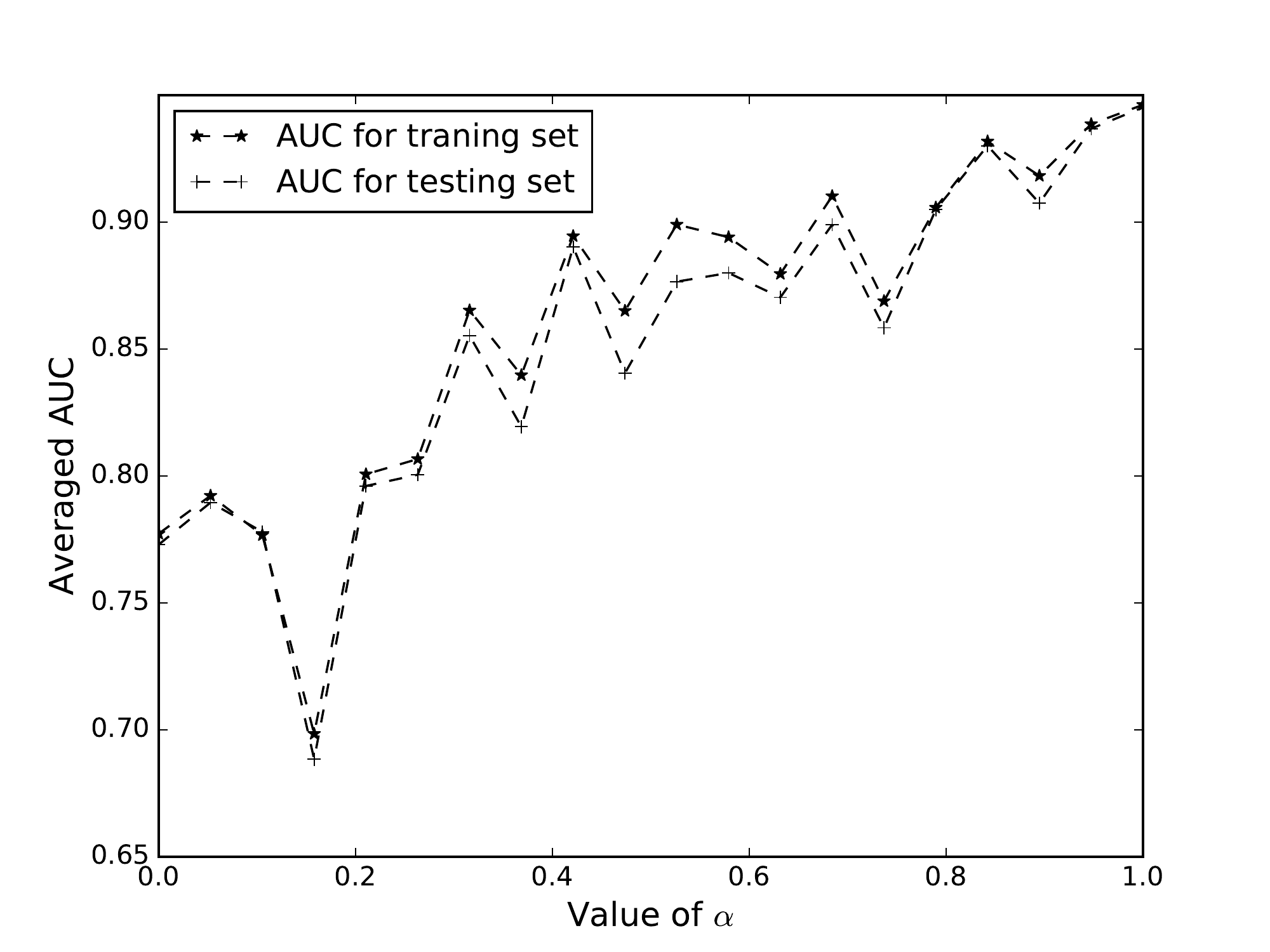}}
\caption{From the left to right, the impact of $\alpha$ on AUC using $\mathcal{D}_l^3,\mathcal{D}_l^5$, $\mathcal{D}_l^{10}$, respectively. \label{fig-exp2linear}}
\end{figure}

\begin{figure}[h]
\centering
{\includegraphics[scale=0.18]{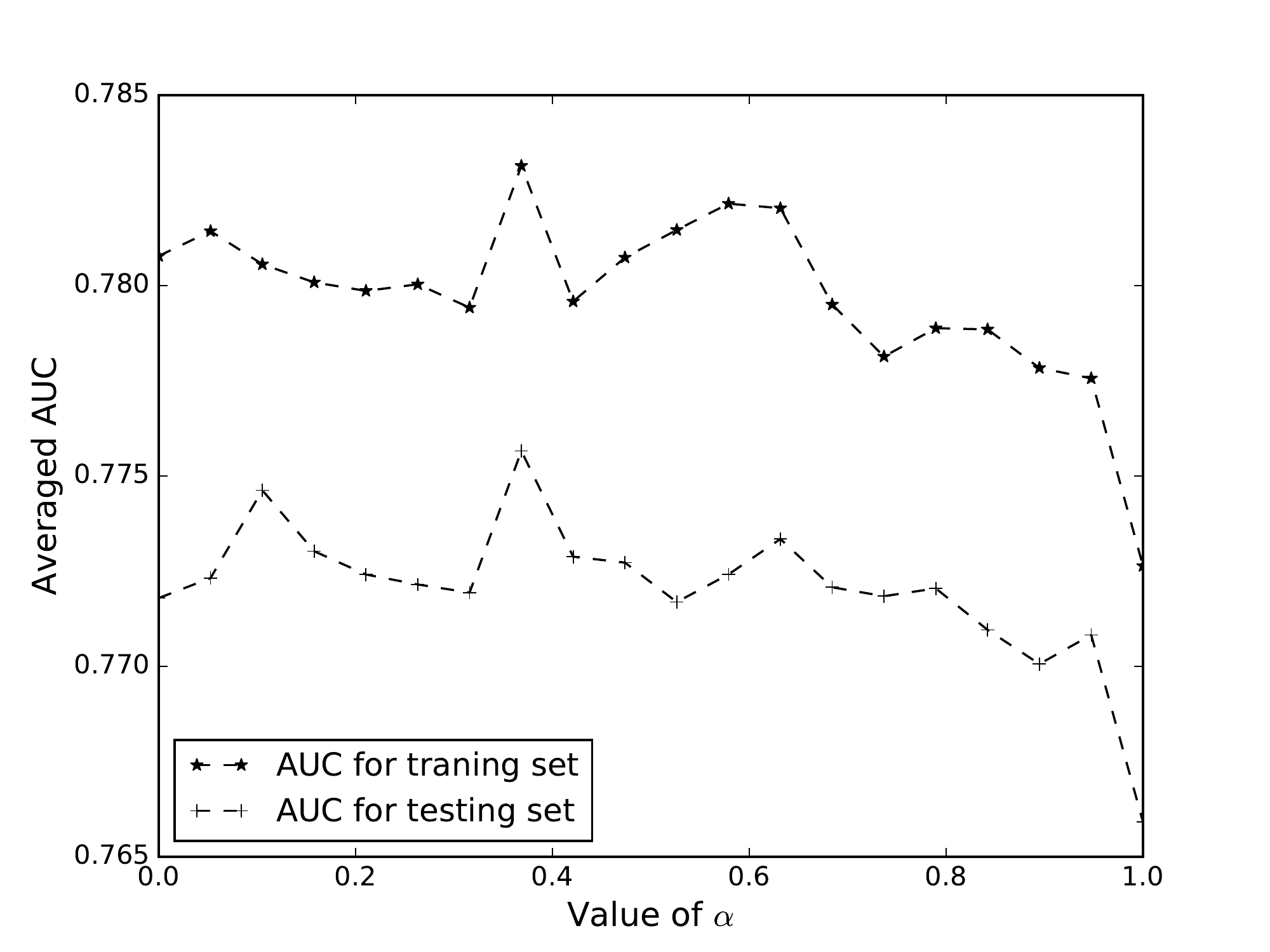}
\includegraphics[scale=0.18]{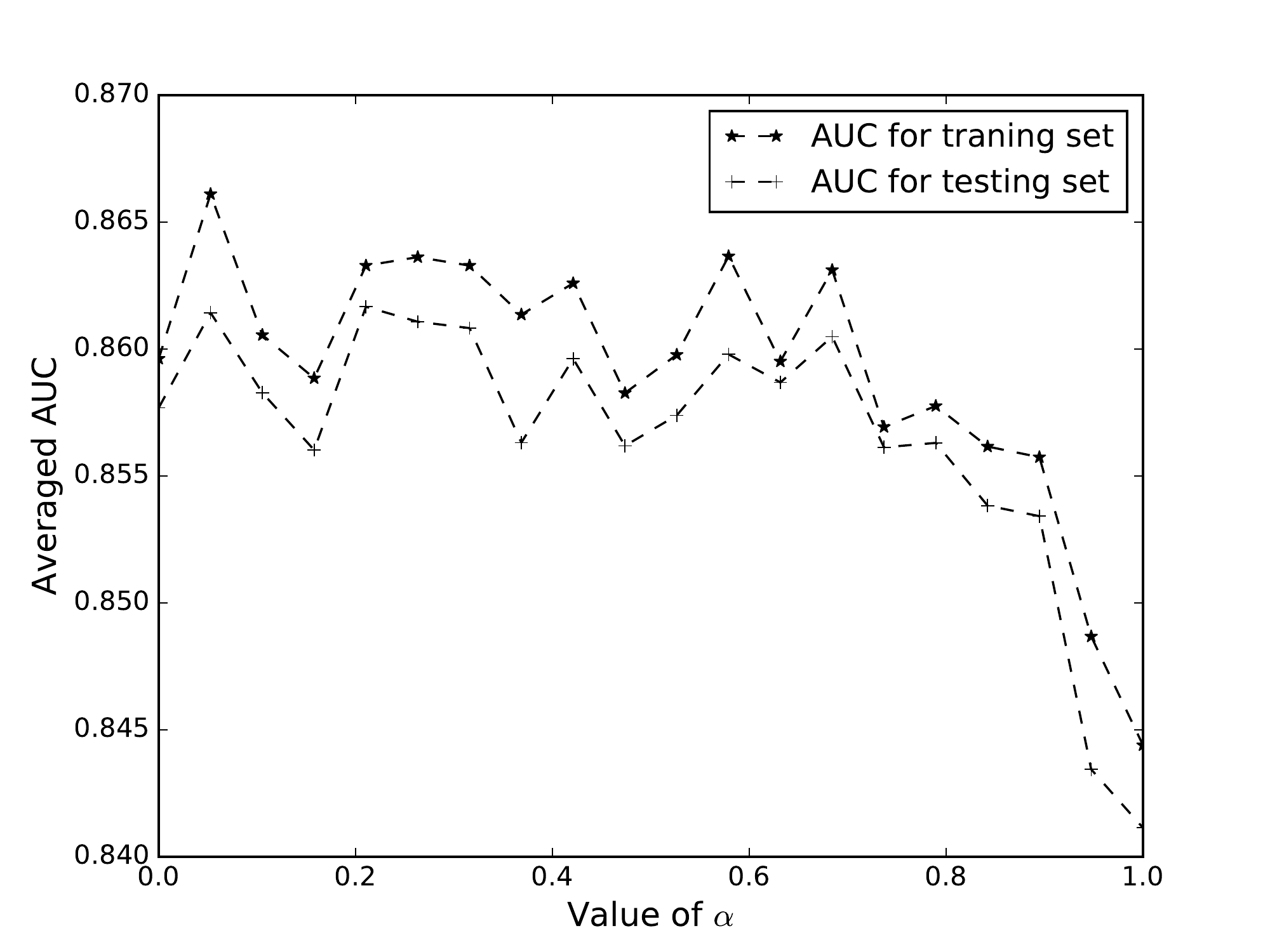}
\includegraphics[scale=0.18]{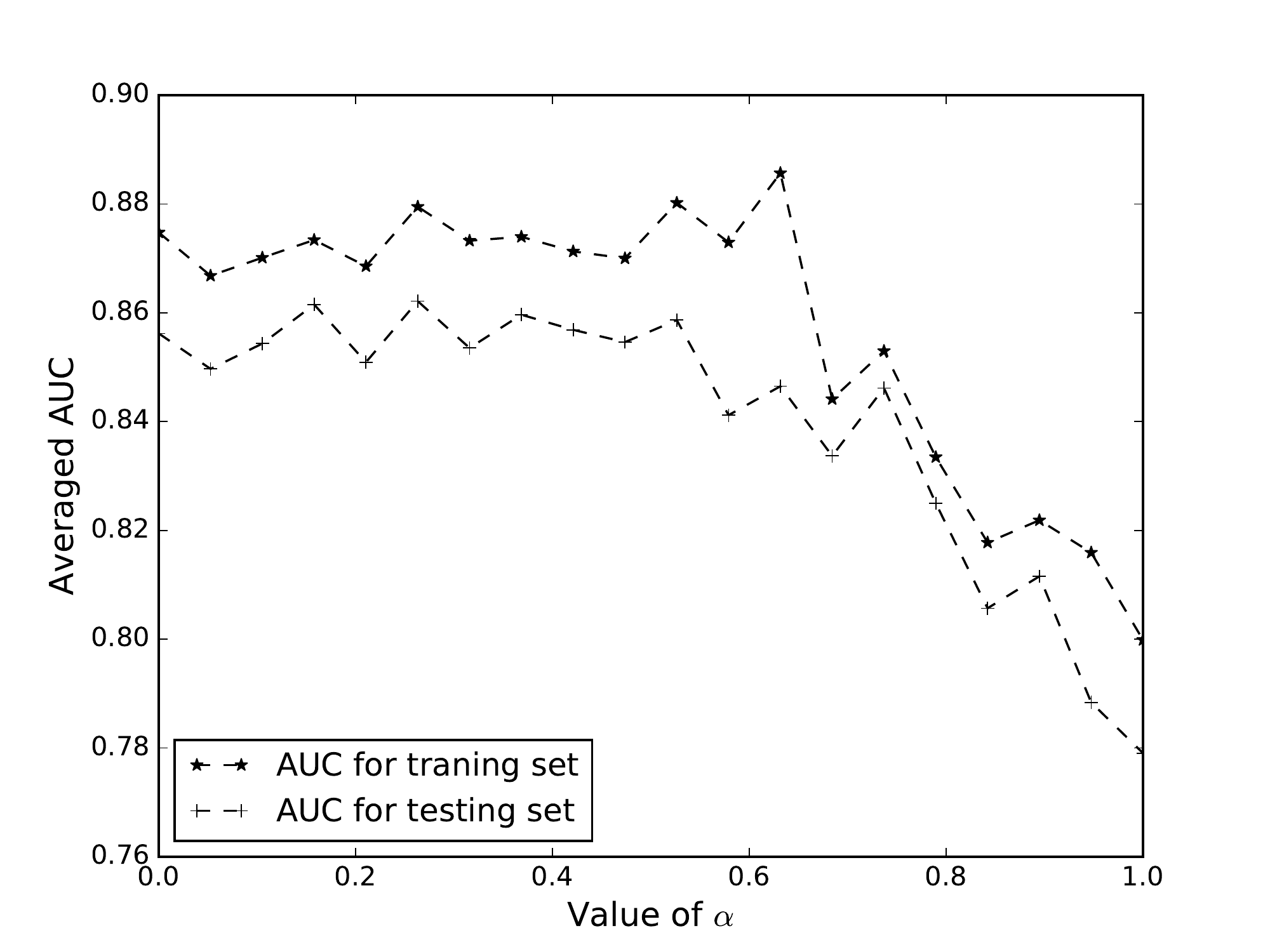}}
\caption{From the left to right, the impact of $\alpha$ on AUC using $\mathcal{D}_{polynomial-3}^3,\mathcal{D}_{polynomial-3}^5$, $\mathcal{D}_{polynomial-3}^{10}$, respectively. \label{fig-exp2random}}
\end{figure}

\begin{figure}[htbp]
\centering
{\includegraphics[scale=0.18]{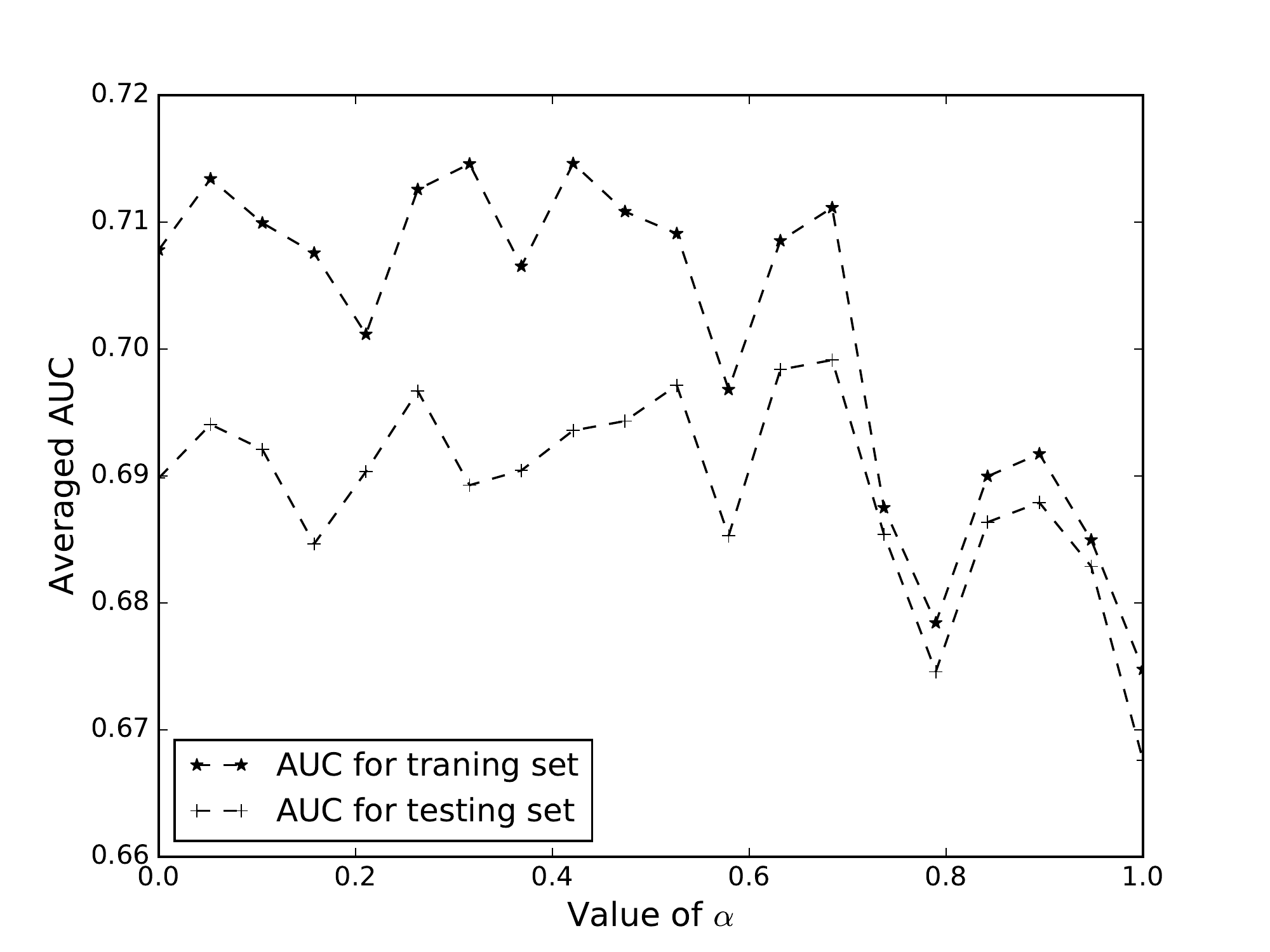}
\includegraphics[scale=0.18]{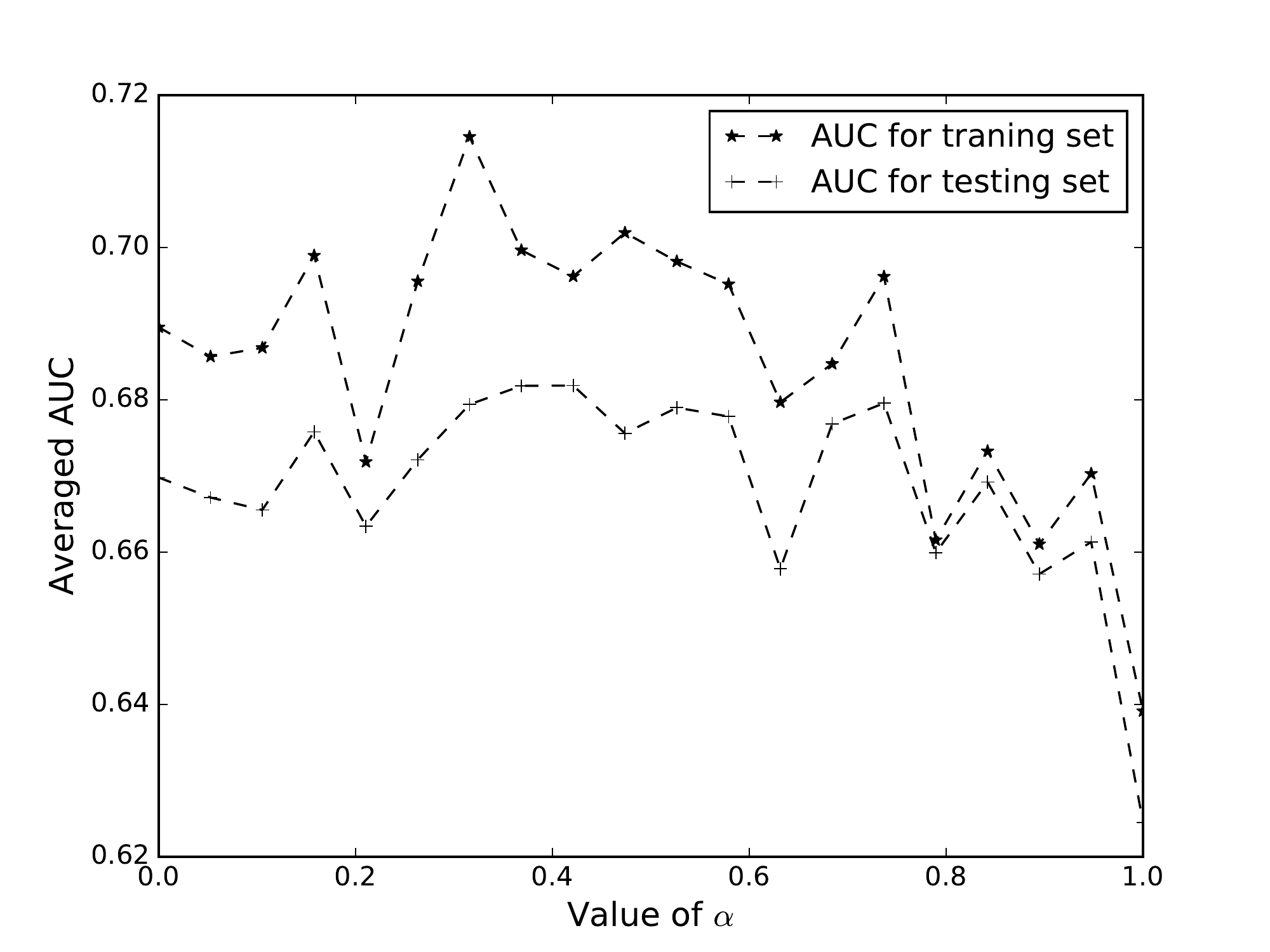}
\includegraphics[scale=0.18]{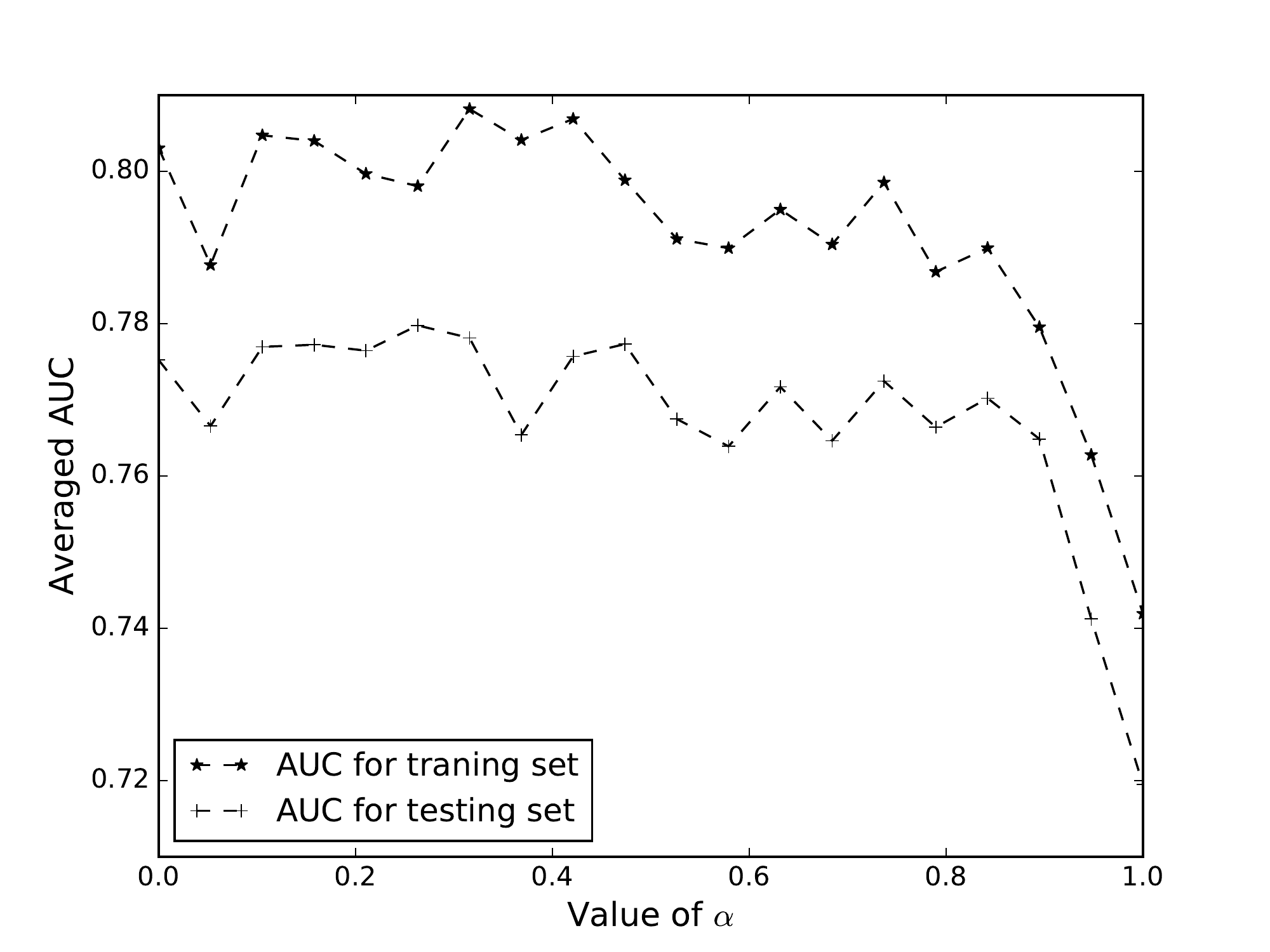}}
\caption{From the left to right, the impact of $\alpha$ on AUC using $\mathcal{D}_{polynomial-15}^3,\mathcal{D}_{polynomial-15}^5$, $\mathcal{D}_{polynomial-15}^{10}$, respectively. \label{fig-exp2random2}}
\end{figure}

In Figure \ref{fig-exp2linear} (results with the linearly generated datasets), though three curves have many monotonicity inflexions, there is a general trend that the greater the value of $\alpha$, the better the performance. NN-MCDA obtains the best results when $\alpha$ is between 0.8 and 1.0 on these linearly generated datasets. In contrast, the AUC curves have a general decreasing trend when $\alpha$ increases for the nonlinearly generated datasets (Figures \ref{fig-exp2random} and \ref{fig-exp2random2}). In the extreme cases, where only the nonlinear MLP component (i.e., $\alpha=0$) or the linear component (i.e.,$\alpha = 1$) works, the model obtains the greatest or smallest average AUC.

Note that the three datasets have very different patterns. Datasets $\mathcal{D}^n_l$ use a set of simple linear marginal value functions whereas $\mathcal{D}^n_{polynomial-3}$ and $\mathcal{D}^n_{polynomial-15}$ simulate more complicated patterns. Theoretically, a full complexity model (MLP) can perfectly capture any patterns in the data at the cost of very large numbers of iterations and data samples for convergence. In practice, we often do not have sufficient data or computational time to achieve the optimal MLP solution. In this simulation experiment (31,125 data points and 250 iterations to fit the model), a pure MLP model (NN-MCDA with $\alpha=0$) does not always lead to the best outcome. This result indicates that, in real-world managerial decision making, a full complexity model is usually not the best one not only because of the lack of interpretability, but also the limited data and computational resources to optimize the model. It is sensible to allow the model to automatically adjust the trade-off coefficient $\alpha$ to avoid the scenarios where a very complex model is used to fit simple data, or a simple model is used to fit complex data.

\subsubsection{Experiment III: performance in fitting actual marginal value functions}
\label{subsubsec-exp3}

This experiment studies the ability of the NN-MCDA model to reconstruct the actual marginal value functions. From Experiments I and II, we find that the NN-MCDA model with degree equal to 3 has a good balance between prediction performance and computational cost. Therefore, in this experiment, we generate four typical synthetic models with different complexities. Each hypothetical model has three marginal value functions to estimate. Then, we use the synthetic models to generate datasets with the same attribute vectors (model input). We approximate the marginal value functions using an NN-MCDA model with a degree of 3. We also compare the obtained function with a baseline linear regression model. The four synthetic models (from the simplest to extremely complicated) are described as follows:
\begin{itemize}
\item Synthetic model 1: Three linear marginal value functions. The global scores are the linear summation of three marginal values without interactions.
\item Synthetic model 2: Three polynomial functions of degree 3. The global scores are the linear summation of three marginal values with all possible pairwise interactions.
\item Synthetic model 3: A polynomial function of degree 15, a sigmoid function and an exponential function. The global scores are the linear summation of three marginal values without interactions.
\item Synthetic model 4: The Model 3 with pairwise and triple-wise attribute interactions.
\end{itemize}

\begin{figure}[htbp]
\centering
{\includegraphics[scale=0.3]{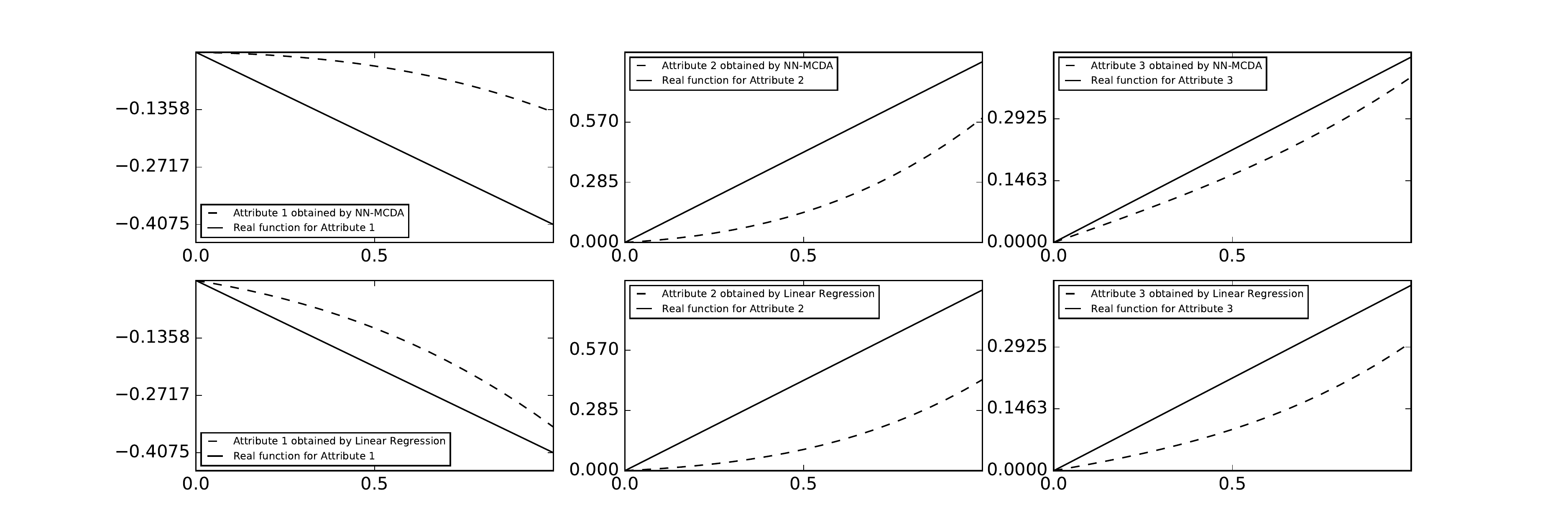}}
\caption{The simulated and actual marginal value functions in model 1. \label{fig-exp3-1}}
\end{figure}

\begin{figure}[htbp]
\centering
{\includegraphics[scale=0.3]{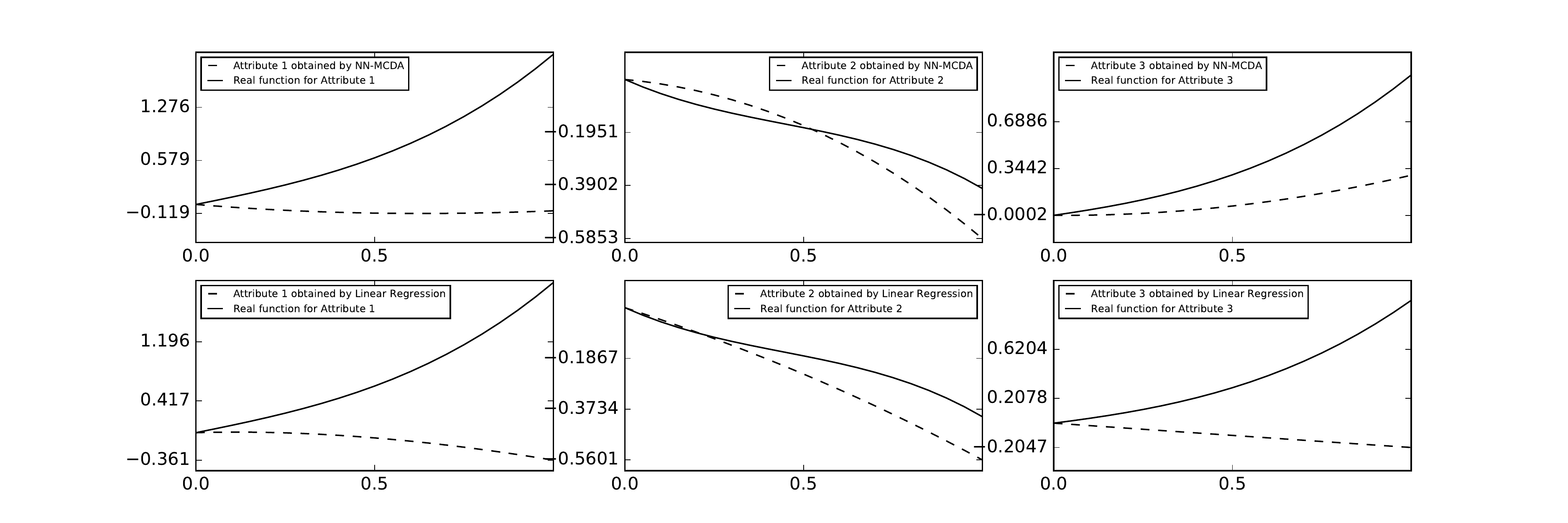}}
\caption{The simulated and actual marginal value functions in model 2. \label{fig-exp3-2}}
\end{figure}

\begin{figure}[htbp]
\centering
{\includegraphics[scale=0.3]{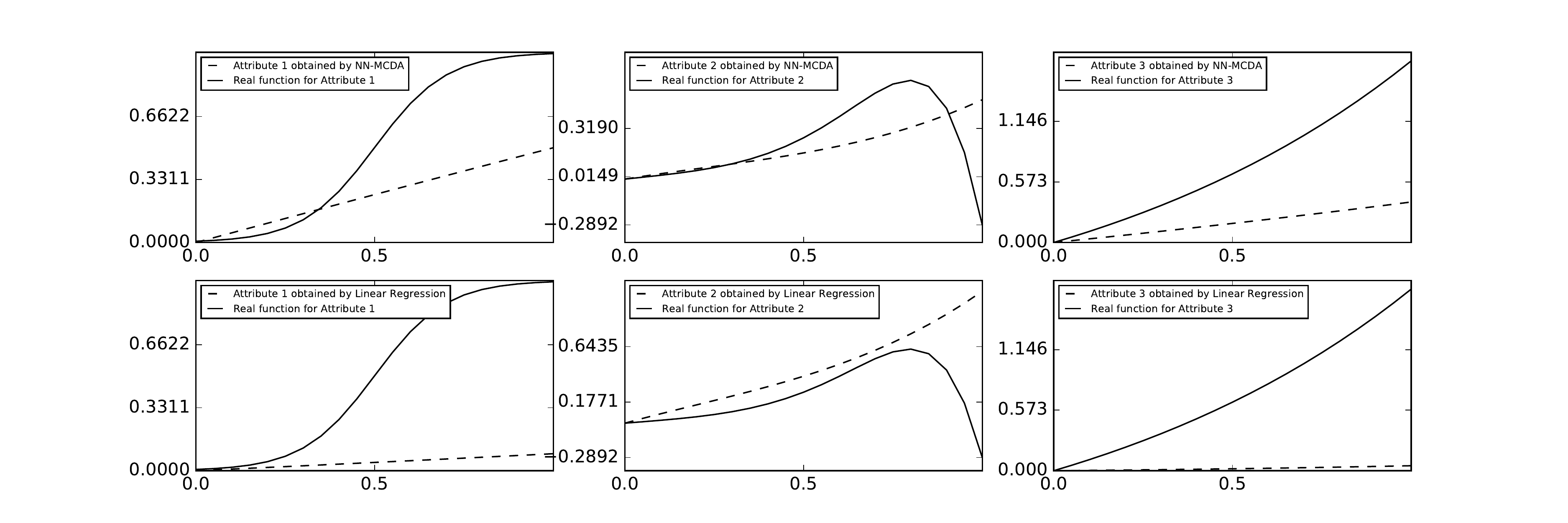}}
\caption{The simulated and actual marginal value functions in model 3. \label{fig-exp3-3}}
\end{figure}

\begin{figure}[h]
\centering
{\includegraphics[scale=0.3]{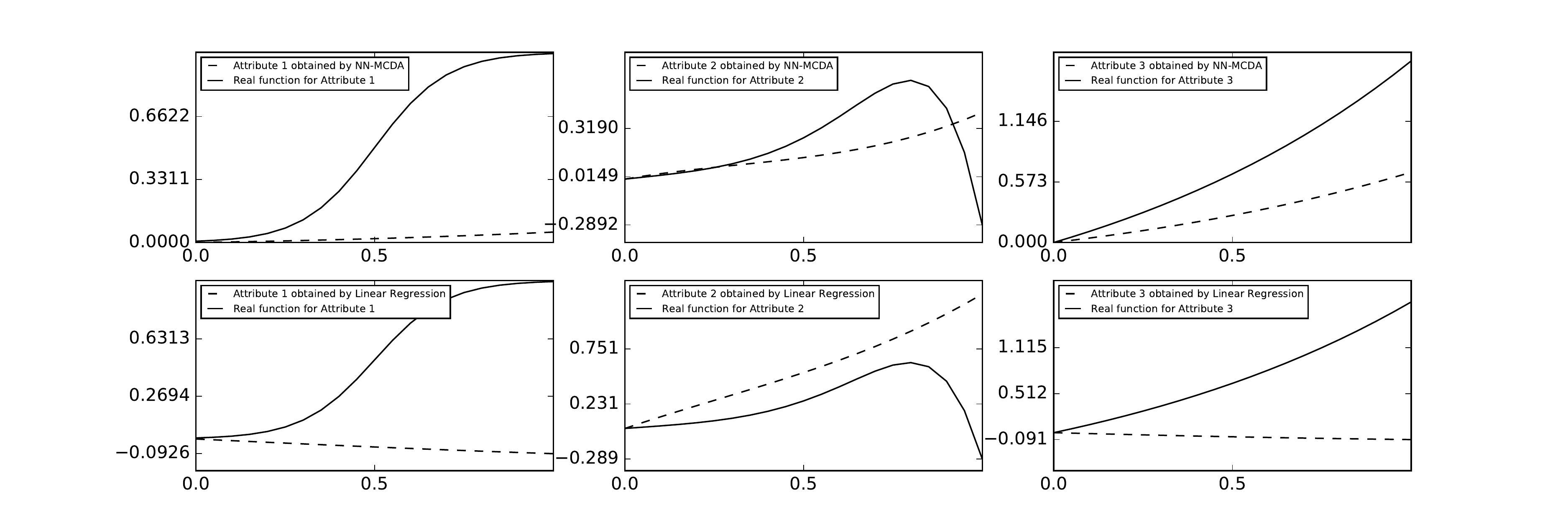}}
\caption{The simulated and actual marginal value functions in model 4. \label{fig-exp3-4}}
\end{figure}

Figures \ref{fig-exp3-1} to \ref{fig-exp3-4} reveal the actual and fitted marginal value functions obtained by the proposed NN-MCDA model and the baseline linear regression model. For a simple model with linear marginal value functions (Synthetic model 1), both the baseline linear regression model and the NN-MCDA can fit the actual functions well. Both models successfully capture the monotonicity of the original marginal value functions. This indicates that the NN-MCDA model is also applicable to simple prediction tasks that do not have attribute interactions or nonlinear associations between attributes and predictions.

When the attribute interactions are considered (Synthetic model 2), the proposed model outperforms the baseline linear regression model. In the first row of Figure \ref{fig-exp3-2}, the NN-MCDA model captures correct monotonicity changes of all three actual marginal value functions. Moreover, for the first and third attributes, it captures the concavity of the original functions. While linear regression model gives opposite monotonicity of the actual functions except for the second attribute. In addition, the linear regression model does not capture the concavity of the actual functions. The bad performance of the linear regression model is due to the fact that it can not capture the interactions between attributes, which often brings distorted interpretability. 

For the model with more complex marginal value functions (Synthetic model 3), the linear regression model performs rather bad. In Figure \ref{fig-exp3-3}, the first and third attributes have very negligible impact on the prediction in the fitted linear regression model. The NN-MCDA model, on the other hand, correctly captures the main characteristics of the three attributes. Both models failed to capture the inflexion point for the second attribute. These results demonstrate that a low-degree NN-MCDA model (3-degree in this study) can fit both lower and higher degree marginal value functions, because of the nonlinear component helps deal with the complexities that cannot be captured by the predefined linear component. In Figure \ref{fig-exp3-4}, if the marginal value function is extremely complex (Synthetic model 4), though low-degree NN-MCDA cannot fully capture the characteristics of the marginal value functions, it still outperforms the baseline linear regression model. However, the real-world decision making behaviors are usually not that complex (15-degree in Synthetics model 3 and 4). We will further validate the applicability and generalizability of the proposed NN-MCDA model with real datasets in the following section.

\subsection{A multiple criteria ranking problem.}
\label{subsec-real}

QS world university ranking organization\footnote{It is an annual publication of university rankings by Quacquarelli Symonds (QS). \url{https://www.topuniversities.com}} provides five carefully-chosen indicators to measure the universities' capacity in producing the most employable graduates, including employer reputation, employer-student connection, alumni outcomes, partnerships with employers and graduate employment rate. The metric of employer reputation, regarded as a key performance indicator, is based on over 40,000 responses to the QS Employer Survey. In this experiment, we apply the proposed NN-MCDA model to a multiple criteria ranking problem that predicts the employer reputation (human decision) using the other four quantitative indicators\footnote{\textbf{Employer-student connection} (EC). This indicator involves the number of active presences of employers on a university's campus over the past 12 months. Such presences are in form of providing students with opportunities to network and acquire information, organizing company presentations or other self-promoting activities, which increase the probability that students have to participate in career-launching internships and research opportunities.\\ 
\textbf{Alumni outcomes} (AO). The scores based on the outcomes of a university's graduates produced. A university is successful if its graduates tend to produce more wealth and scientific researches.\\
\textbf{Partnerships with employers} (PE). The number of citable and transformative researches which are produced by a university collaborating successfully with global companies.\\
\textbf{Graduate employment rate} (GER). This indicator is essential for understanding how successful universities are at nurturing employability. It involves measuring the proportion of graduates (excluding those opting to pursue further study or unavailable to work) in full or part time employment within 12 months of graduation.}. The descriptive statistics are shown in Table \ref{tab-rankUdesription}.

\begin{table}[htbp]
\centering
\caption{The descriptive statistics of data. {Original data is normalized into $[0,1]$ in the following experiment.} \label{tab-rankUdesription}}

\resizebox{0.95\textwidth}{!}{
\begin{tabular}{lllll}
    \hline
          & Mean  & Std Dev & Minimum & Maximum \\
    \hline
    Employer-student connection & 74.63 & 11.17 & 44.80 & 100.00 \\
    Alumni outcomes & 55.25 & 16.74 & 26.40 & 100.00 \\
    Partnerships with employers & 73.27 & 10.76 & 46.10 & 100.00 \\
    Graduate employment rate & 77.88 & 7.93  & 54.00 & 100.00 \\
    \hline
    \end{tabular}%
}
\end{table}

\begin{figure}[h]
\centering
{\includegraphics[scale=0.35]{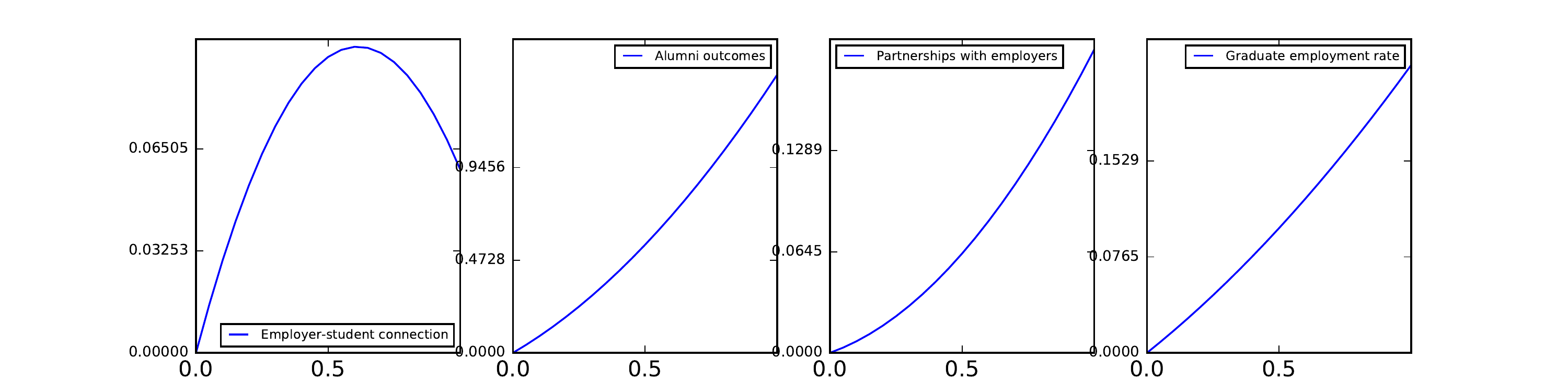}}
\caption{Marginal value functions obtained by NN-MCDA for the university data. We use polynomial functions of degree 2 to approximate the model. The black dashed line is the baseline rate satisfying $p(\hat{y}=1|\mathbf{x})=p(\hat{y}=0|\mathbf{x})=0.5$. Same in following Figures. \label{fig-ranku-nnmcda}}
\end{figure}

\begin{figure}[h]
\centering
{\includegraphics[scale=0.35]{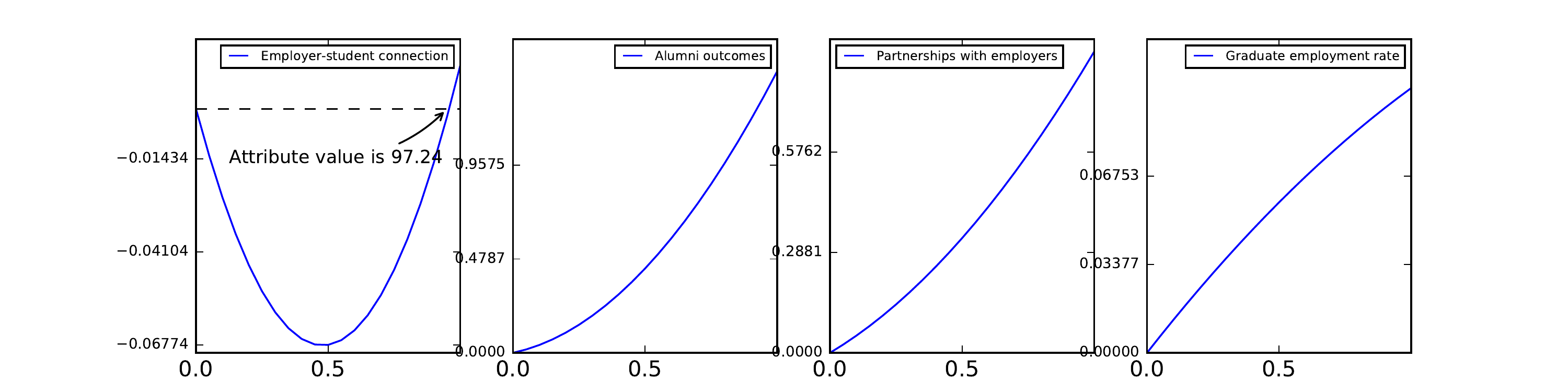}}
\caption{Marginal value functions obtained by logistic regression model for the university data. \label{fig-ranku-linear}}
\end{figure}

\begin{figure}[htbp]
\centering
{\includegraphics[scale=0.35]{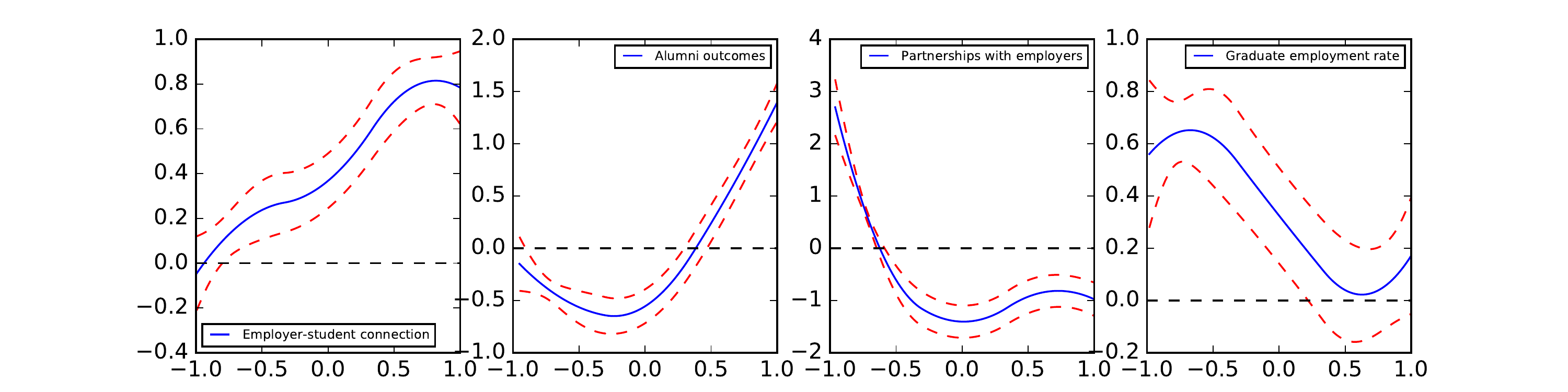}}
\caption{Marginal value functions obtained by GAM with 5 splines that are in 2-degree polynomials for the university data. The red dashed lines are the 95\% confidence bands.\label{fig-ranku-GAM}}
\end{figure}

There are $\binom{250}{2}$ pairwise comparisons among 250 universities. To determine the pre-defined degree of polynomials, we respectively set $D_j$ as 1, 2, and 3. We use the same fivefold cross-validation process to fit the model. We record the averaged AUC for both training and testing sets. We also select three baseline models, including a 3-layer MLP, a logistic regression model and a GAM. The results of the average AUC are presented in Table \ref{tab-rankUMSE}. For each pre-defined degree, the full complexity model always obtain the best results whereas the logistic regression model performs the worst. Since NN-MCDA can model attribute interactions, it slightly outperforms the GAM. We depict the marginal value functions obtained by NN-MCDA, logistic regression model and GAM in Figures \ref{fig-ranku-nnmcda}, \ref{fig-ranku-linear} and \ref{fig-ranku-GAM}, respectively.

\begin{table}[htbp]
\centering
\caption{Given polynomials of pre-defined degree, the averaged AUC for training and testing sets. The value in parentheses is $\alpha$ when convergence. There are minor differences between averaged AUC when pre-defined degrees are 2 and 3. Using polynomial of degree 2 is sufficient because more complex model does not significantly increase AUC. \label{tab-rankUMSE}}
\resizebox{0.95\textwidth}{!}{
\begin{tabular}{lllll}
    \hline
          & NN-MCDA  & Logistic regression & MLP & GAM (5 splines)  \\
    \hline
    1-degree Training  & 0.66 (0.8988) & 0.61 (1.00) & 0.71(0.00) & 0.64 \\
    1-degree Testing & 0.64 &  0.57 & 0.70  & 0.63\\
    2-degree Training & 0.69 (0.4969)  & 0.64 (1.00) & 0.73 (0.00) & 0.67 \\
    2-degree Testing & 0.68 & 0.63 & 0.71 & 0.66\\
    3-degree Training & 0.69 (0.4351) &  0.65 (1.00) & 0.74 (0.00) & 0.68 \\
    3-degree Testing & 0.68 &  0.65 & 0.73  & 0.66 \\
    \hline
    \end{tabular}%
}
\end{table}

In Figures \ref{fig-ranku-nnmcda} and \ref{fig-ranku-linear}, the vertical axis is the individual attributes contributions to employer reputation. For the attributes \textit{alumni outcomes}, \textit{partnerships with employers} and \textit{graduate employment rate}, the value functions obtained by NN-MCDA and logistic regression model exhibit a monotonically increasing trend, which makes sense based on our common knowledge. For attribute \textit{employer-student connection}, GAM obtains a generally increasing curve. Although it also captures a slight dip of the increasing trend for \textit{employer-student connection}greater than 0.8, such effect is not as clear as that in the curve captured by NN-MCDA. For attributes \textit{alumni outcomes}, \textit{partnerships with employers} and \textit{graduate employment rate}, GAM obtains quite different and unstable curves that are difficult to explain.

\subsection{Predicting geriatric depression risk.}
\label{subsec-preddeprisk}

Depression is a major cause of emotional suffering in later life. Geriatric depression reduces the quality of older adults' life and increases the risk for acquiring other diseases and committing suicide \citep{alexopoulos2005depression}. The existing literature empirically studied the risk factors of geriatric depression but few of them gave insight into how these risk factors affect the prevalence of geriatric depression in details (i.e.: the shape of the marginal value functions). It is more managerially helpful for clinical decision making to prevent older adults from being depressed if we can understand how each risk factor influences the risk for depression at different scales.

The Health and Retirement Study (HRS) is a nationally representative longitudinal study of US adults \citep{bugliari2016rand}. It has been widely used in many medical studies because of its massive information about older adults demographics, health status, health care utilization and costs, and other useful variables \citep{pool2018association}. We sample the data in 2014 ($N=17,696$). In this experiment, given five pre-determined attributes (risk factors) that have been found to be associated with geriatric depression, we want to capture the detailed effect of these risk factors at different scales, which are represented by marginal value functions. The descriptive statistics are described in Table \ref{tab-descrpdepre}\footnote{Center for Epidemiologic Studies Depression (CES-D) scale is a self-report measure of the frequency of 20 depressive symptoms during the past week \citep{radloff1991use}. It is one of the most popular index assessing the risk for being depressed \citep{beardslee2013prevention, brent2015effect, garber2009prevention}. Age in years at the end of the survey is calculated from the respondent birth date and beginning survey date. The samples vary from 18 to 104 but very few of them are within 18-45 and 93-104 years \citep{blazer1991association, mirowsky1992age}. Degree of education is measured by the years of getting education. It is a categorical variable varying from 0 to 18. Note that for respondents whose degree of education is higher than 17 years, we set the degree of education as 18 \citep{ladin2008risk, murrell1983prevalence}. Marital status is represented by the length of the longest marriage that respondent ever had. It is a continuous variable and varies from 0 to 74.2 years \citep{pearlin1977marital, kessler1982marital, penninx1998depressive}. Out-of-pocket expenditure refers to the expenses that the respondent pays directly to the health care provider without a third-party (insurer, or State). We consider the out-of-pocket medical expenditure in previous 2 years \citep{gadit2004out}. Body mass index is the weight divided by the square height \cite{bugliari2016rand}, which determines whether a respondent is overweight \citep{dong2004relationship, luppino2010overweight, ross1994overweight}.}.

\begin{table}[h]
\centering
\caption{Descriptive statistics of variables used in experiment. \label{tab-descrpdepre}}
\resizebox{0.95\textwidth}{!}{
\begin{tabular}{llllll}
    \hline
    \textbf{Outcome} & Mean  & Std Dev & Minimum & Maximum & Type \\
    \hline
    Center for Epidemiologic Studies Depression (CES-D) & 1.52  & 2.05  & 0     & 8     & Categ \\
    \hline
    \textbf{Variables} &       &       &       &       &  \\
    \hline
    Age in years (AGE) & 67.42 & 10.99 & 18    & 104   & Cont  \\
    Degree of education (EDU) & 12.83 & 3.44  & 0     & 18    & Categ   \\
    Marital status (MS) & 30.13 & 18.19 & 0     & 74.2  & Cont  \\
    Out-of-pocket expenditure (OPE) & 2,978.72 & 7,696.11 & 0     & 232,255.46 & Cont \\
    Body mass index (BMI) & 28.24 & 7.14  & 0     & 76.6  & Cont \\
    \hline
    \end{tabular}%
}
\end{table}

\begin{figure}[h]
\centering
{\includegraphics[scale=0.35]{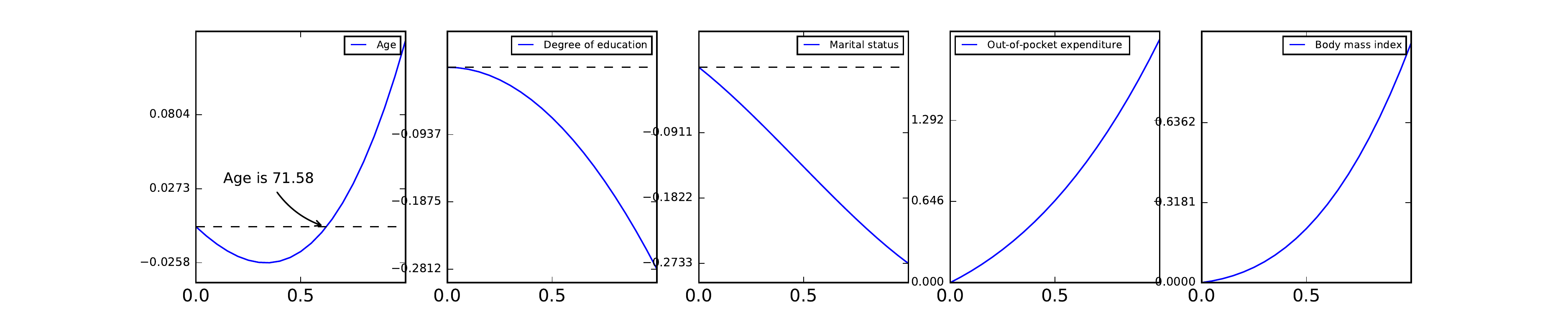}}
\caption{Marginal value functions obtained by NN-MCDA for the depression data. From the left to right, the attributes are age, degree of education, marital status, out-of-pocket expenditure, and BMI. We use polynomial functions of degree 3 to approximate the model. The black dashed line is the baseline rate satisfying $p(\hat{y}=1|\mathbf{x})=p(\hat{y}=0|\mathbf{x})=0.5$. Same in following Figures. \label{fig-depre-nnmcda}}
\end{figure}

\begin{figure}[h]
\centering
{\includegraphics[scale=0.35]{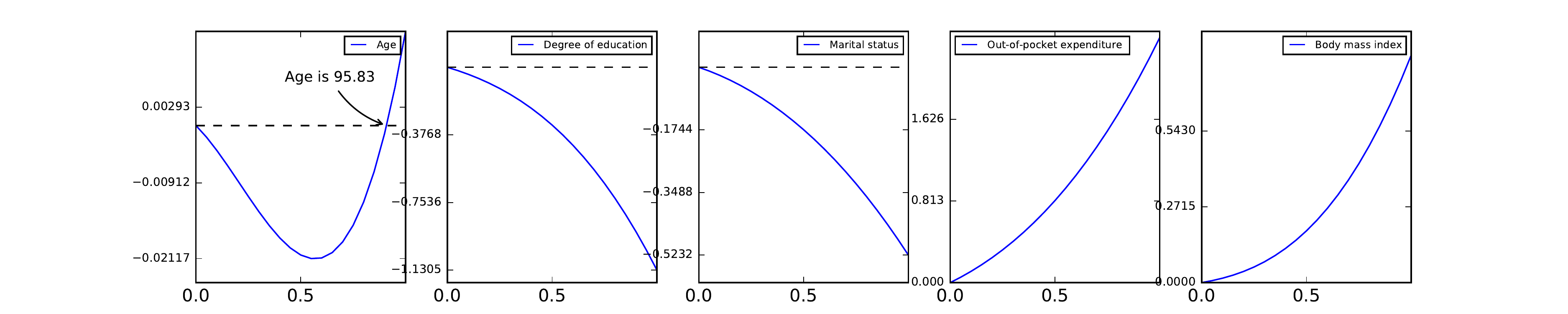}}
\caption{Marginal value functions obtained by logistic regression model for the depression data. \label{fig-depre-linear}}
\end{figure}

\begin{figure}[h]
\centering
{\includegraphics[scale=0.35]{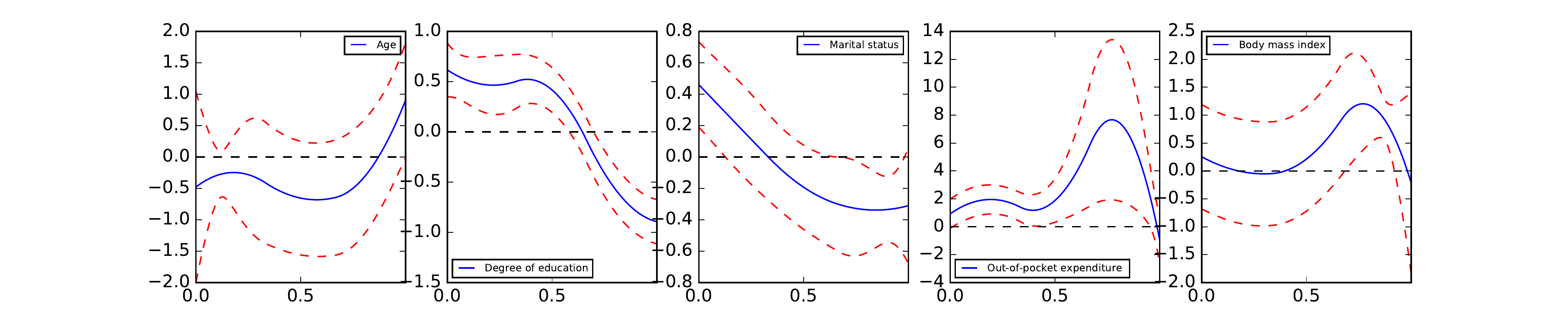}}
\caption{Marginal value functions obtained by GAM with 5 splines that are in 3-degree polynomials for the depression data. \label{fig-depre-GAM}}
\end{figure}

The respondents with CES-D scores higher than or equal to 1 are assumed to be at risk for depression (positive samples). There are totally 9,816 positive samples and 7,880 negative samples. We randomly choose 90\% from both positive and negative samples to train the model. We train the NN-MCDA, MLP, GAM, and logistic regression model for 30 times, and present the average results in Table \ref{tab-preddepreMSE}. The obtained marginal value functions are visualized in Figures \ref{fig-depre-nnmcda}, \ref{fig-depre-linear}, and \ref{fig-depre-GAM}.

\begin{table}[htbp]
\centering
\caption{Averaged AUC and $\alpha$ for training and testing sets. \label{tab-preddepreMSE}}
\resizebox{0.85\textwidth}{!}{
\begin{tabular}{llll}
    \hline
          & Training set & Testing set & $\alpha$ \\
    \hline
    NN-MCDA & 0.678 & 0.669 & 0.393 \\
    Logistic Regression & 0.621 & 0.593 & 1 \\
    MLP & 0.698 & 0.668 & 0 \\
    GAM (5 splines) & 0.628 & 0.613 & - \\
    \hline
    \end{tabular}
    }
\end{table}

We first analyze the similar conclusions by three baseline interpretable models. For the last four attributes, both NN-MCDA and logistic regression models capture similar monotonic trends. As for \textit{degree of education} and \textit{marital status}, the curves are under the baseline rate indicating higher education and the longer length of marriage can reduce the risk for depression. This is consistent with the medical literature \citep{penninx1998depressive, ladin2008risk}. More specifically, both models find that the attributes \textit{out-of-pocket expenditure} and \textit{Body mass index} would increase the risk for depression. Since the obtained value functions are in a convex shape, the growth of the risk will increase along with the increase of these attribute values.

Both the NN-MCDA and logistic regression obtain a convex curve for attribute \textit{age} with part of the curve being negative and the rest being positive (see Figures \ref{fig-depre-nnmcda} and \ref{fig-depre-linear}). This indicates that the risk of depression does not increase while aging if the adult is younger than a threshold. The risk of depression increases fast after an adult passes an age threshold. The threshold for NN-MCDA is 71.58, which makes sense because most adults younger than 71.58 could be enjoying their retirement and their body functions do not degrade much. However, the threshold for the logistic regression is 95.83, which seems unrealistic and inconsistent with the literature \citep{blazer1991association}. 

Similar to the previous experiment, GAM obtains less stabler curves, which are relatively more difficult to interpret. For attribute \textit{age}, GAM obtains similar patterns as NN-MCDA and logistic regression models with an age threshold around 90, which is inconsistent with the literature \citep{ blazer1991association}. For attributes \textit{degree of education} and \textit{marital status}, GAM even obtains quite counter-intuitive (if not wrong) results, indicating that the increase in educational and marriage time results in the higher risk of depression.

\subsection{Predicting the success of bank telemarketing.}
\label{subsec-predtele}

In this section, we employ a benchmark telemarketing datasets \footnote{The dataset is available on \url{https://archive.ics.uci.edu/ml/datasets/Bank+Marketing}. The descriptions about the attributes and data, please refer to the website.} to further verify the efficacy of NN-MCDA. In this task, we aim to predict the success of selling bank long-term deposits by making telemarketing calls \citep{moro2014data}. After data preprocessing, there are 40,787 records and 48 attributes (11 of them are numeric and the rest are binary). In this experiment, we focus on the contributions of 11 numeric attributes in the linear component, and import all attributes to the nonlinear component. We utilize the 3-degree polynomial functions and randomly choose 80\% of data to train the NN-MCDA model for 100 times. In addition to leading to comparable prediction performance (the averaged AUC is around 0.889), NN-MCDA can provide explicit marginal value functions to depict the detailed contributions of individual attributes.

\begin{figure}[h]
\centering
{\includegraphics[scale=0.3]{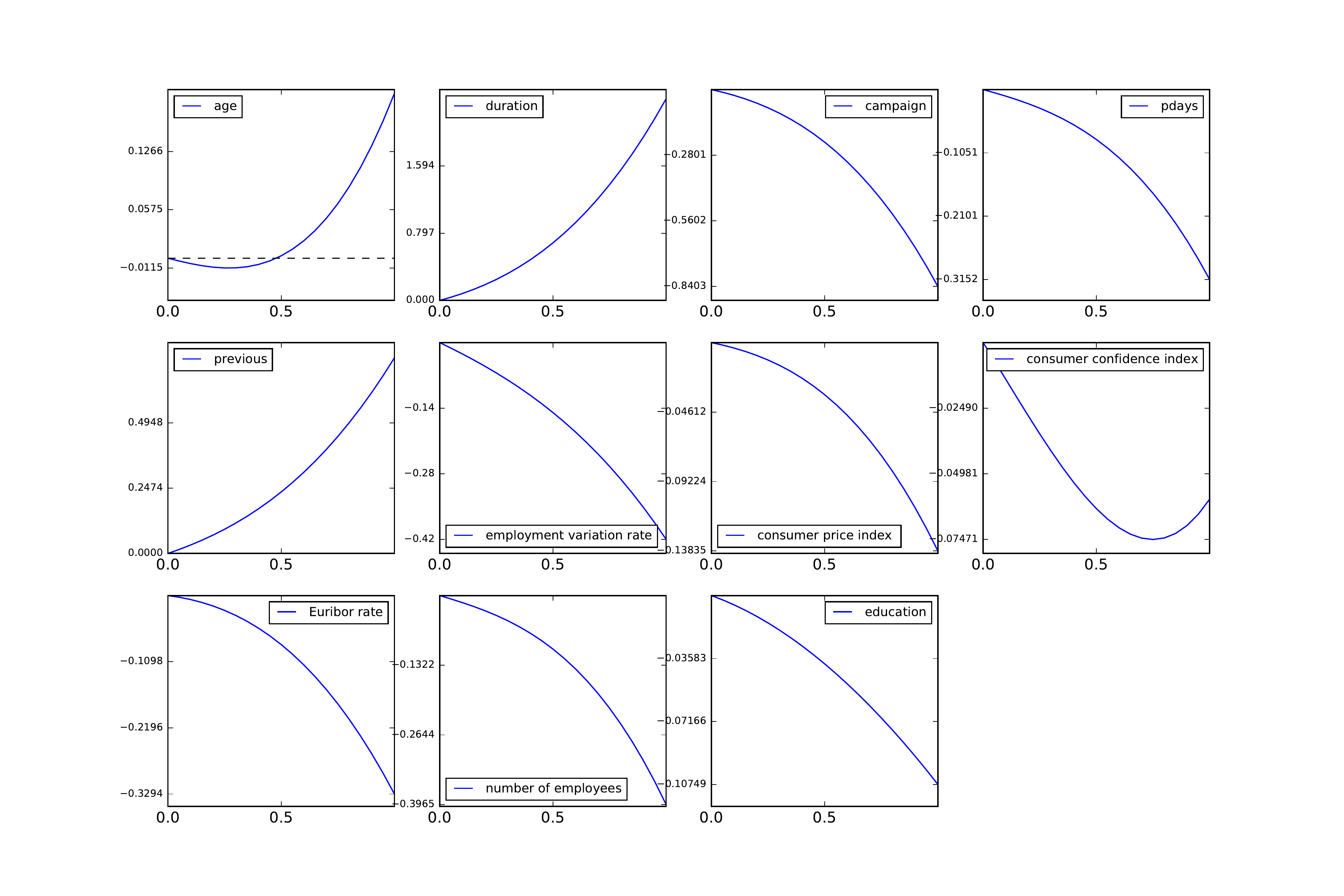}}
\caption{Marginal value functions obtained by NN-MCDA for the telemarketing data. \label{fig-bank}}
\end{figure}

Figure \ref{fig-bank} presents the normalized marginal value functions for all 11 numeric attributes.\footnote{duration: last contact duration; campaign: number of contacts performed during this campaign and for this client; pdays: number of days that passed by after the client was last contacted from a previous campaign; previous: number of contacts performed before this campaign and for this client.} We find that the marginal value function of the age indicates that when a client is younger than 55, he or she is less likely to buy long-term deposit. However, over the course of aging, the client tends to save for future costs. That implies that an experience DM (a telemarketing campaign manager) should target at older clients rather than younger ones. Another interesting pattern is about the Euribor rate. NN-MCDA shows that the Euribor rate is negatively associated with the probability for long-term deposit subscription, and such probability declines more quickly along with the increase of the Euribor rate. Consumer confidence index (CCI) measures the consumers' attitude towards the economy. The original value of the consumer confidence index ranges from -50.8 to -26.9 indicating that the clients are pessimistic towards investments, and thus they may tend to purchase less long-term deposits and keep the cash. Nevertheless, while the CCI increases, the probability of long-term deposit subscription starts to increase as well because the clients becomes more confident in the economy. Similarly, we can interpret the marginal value functions for all variables. Such interpretability extends the DM's understanding of the clients' investment propensity, and could inform better telemarketing strategy.

The experiments with real data demonstrate that the proposed NN-MCDA can effectively capture the patterns in human decision making through learning the marginal value functions, which characterize the contribution of individual attributes to the predictions. The NN-MCDA presents good potential in enhancing the empirical studies through providing a detailed marginal value function instead of a single coefficient for each attribute. Moreover, the prediction performance of NN-MCDA is close to a full complexity model (MLP), and much better than that of baseline interpretable models (GAM and logistic regression model). This finding is further verified through comparing its performance with that reported in the literature.

\section{Discussion}
\label{sec-dis}

In this section, we summarize the insights from the experiments, discuss the use and extension of the proposed NN-MCDA model, and compare the proposed NN-MCDA with ensemble learning.

\subsection{Attribute importance.}
\label{subsec-attimp}

\begin{figure}[h]
\centering
{\includegraphics[scale=0.25]{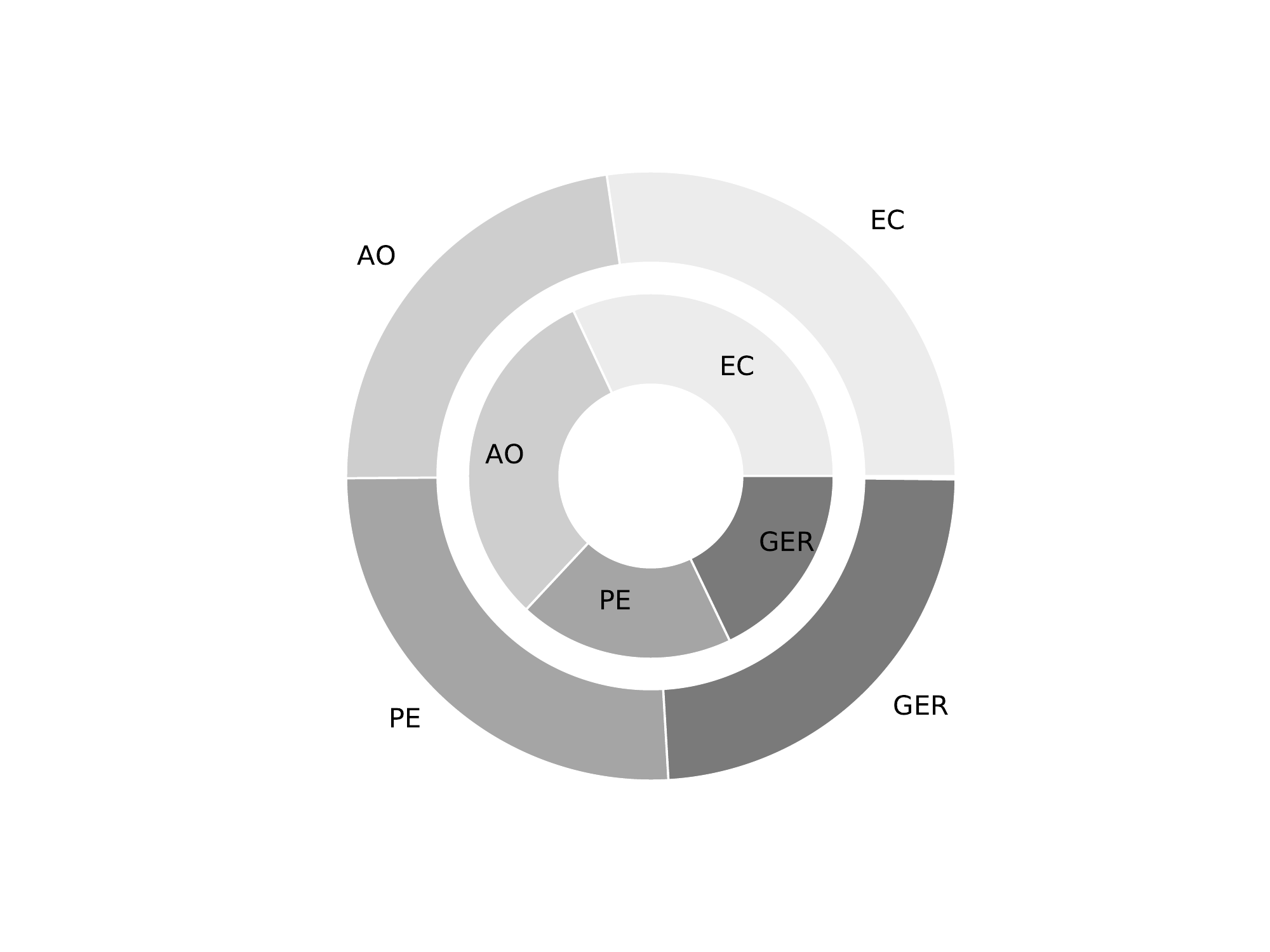}
\includegraphics[scale=0.25]{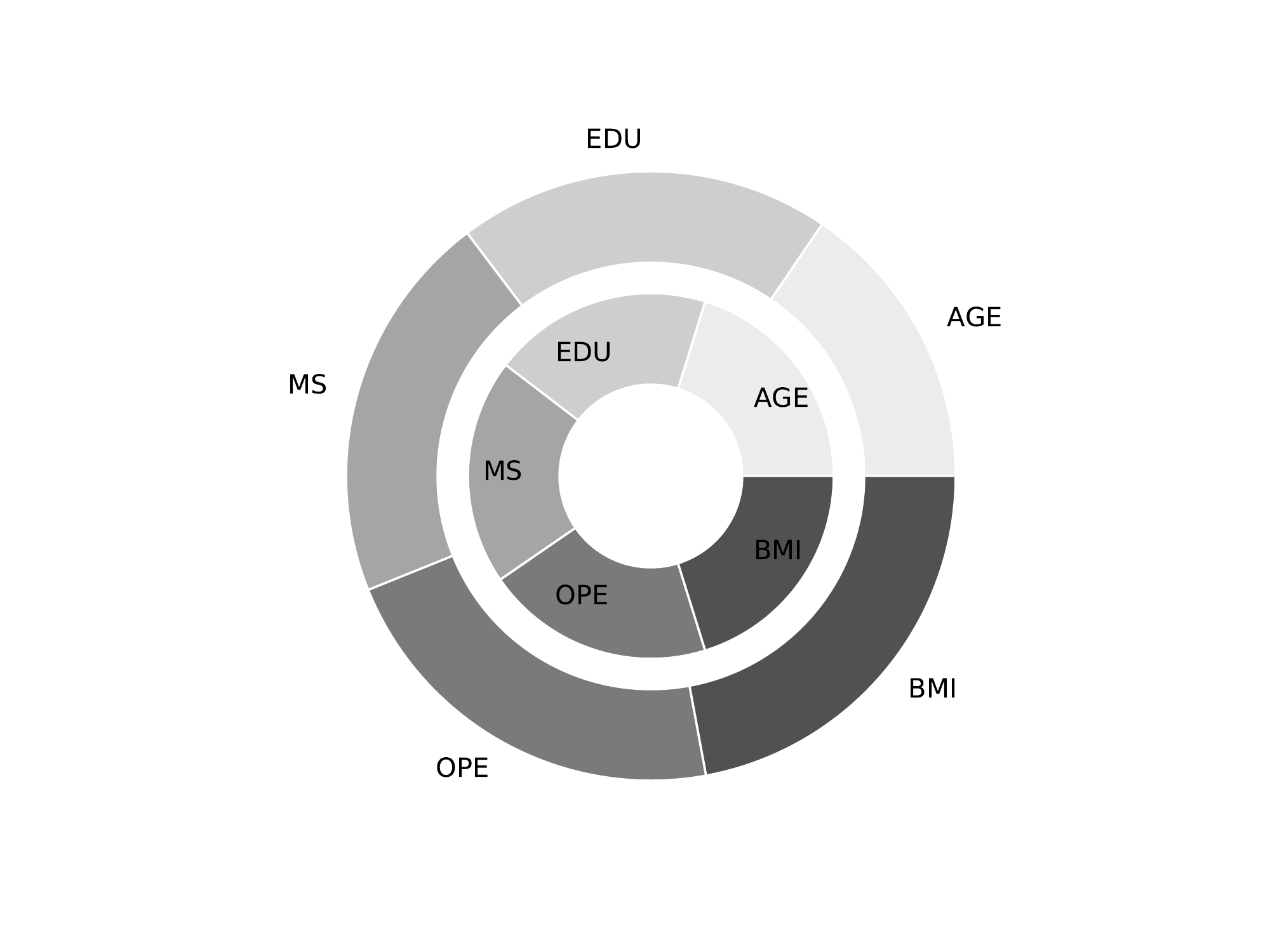}
}
\caption{Normalized attribute weights in univerisity employee reputation ranking and geriatric depression prediction. The inner circles are normalized attribute weights that are obtained by linear regression model and the outer ones are obtained by NN-MCDA.\label{fig-attributeweight}}
\end{figure}

To explain the importance of each attribute, we present the normalized attribute weights obtained from previous experiments in Figure \ref{fig-attributeweight}. In the university ranking problem, NN-MCDA assigns 0.2732, 0.228, 0.258, and 0.239 to attributes \textit{employer-student connection}, \textit{alumni outcomes}, \textit{partnerships with employers} and \textit{graduate employment rate} whereas a regression model assigns 0.3198, 0.3107, 0.1906, and 0.1789 to them, respectively. Both models determine that \textit{employer-student connection} is the most important attribute, however, they give different orders of other three attributes. Given the limited resources and the obtained importances of attributes, maintaining a good employer-student connection with frequent employer presences on campus is the most effective method to achieve good employer reputation of a university.

As for the depression prediction problem, a regression model assigns almost equal importance to the five attributes, which is not intuitive for the DM. On the other hand, the NN-MCDA provides a an order of the attributes according to the importance: $ BMI \sim OPE \succ MS \sim EDU \succ AGE$ (0.221, 0.218, 0.208, 0.198, 0.155). Such order suggests that obesity and loads of expenditure are the most important risk factors of becoming depressed. While estimating an old adult's risk for depression, a DM (general physician, specialists and geriatricians) should prioritize the problems related to the older adult's body weight and economical conditions.

\subsection{Interpreting the trade-off coefficient.}
\label{subsec-pretest}

As a key coefficient, determining the value of $\alpha$ is important to practical applications. It reflects the influence of high-order interactions and complex nonlinearity of variables on the final decisions. In this regard, the NN-MCDA model can be used to explore the complexity of the learning problem. If the convergence $\alpha$ is very small, it indicates that the data is highly complex and the assumption of preference independence is not valid. The DM should then use latest models that account for attribute interactions, such as a Choquet integral-based model \citep{aggarwal2019modelling} or full complexity machine learning models. On the contrary, if $\alpha$ is close to 1, the DM is recommended to use a simpler model to avoid massive computational time and non-interpretable results. 

In the case of prediction for depression (Table \ref{tab-preddepreMSE}), the convergence $\alpha$ is around 0.4. It indicates that the involved attributes are possibly interacted in this problem. Some related medical studies also empirically demonstrated some interactions between attributes, for example, older adults that are relatively young with higher degree of education may be involved in more social activities and have a more contented life, which lead to lower risk for depression \citep{li2014meta}. Moreover, these older adults with longer marriage will obtain more family support and thus they have lower risk for depression \citep{pearlin1977marital}. We usually address the pairwise interactions because they are easier to interpret and can be visualized by a heating map \citep{caruana2015intelligible}. Given convergence $\alpha$ and marginal value functions obtained \textit{in the presence of} high-order correlations among attributes, we could develop algorithms to use lower-ordered attribute interactions, e.g.: pairwise and triple-wise interactions, to approximate the higher ones. The framework can then be extended to a two-step procedures, including determining marginal value functions and deducing possible lower-ordered correlations.

\subsection{Extending the NN-MCDA framework.}
\label{subsec-modification}

The proposed NN-MCDA presents a general modelling framework, which can be easily extended to enhance the performance and adaptivity for various problems. In this section, we discuss the three extensions, including adding regularizations replacing the model in the nonlinear component, and incorporating attributes in the nonlinear component.

\subsubsection{Adding regularizations}
\label{subsubsec-reg}

For some mission-critical cases where the data is complex, the convergence $\alpha$ could be very small. However, the DM still requires certain level of model interpretability to facilitate their decision making. Therefore, we opt for adding a regularization term to prevent the model from being too complicated and non-interpretable. The inclusion of the regularization term also helps prevent the over-fitting problem. For example, we can revise the original MSE as follows $MSE_1 = MSE + (1-\alpha)^2$. The added regularization term $(1-\alpha)^2$ allocates more weight to the linear component at the cost of lower fitting accuracy. We can also change the regularization term to $(2\alpha-1)^2$, which leads to a balanced model that favors a model with equal weights to the linear and nonlinear components. The exact form of the loss function should be selected according to the problem settings.

\subsubsection{Replace the model in the nonlinear component.}
\label{subsubsec-rep}

Given different types of datasets, the proposed NN-MCDA model can be modified by replacing the neural networks in the nonlinear component by other network structures. In subsection \ref{subsubsec-exp1}, NN-MCDA has difficulty in handling extremely complex data ($\mathcal{D}^n_{polynomial-3}$ and $\mathcal{D}^n_{polynomial-15}$). To improve the performance on this data, we can introduce more layers in the MLP or increase the number of neurons in each layer. For image classification problem, we can replace the MLP with a convolutional neural network (CNN), and use the features obtained from CNN as the input for the linear component. To fit time-series or free text data, we can replace the MLP with a recurrent neural network.

\begin{figure}[h]
\centering
{\includegraphics[scale=0.4]{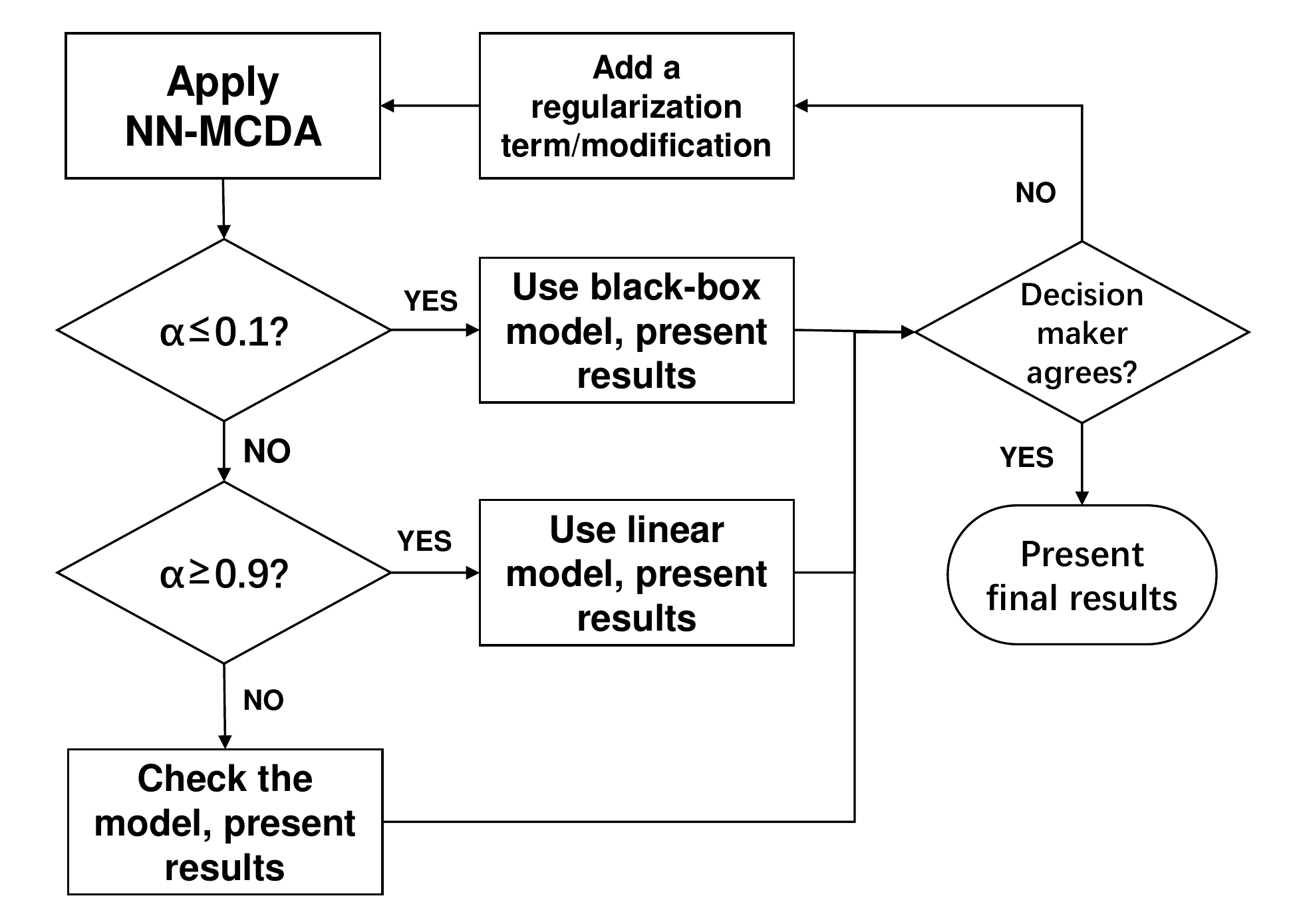}}
\caption{A heuristic user-interactive process for modifying the proposed NN-MCDA. \label{fig-flowchart}}
\end{figure}

In addition, the proposed model can be progressively modified by iteratively interacting with the DM. We provide a user-interactive process to determine the ultimate model. The framework is shown in Figure \ref{fig-flowchart} and explained as follows:
\begin{enumerate}
\item[Step 1.] We first apply the NN-MCDA model to the management problem. While the model converges, we obtain the value of $\alpha$.
\item[Step 2.] If $\alpha > 0.1$, go to step 3. If $\alpha \leq 0.1$, which indicates that the data are potentially very complex. We opt for a full complexity black-box model to achieve higher accuracy and present the results to the DM. If the DM agrees to use the black-box model, the process is end. Otherwise, we add a regularization term (e.g.: use $MSE_1$) to the original NN-MCDA, and go back to Step 1.
\item[Step 3.] If $\alpha < 0.9$, go to step 4. If $\alpha \geq 0.9$, which indicates that a simple additive model is sufficient to fit the data, we explain the results to the DM. If the DM is satisfied with the accuracy, the process ends. Otherwise, we can modify the NN-MCDA model by increasing the complexity of the nonlinear component (e.g.: using deeper MLP or other neural net-based model). Then, we go back to Step 1.
\item[Step 4.] If $0.1 < \alpha < 0.9$, we present the underlying model and results to the DM. If there are no further requirements, the process is end. If the DM requires further modifications (such as adding regularization terms or modifying the nonlinear component), we modify the model accordingly and then go to Step 1.
\end{enumerate}

\subsubsection{Flexible inclusion of attributes in the nonlinear component}
\label{subsub-flexileattri}

In practice, the human decision making usually focuses on a small number of key attributes/criteria \citep{ribeiro2016should}. However, there could exist other minor attributes that do not directly contribute to the prediction, but could affect the prediction through non-traceable complex interactions with other attributes (for example, the interaction between the nonlinear transformation of an attribute and the nonlinear transformation of five other attributes). These minor attributes can be incorporated by the nonlinear component.

\begin{figure}[h]
\centering
{\includegraphics[scale=0.4]{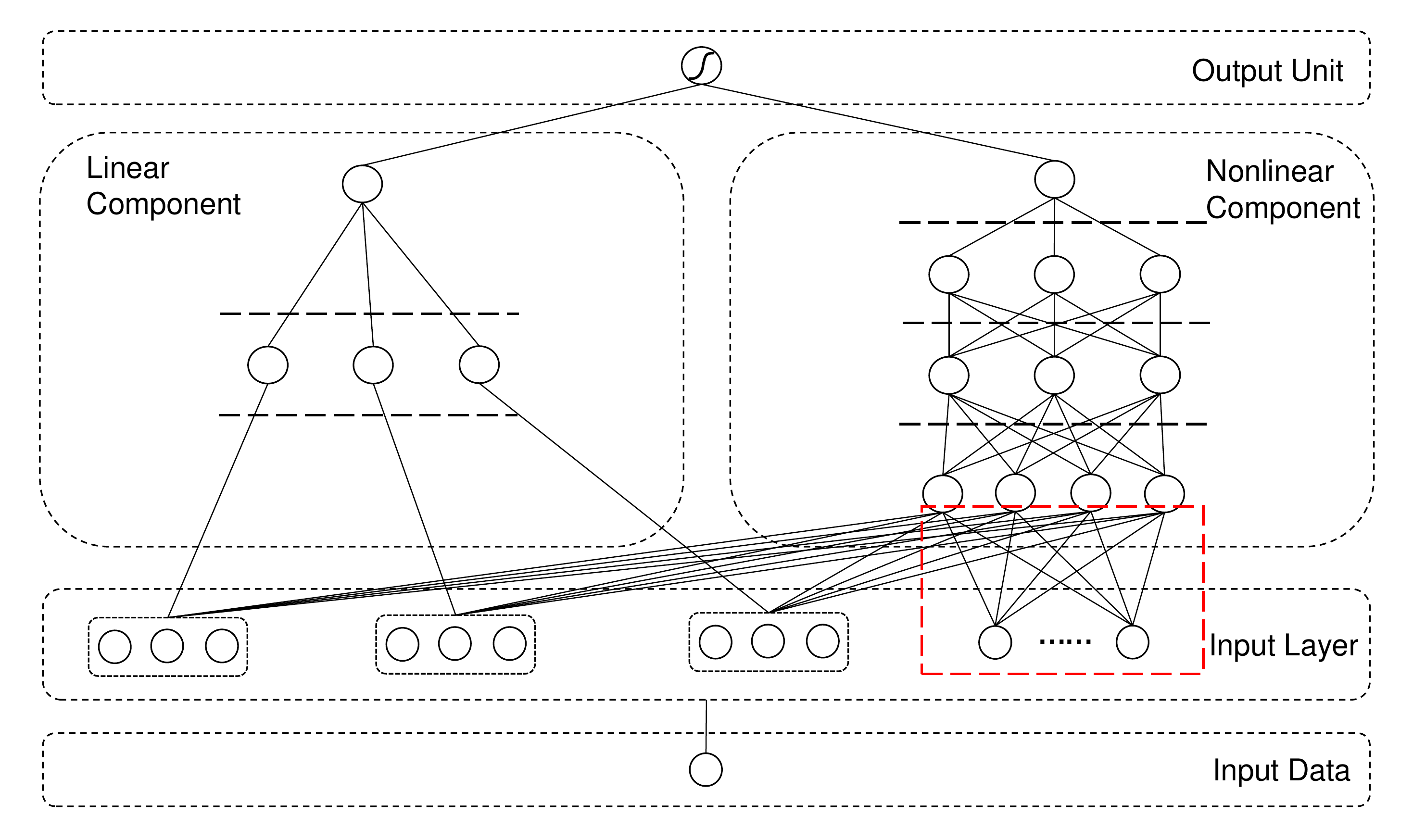}}
\caption{A new framework for NN-MCDA considering more attributes in the nonlinear component. \label{fig-framework2}}
\end{figure}

\begin{figure}[h]
\centering
{\includegraphics[scale=0.3]{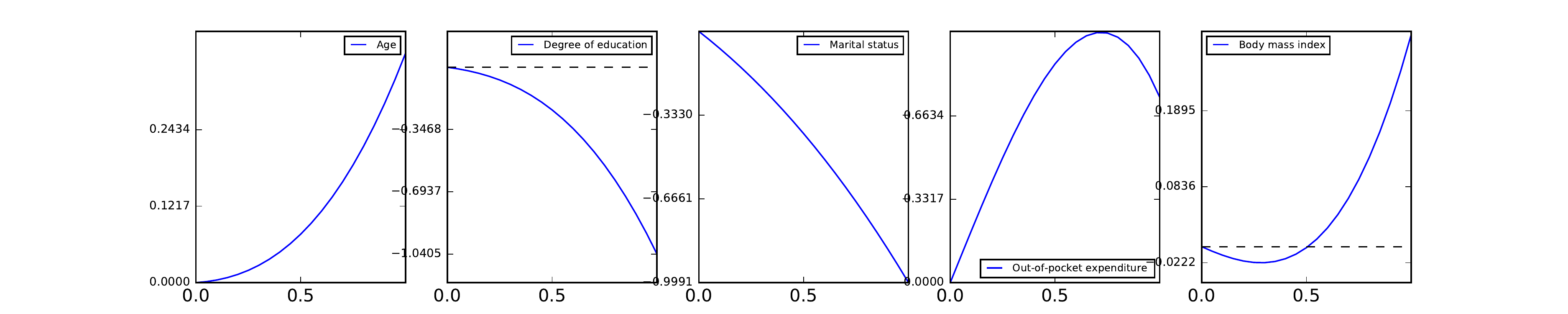}}
\caption{For depression data, the marginal value function obtained by linear component while adding two more attributes to the nonlinear component. \label{fig-depressFunc_1}}
\end{figure}

\begin{figure}[h]
\centering
{\includegraphics[scale=0.3]{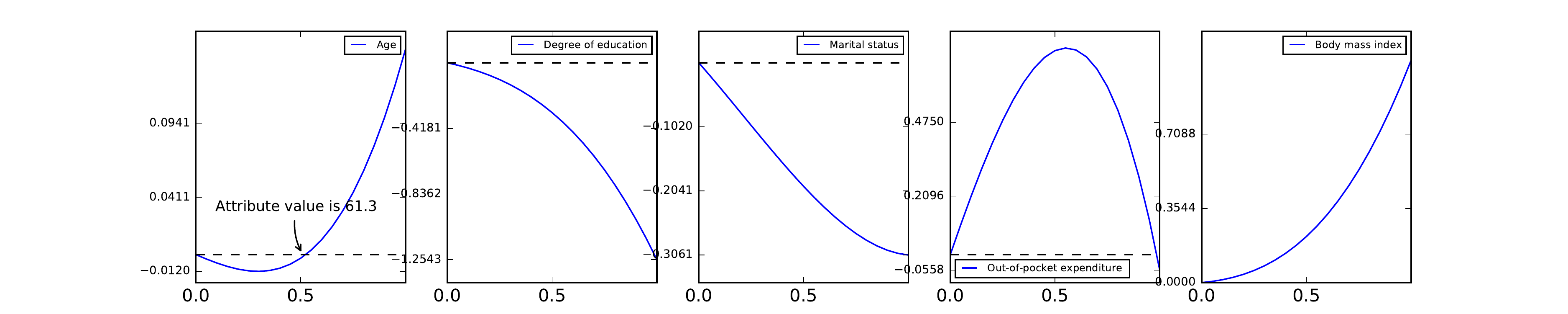}}
\caption{For depression data, the marginal value function obtained by linear component while adding four more attributes to the nonlinear component. \label{fig-depressFunc_2}}
\end{figure}

In the geriatric depression experiment, the gender of an older adult may not directly indicate a difference in the risk for depression, however, it might still influence the prediction through complex interactions with other attributes. We further extend the NN-MCDA model in incorporate gender and smoking status into the nonlinear component (as shown in Figure \ref{fig-framework2}). We find that the incorporation of these attributes indeed improves the prediction accuracy (the AUC for testing set increases from 0.669 to 0.675), while still maintains the similar marginal value functions in the linear component (see Figure \ref{fig-depressFunc_1}). If we add two more attributes, for instance whether the respondent received any home cares in last two years and whether the health problem limited his/her work, the AUC increases more obviously (from 0.675 to 0.708) and the marginal value functions still provide convincing results (see Figure \ref{fig-depressFunc_2}).

\subsection{Joint training process.}
\label{subsec-joint}

In the NN-MCDA model, the linear and nonlinear components are combined by a trade-off coefficient $\alpha$. Their sum is then fed to a common logistic function for a joint training process. Note that this joint training process is different from ensemble learning \citep{cheng2016wide}, in which multiple classifiers are trained individually and their predictions are simply combined after every model is optimized separately. For example, an ensemble learning approach could have a linear logistic regression model and an MLP model to make predictions for the same dataset separately, and then integrate the prediction results of the two models. The joint training process indicates that the linear and nonlinear components are connected. While we tune the parameters in one component, the other component will be affected. If the model is at its global optimal, the predictions can be made.

\section{Conclusion and future work.}
\label{sec-conc}

In this paper, we proposed a framework for a novel hybrid machine learning model, namely the NN-MCDA, which combines traditional MCDA model and neural networks. MCDA uses marginal value functions to describe the explicit contribution of individual attributes to the predictions, while neural network considers the implicit high-order interrelations among attributes. The framework automatically balances the trade-off between two components. NN-MCDA is more interpretable than a full complexity model and maintains similar predictability.

We present simulation experiments to demonstrate the effectiveness of NN-MCDA. The experiments show that (1) polynomial of higher degrees do not always improve on accuracy; (2) There is a trade-off between the interpretability and the predictability of the model. NN-MCDA can achieve a good balance between them; (3) Given simple data, NN-MCDA performs as good as interpretable model, while given more complex data, NN-MCDA outperforms an interpretable model. We also present how to apply the NN-MCDA framework to real-world decision making problems. These experiments with real data demonstrate the good prediction performance of NN-MCDA and its ability in capturing the detailed contributions of individual attributes.

We envisage the following directions for future researches based on the NN-MCDA framework. First, we can further enhance the interpretability of the model through proposing algorithms to approximate the attribute interactions after obtaining the marginal value functions. Second, further studies are needed to validate the effectiveness of the NN-MCDA variants that introduced in the discussion section. Last, but not the least, applying the proposed framework to a variety of real-world decision making and prediction problems constitutes another interesting direction for future work.


\appendix

\section{Proof of Proposition 1}
\label{app-1}
\begin{proof}
Assume the activation function has the following linear form $a_l(\mathbf{x})=\mathbf{A}_l^T\mathbf{x}+\mathbf{b}'_l$. The output of the $l$-th layer is \begin{align}
    \mathbf{z}_l(\mathbf{x})&=a_l(\mathbf{W}^T_l\mathbf{z}_{l-1}+\mathbf{b}_l)=\mathbf{A}_l^T(\mathbf{W}^T_l\mathbf{z}_{l-1}+\mathbf{b}_l)+\mathbf{b}'_l \\
    &= (\mathbf{W}_l\mathbf{A}_l)^T\mathbf{z}_{l-1}+(\mathbf{A}_l^T\mathbf{b}_l+
\mathbf{b}'_l)   
\end{align}
leading to a linear combination at each unit. The global value of the $i$-th attribute vector is 
\begin{align}
    U(\mathbf{x}_i) &= \alpha {z}_{i}^{linear} + (1-\alpha) {z}_{i}^{nonlinear}\\
    &= \alpha \mathbf{w}^T\mathbf{P}_i + (1-\alpha) \mathbf{h}^T((\mathbf{W}_{L-1}\mathbf{W}_{L-2}\dots\mathbf{W}_1\mathbf{A}_1\dots\mathbf{A}_{L-1})^T\Phi(\mathbf{x}_i)+\mathbf{b})
\end{align}
is apparently in an additive form. 
\end{proof}

\section{Proof of Proposition 2}
\label{app-2}
\begin{proof}
Without loss of generality, assume the activation function is a sigmoid function. When $L=1$, assume there are $K_1$ units in the first layer and the output of $k$-th unit in the nonlinear component is $z_1^k(\mathbf{x}_i) = \sigma(\mathbf{w}_k^T\mathbf{x}_i+b_k)$. The output of this layer is 
\begin{align}
    z_i^{nonlinear}&=\sum\nolimits_{k = 1}^{{K_1}} {{h_k}\sigma \left( {w_1^k{x_i} + b_1^k} \right)} \\&= f(x_{i,1},x_{i,2},\dots,x_{i,n}) \label{eq-nonlinear_simple} 
\end{align}
where $f(x_{i,1},x_{i,2},\dots,x_{i,n})$ is a function of each attribute value. For example, when $K_1=2$, we have
\begin{align}
f(x_{i,1},\dots,x_{i,n}) = \frac{h_1+h_2 + h_2 e^{-(\mathbf{w}_1^T\mathbf{x}_i+b_1)} + h_1e^{-(\mathbf{w}_2^T\mathbf{x}_i+b_2)}}{1+e^{-(\mathbf{w}_1^T\mathbf{x}_i+b_1)}+e^{-(\mathbf{w}_2^T\mathbf{x}_i+b_2)} + e^{-(\mathbf{w}_1^T\mathbf{x}_i+\mathbf{w}_2^T\mathbf{x}_i+b_1+b_2)}}  
\end{align}
where $\mathbf{w}_k^T\mathbf{x}_i = \sum\nolimits_{j = 1}^{{n}} {w_k^jx_{i,j}}$ and $k=\{1,2\}$. When $L\ge 2$, let $f_{l-1}^k(x_{i,1},\dots,x_{i,n})$ be the output of the $k$-th unit in the previous layer. We have
\begin{align}
input_2^u &= \sum\nolimits_{k = 1}^{{K_1}} {w_1^{u,k}f_1^k\left( {{x_{i,1}}, \ldots {x_{i,n}}} \right)} ,\\
output_2^u &= \sigma \left( {input_2^u} \right) = f_2^u\left( {f_1^1\left( {{x_{i,1}}, \ldots {x_{i,n}}} \right), \ldots f_1^{{K_1}}\left( {{x_{i,1}}, \ldots {x_{i,n}}} \right)} \right)\\
 &\vdots \nonumber\\
output_L^u &= f_L^u ( {f_{L - 1}^1\left( { \cdots f_2^{{K_2}}\left( {f_1^1\left( {{x_{i,1}}, \ldots {x_{i,n}}} \right), \ldots f_1^{{K_1}}\left( {{x_{i,1}}, \ldots {x_{i,n}}} \right)} \right)} \right)}, \label{eq-nonlinear}\\ 
& \quad f_{L-1}^{2}\left( \cdots \right),     \dots, f_{L-1}^{K_{L-1}} \left( \cdots \right)    ), \nonumber \\
z_i^{nonlinear} &= \sum\nolimits_{u = 1}^{{K_L}} {{h_u}output_L^u,} 
\end{align}
where $input_l^u$ is the input of the $u$-th unit at the $l$-th layer. The form of Eq.(\ref{eq-nonlinear}) is complex and it is difficult to simulate its explicit form. Apparently, there are higher-order transformations of attribute interactions, which are more complicated than the forms in Eq.(\ref{eq-nonlinear_simple}).
\end{proof}
\bibliographystyle{elsarticle-harv} 
\bibliography{mybibfile}





\end{document}